\documentclass[11pt,letterpaper]{article}
\usepackage[T1]{fontenc}
\usepackage[utf8]{inputenc}
\usepackage{amsmath,amssymb,bm}
\IfFileExists{newtxtext.sty}{\usepackage{newtxtext,newtxmath}}{\usepackage{mathptmx}}
\usepackage[margin=0.82in,top=0.62in,bottom=0.64in,footskip=0.34in]{geometry}
\usepackage{graphicx,array,booktabs,longtable,tabularx}
\usepackage[table]{xcolor}
\usepackage{caption,float,placeins,flafter}
\usepackage{listings,fancyhdr}
\usepackage[hidelinks,breaklinks=true]{hyperref}
\usepackage{xurl}
\fancypagestyle{plain}{\fancyhf{}\fancyfoot[R]{\thepage}
  }
\definecolor{listingbg}{HTML}{F1F4F5}
\definecolor{tablehead}{HTML}{E3ECED}
\hypersetup{pdfauthor={Sabik Bin Sultan; Shafi Bin Sultan; Safwan Sadad}}
\begin{document}

\begin{center}
{\LARGE\bfseries Zephyron: Integrated Design and Analytical Evaluation of a Solar-Assisted Mobile Manipulator for Multimodal Environmental Reconnaissance and Distributed Visual Inference\par}
\vspace{9pt}
Sabik Bin Sultan\textsuperscript{1}   Shafi Bin Sultan\textsuperscript{2}   Safwan Sadad\textsuperscript{3}\\[4pt]
{\small \textsuperscript{1} Bangladesh Air Force Shaheen College Kurmitola | sabikbinsultan@gmail.com\\
\textsuperscript{2} St.Joseph Higher Secondary School | shafibinsultan0207@gmail.com\\
\textsuperscript{3} Greenland Residential School | Avoidsafwan@gmail.com}
\end{center}
Literature-informed design and analytical evaluation

\FloatBarrier
\section*{Abstract}

Environmental reconnaissance requires mobile platforms to transport sensors, preserve measurement context, and return interpretable evidence under limited energy and communication availability. This study develops a literature-informed engineering design for Zephyron, a four-wheel rover combining a front manipulator, environmental screening instruments, distributed computer vision, local recording, and a raised rear solar module. The design preserves the photographed prototype arrangement while replacing unsupported numerical assumptions with an explicit component and geometry baseline. A reproducible search retrieved 5,000 bibliographic records and retained 4,858 unique records for metadata screening, followed by targeted examination of primary literature and manufacturer documentation. The analytical baseline uses 165 mm wheels, a 12 kg gross mass budget, a 72 Wh conservative battery-energy basis, and a 20 W photovoltaic module. Under an assumed rolling-resistance coefficient of 0.04, steady ascent of a 10 degree grade requires approximately 0.517 N m per wheel with equal load sharing. An illustrative 40 W continuous-motion load yields 1.44 h from 57.6 Wh usable energy; a 25 percent driving duty yields 4.19 h without solar input. These are calculated scenarios, not measured rover performance. Sensor models show why integration time, calibration, temperature, and communication delay constrain mission interpretation. A quality-aware stop-and-sample policy connects those constraints to mission execution. Candidate lightweight detectors, reference-based sensor learning, and executable data-integrity checks define a reproducible machine-learning evaluation pathway. The contribution is a traceable design and evaluation framework supported by editable three-dimensional models, subsystem diagrams, and reproducible analytical data. Experimental validation remains necessary before assigning payload, endurance, detection, or field-operating ratings.

Keywords: mobile robotics; environmental sensing; solar-assisted rover; measurement uncertainty; distributed vision; mission planning; engineering design

\FloatBarrier
\section{Introduction}

Ground robots can place instruments and cameras near environmental hazards while maintaining a separation between an operator and the immediate inspection area. Their usefulness depends on more than the number of sensors mounted to a chassis. A useful measurement must be associated with a sampling procedure, an instrument state, a position, and a time; the operator must also know whether the robot can return with its remaining energy. Rescue-robot and environmental-monitoring studies therefore provide complementary design evidence. Field robotics establishes the importance of mobility, communications, and recoverability, whereas sensing studies establish the limitations of calibration, response time, and spatial sampling~\cite{LIT_QUINCE2011},~\cite{LIT_GAS2019},~\cite{LIT_RADNAV2021}.

Zephyron is developed here as a compact, supervised environmental reconnaissance platform. Photographs establish its recognizable arrangement: four corner wheel modules, an aluminum chassis, a carbon-patterned central body, a forward manipulator, paired front lamps, a camera and display on the nose, and a solar panel carried above the rear chassis. The intended functional scope includes water-quality screening, combustible-gas indication, ultraviolet and radiation monitoring, proximity sensing, outdoor geotagging, visual inspection, local recording, telemetry, and retrieval of small objects. The physical arrangement is treated as a design constraint. Dimensions and ratings are new engineering selections rather than measurements recovered from perspective photographs.

\begin{figure}[htbp]
\centering
\includegraphics[width=0.467\linewidth]{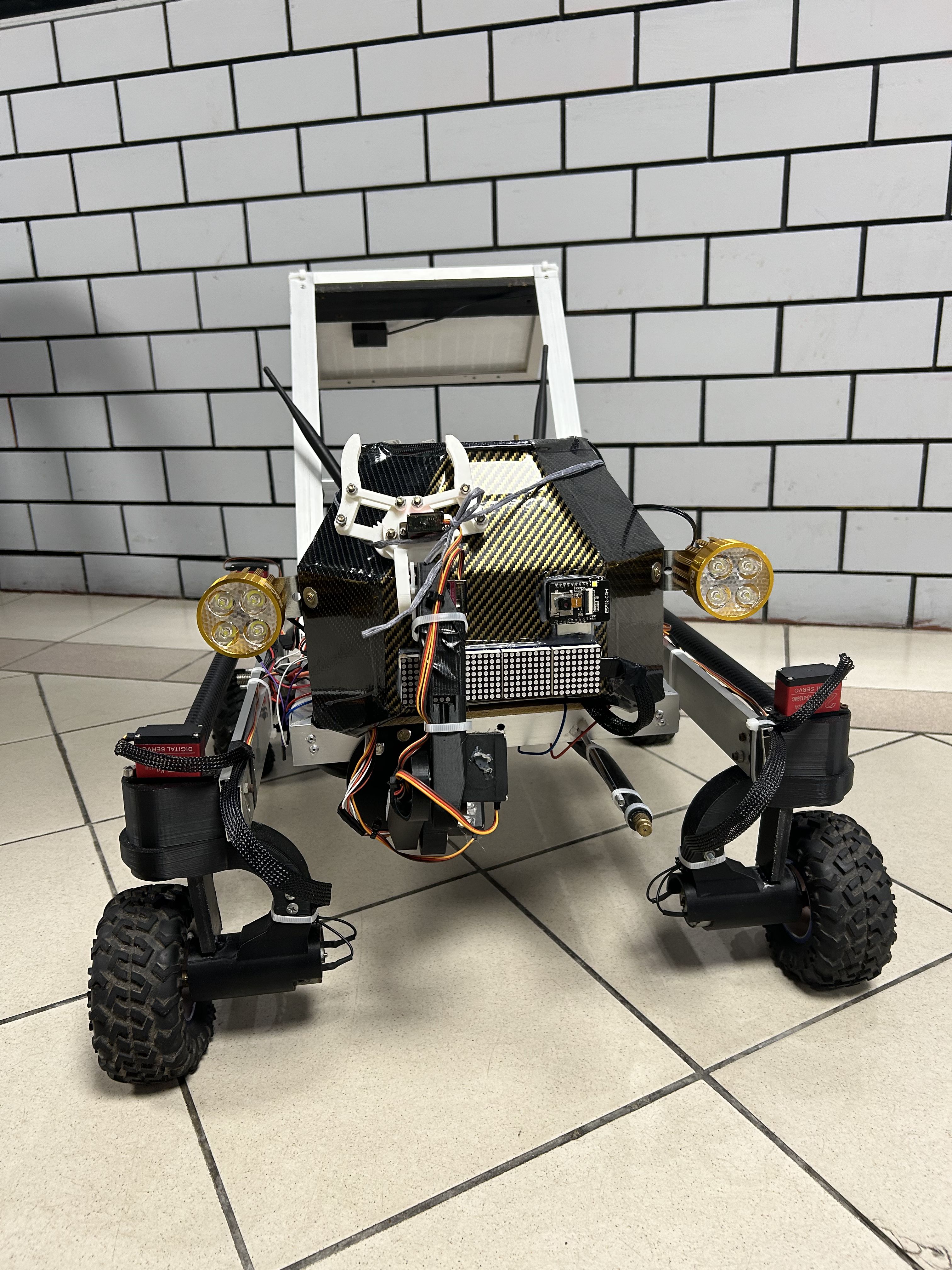}
\hfill
\includegraphics[width=0.467\linewidth]{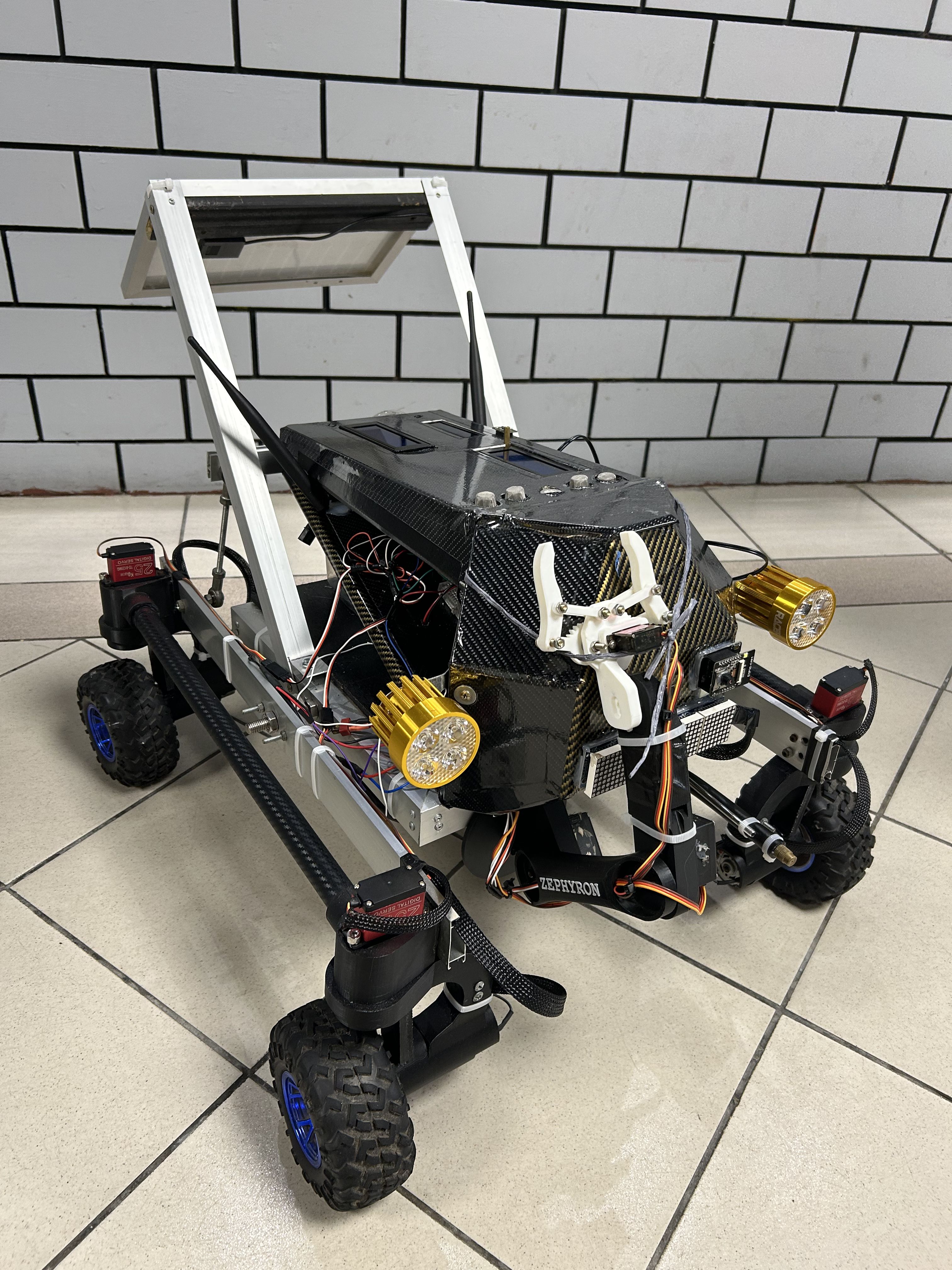}
\caption{Author-supplied photographs of the assembled Zephyron rover. Left: front view showing the folded manipulator, paired lamps, camera and wheel modules. Right: oblique view showing the chassis, electronics bay and elevated rear solar support. Original photographs are reproduced without hardware alteration; no dimensions or performance measurements are inferred.}
\label{fig:PHOTO}
\end{figure}

A common difficulty in low-cost robotics is the transfer of a component specification into an unsupported system claim. A motor's stall torque does not define a continuous wheel load; a camera's recording rate does not establish end-to-end detection latency; and the accuracy of an individual sensor does not determine the uncertainty of a field sample. The same distinction applies to visual evidence. A manufactured prototype photograph supports the existence and arrangement of visible mechanisms, while a Blender rendering supports communication of a proposed design. Neither constitutes a calibrated field trial.

This work addresses three engineering questions. First, can the photographed arrangement be assigned a coherent component, mass, geometry, and power baseline without relying on unsupported prototype numbers? Second, which mechanical and measurement constraints dominate the feasible operating envelope? Third, how should the rover schedule travel, sampling, and reporting so that measurement quality and remaining energy are explicit? These questions are approached through primary-source review, deterministic calculations, and a proposed experimental protocol. The manuscript does not claim a new sensor principle or a trained detection model.

The contribution comprises a traceable baseline; linked mobility, energy, manipulator, and sensing calculations; a mission policy that checks feasibility before accepting an action; and an editable visual and analytical package. The policy is a proposed integration method whose benefit must be tested against matched baselines. This distinction permits a rigorous design paper while identifying the experimental evidence needed for a later performance paper.

\FloatBarrier
\section{Research method and evidence traceability}

\subsection{Source hierarchy and exclusion of unsupported quantities}

Evidence is assigned to five classes: directly observed photographic features; verified component specifications; findings reported by external studies; selected design assumptions; and results calculated from those assumptions. The original concept paper is used only to establish intended functions. All of its numerical claims and purported performance results are excluded from the present analysis. No sample values, repeated trials, error bars, training curves, or statistical significance values have been reconstructed from it.

The supplied photographs do not include a reliable scale reference, and their viewpoints contain perspective distortion and occlusion. Absolute chassis dimensions, component mass, and load ratings therefore cannot be established by photogrammetric inspection alone. Instead, selected component envelopes constrain a new geometry baseline, while photographs constrain relative arrangement. This procedure preserves the recognizable rover without implying that a rendered assembly is an accurate dimensional survey of the earlier physical prototype.

Manufacturer documentation is used for component envelopes, electrical ratings, response characteristics, and stated limitations. Primary papers are used for methods, experimental precedents, and interpretation of sensing or locomotion problems. A benchmark remains associated with its original platform, environment, hardware, and evaluation protocol. When an external study uses simulation or a physical surrogate rather than the environmental quantity of interest, that distinction accompanies its result in the comparison and figure caption.

\subsection{Literature search and focused review}

Ten reproducible Crossref queries covered rescue ground vehicles, mobile environmental sensing, gas mapping, radiation surveys, water sensing, low-cost mobile platforms, and related navigation topics. The search retrieved 500 records per query, producing 5,000 records before deduplication. DOI matching, with normalized-title fallback, reduced this set to 4,858 unique records. A transparent title and available-abstract keyword screen flagged 1,630 records for closer relevance consideration. This is a metadata discovery corpus; it is not a claim that thousands of complete papers were read, and keyword flags are not a formal quality assessment.

The focused platform review examined 25 primary robotics and sensing studies with accessible detail relevant to Zephyron. Additional primary method papers support the proposed learning models and calibration procedures. Search queries, retrieval dates, raw responses, deduplicated metadata, and source-level notes are retained in the accompanying archive. The focused review was purposive rather than a preregistered systematic review. It is appropriate for engineering design support, but it cannot support estimates of the prevalence of methods across the entire robotics literature. Citation counts and journal reputation were not used as substitutes for inspecting experimental conditions.

\subsection{Reproducible analytical workflow}

The selected baseline is recorded separately from calculated quantities. Each deterministic analysis stores its independent-variable grid and resulting values in a machine-readable table. Units accompany variables, and limiting assumptions are stated with each model. Curves represent parameter sensitivity, expected counting uncertainty, or scenario planning. They do not contain synthetic noise intended to resemble experimental observations. Published graphs are reproduced only when the source license permits reuse, with attribution directly beneath the figure.

The Blender model is an editable design asset. Mission scenes depict sampling and inspection arrangements using the same rover assembly, rather than inventing a different platform for each task. Native Canva designs document subsystem flow. Their arrows represent intended information, command, or power relationships; they do not establish a verified wiring harness, software implementation, or integrated test. A manufacturing release would additionally require controlled drawings, tolerances, component selections, circuit schematics, firmware, and acceptance tests.

\FloatBarrier
\section{Related work}

\subsection{Disaster robotics as an integrated operational system}

The Quince redesign illustrates why disaster robots cannot be assessed as an assortment of individually capable parts. Added mission equipment affected vehicle loading, stair traversal and clearances, while communications, operator interfaces and recovery provisions required coordinated changes. The transferable lesson is to evaluate the complete configuration in its intended operating mode, including the arm and cabling. It does not establish that an independently designed four-wheel rover can reproduce Quince's access capability~\cite{LIT_QUINCE2011}.

\begin{figure}[htbp]
\centering
\includegraphics[width=6.300in,height=6.650in,keepaspectratio]{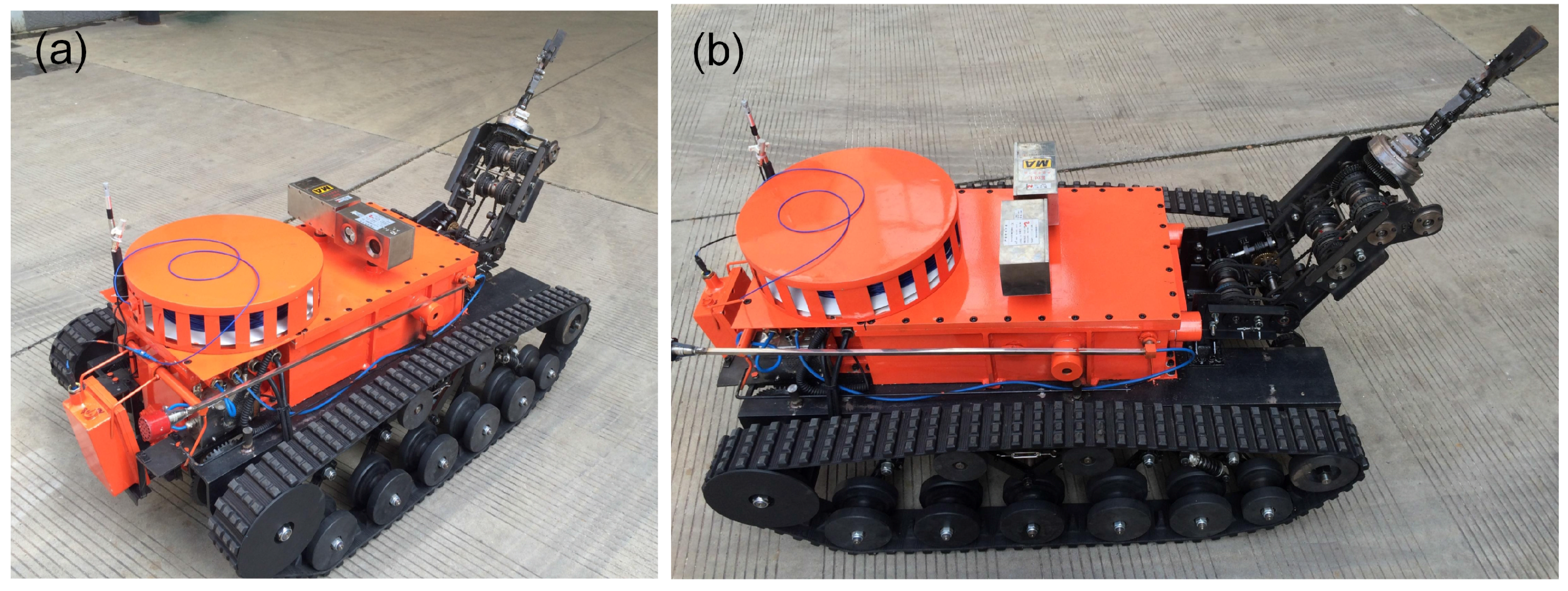}
\caption{Published coal-mine search-and-rescue robot in two views. Reproduced from~\cite{LIT_COAL2017} (source Fig. 3), \copyright{} 2017 the authors, under CC BY 4.0 (\url{https://creativecommons.org/licenses/by/4.0/}). No content changes.}
\label{fig:Coal_mine_rescue_robot}
\end{figure}

Zhao and colleagues' underground coal-mine system provides a complementary field perspective. Their trials and responder training revealed slipping, trajectory deviation, relay handling and cable-management difficulties alongside requests for better viewing arrangements and simpler operation. Such observations motivate tests that include environmental contamination, restricted sight lines and tether behaviour. Mechanical appearance and a nominal motor rating cannot substitute for those operational checks, and another platform does not inherit the original system's qualification~\cite{LIT_COAL2017}.

The German Rescue Robotics Center deployment report places training and organizational readiness alongside technical performance. Fire and flood missions required rehearsed teams and operation under disrupted infrastructure; available equipment could remain unused when appropriately trained operators were absent. For Zephyron, this supports defining setup, control handover, return, recovery and data-management procedures as part of the proposed system. The reviewed source is the accessible 2022 preprint of the later journal article, as identified in the citation record~\cite{LIT_DRZ2024}.

CERBERUS demonstrates how subterranean exploration at competition scale combines heterogeneous mobility, multimodal perception, communications support and supervised autonomy. Its integration lessons also distinguish capabilities actually deployed from research-level coordination methods. This makes CERBERUS a useful source of failure modes and evaluation questions, but an inappropriate direct performance baseline for a compact low-cost rover. The comparison therefore considers functions and evidence maturity, since platforms differ in resources and mission conditions~\cite{LIT_CERBERUS2023}.

Schwaiger and colleagues integrate mapping, radiation observation, sampling and valve manipulation using prioritized control modes. Their sampling sequence combines operator-triggered routines with a contamination-aware arm path, while valve manipulation remains teleoperated. Restricted access caused by robot width and outdoor mapping drift show the interaction between geometry, sensing and mission reach. This preprint supports explicit control authority, tool visibility and task-specific reach analysis; it does not imply that Zephyron's existing gripper can execute the same interventions~\cite{LIT_SCHWAIGER2024}.

\subsection{Chemical sensing, mapping and the meaning of a measurement}

Mobile gas measurement involves sensor dynamics as well as spatial navigation. Fan and colleagues use controlled gas sampling and baseline adaptation with a MOX electronic nose, demonstrating why warm-up, drift and exposure state must accompany the measurement stream. Their experiments employ comparatively safe chemical surrogates rather than validating arbitrary hazardous-gas mixtures. Accordingly, Zephyron's proposed gas interface should distinguish raw response, calibrated estimate, inferred class and uncertain detection instead of presenting each channel as a selective concentration instrument~\cite{LIT_GAS2019}.

\begin{figure}[htbp]
\centering
\includegraphics[width=6.300in,height=6.650in,keepaspectratio]{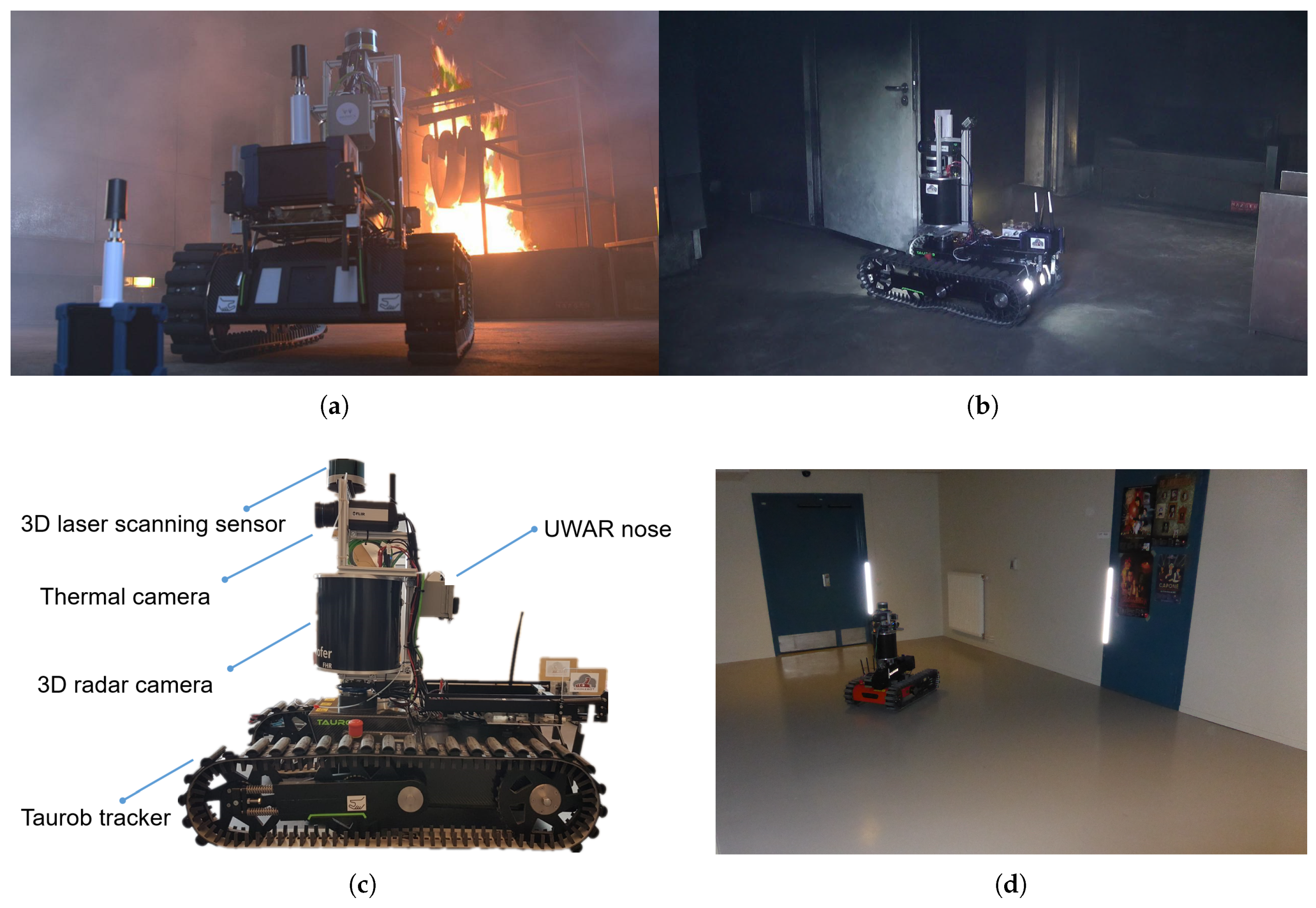}
\caption{SmokeBot operational scenarios recreated in a firefighter training facility, sensor setup, and experimental environment. Reproduced from~\cite{LIT_GAS2019} (source Fig. 8), \copyright{} 2019 the authors, under CC BY 4.0 (\url{https://creativecommons.org/licenses/by/4.0/}). No content changes.}
\label{fig:SmokeBot_gas_sensing_platform}
\end{figure}

Multi-compound gas mapping further separates chemical identity from concentration. Hernandez Bennetts and colleagues combine class probabilities with concentration observations to form compound-specific mean and variance maps. They also identify slow response, drift and limited quantitative ground truth as constraints. Their study motivates retaining uncertainty and classification confidence in the data model. It does not justify treating an interpolated coloured map as an independently verified distribution of a particular hazardous substance~\cite{LIT_MULTIGAS2014}.

Simulation can make these distinctions testable before field integration. GADEN combines gas dispersion and sensor-response models within a modular robotics framework, with wind-tunnel comparisons revealing weaker agreement in difficult low-flow conditions. Robot-induced airflow is not represented. A Zephyron simulation based on this literature should therefore declare the flow model, source assumptions and sensor response explicitly. A simulated plume is useful for controlled algorithm comparison, but cannot calibrate the physical sensor or demonstrate reliable chemical identification~\cite{LIT_GADEN2017}.

Information-driven gas distribution mapping links the sampling route to the expected reduction in uncertainty. Gongora and colleagues combine gas and wind estimates with planning, reporting both simulated studies and physical wind-tunnel demonstrations. Their published RMSE-versus-distance graph concerns simulation ground truth, as identified in the reproduced figure caption. The design implication is to assess what additional information a route obtains for its travel and sensing costs, while acknowledging computation and model assumptions~\cite{LIT_INFOMAP2023}.

\begin{figure}[htbp]
\centering
\includegraphics[width=6.300in,height=6.650in,keepaspectratio]{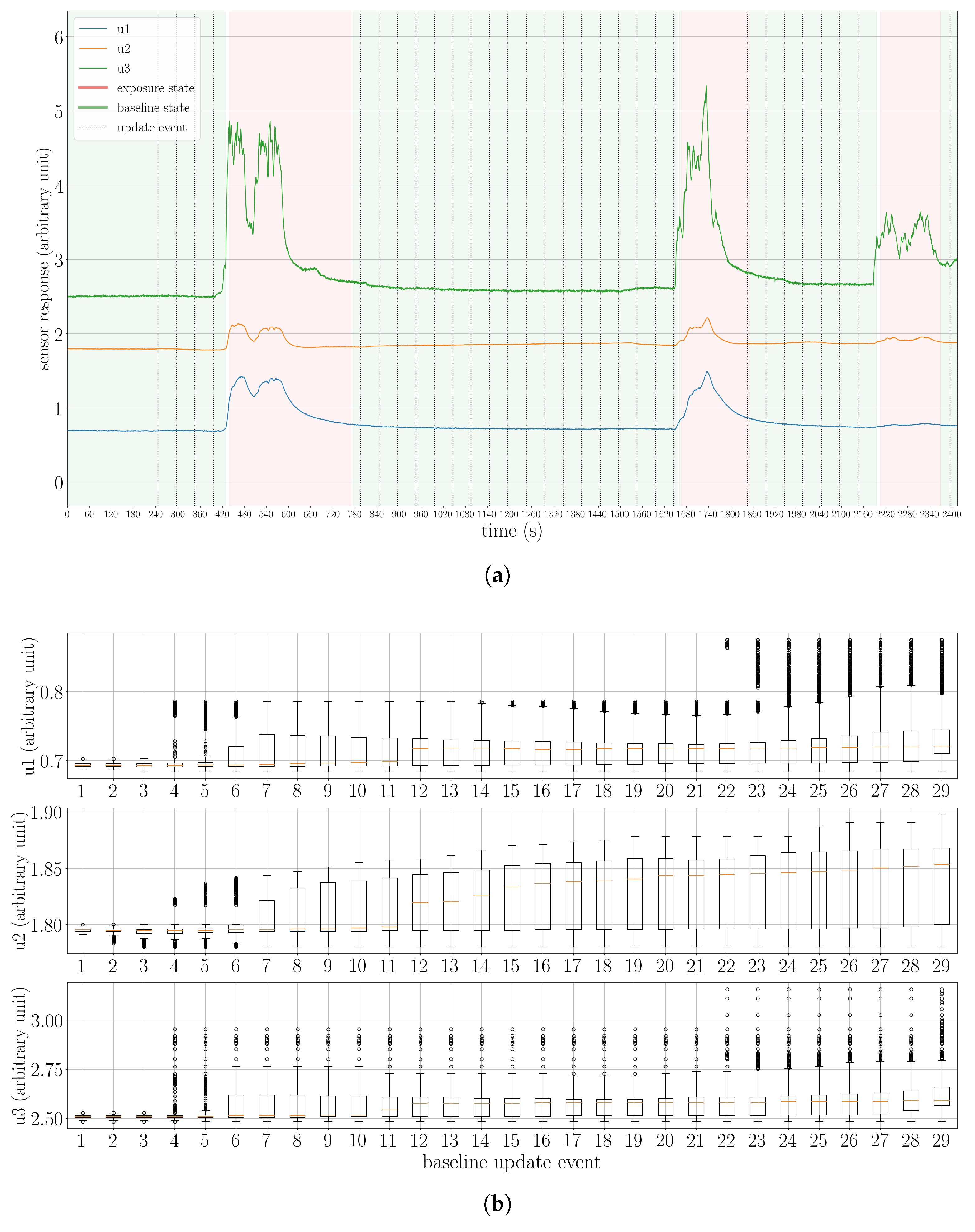}
\caption{Published MOX sensor responses, assigned states, and periodically learned baseline offsets from Experiment 2 of~\cite{LIT_GAS2019}. These are measurements on the authors' gas-sensing platform. Reproduced from source Fig. 9, \copyright{} 2019 the authors, under CC BY 4.0 (\url{https://creativecommons.org/licenses/by/4.0/}). No content changes.}
\label{fig:Published_MOX_drift}
\end{figure}

Semantic gas-source localization introduces another distinction: evidence that a substance is present is different from proof of its source. Monroy and colleagues use recognized objects and object--gas relationships to guide search. However, their laboratory substitutions and assumed source-confirmation step delimit the demonstrated capability. An object identified by a camera may prioritize an inspection target, but should remain a hypothesis until a specified confirmation procedure resolves the source. This is especially relevant when object recognition is proposed as a rover function~\cite{LIT_SEMANTICGAS2018}.

\begin{figure}[htbp]
\centering
\includegraphics[width=6.300in,height=6.650in,keepaspectratio]{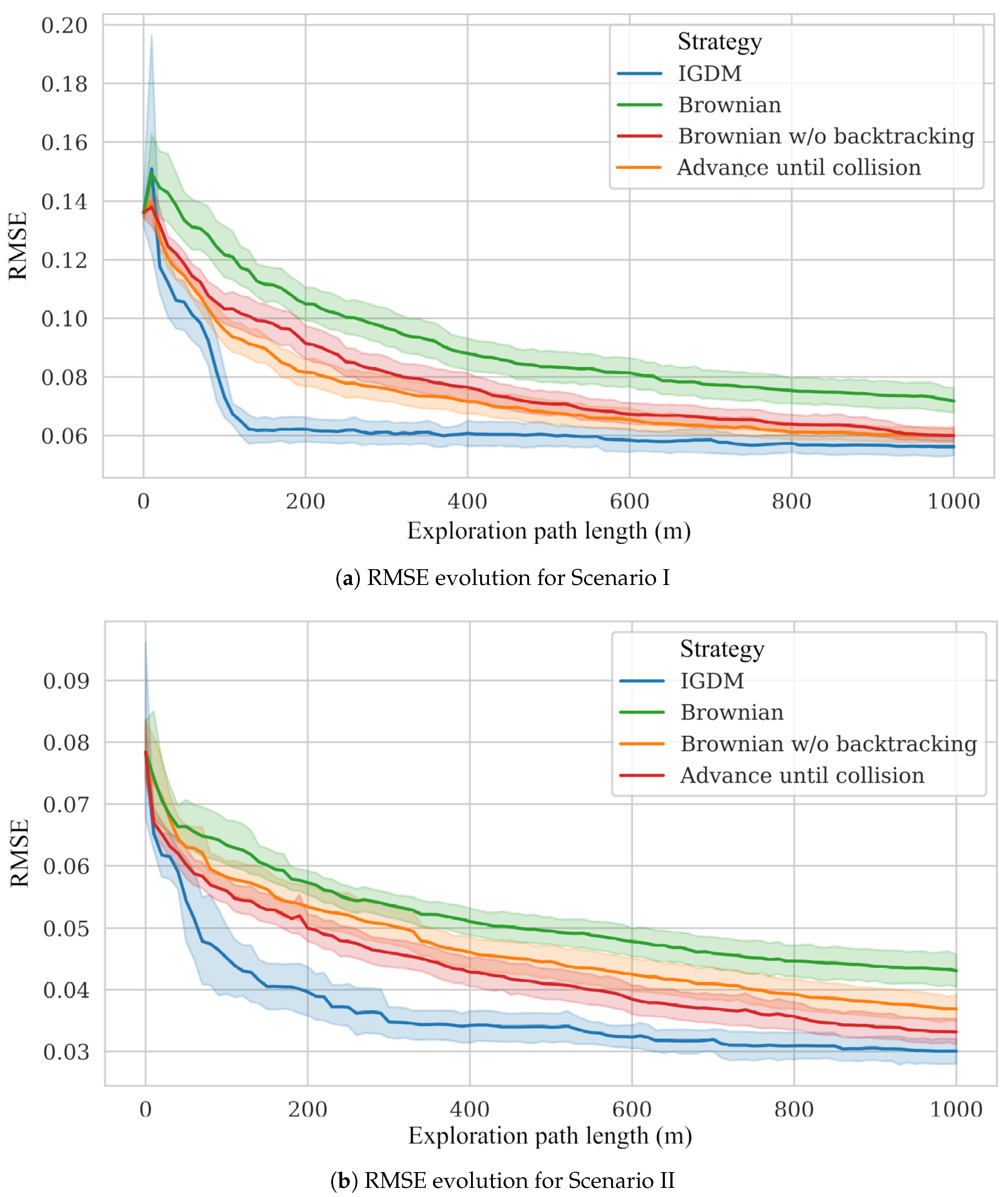}
\caption{Published gas-map RMSE relative to simulation ground truth as exploration path length increases~\cite{LIT_INFOMAP2023}. Each curve averages six starting positions; shading indicates one standard deviation. These are the source authors' simulations. Reproduced from source Fig. 10, \copyright{} 2023 the authors, under CC BY 4.0 (\url{https://creativecommons.org/licenses/by/4.0/}). No content changes.}
\label{fig:Published_gas_mapping_RMSE}
\end{figure}

Remote gas sensing provides a related but different planning model. The next-best-smell study balances viewing coverage, travel and acquisition time using a beam-based detector, with physical trials involving methane held in transparent containers. It expressly leaves quantitative source-estimation assessment outside its scope. The scheduling principle transfers to mobile sensing, whereas remote-beam visibility geometry does not transfer directly to a point MOX sensor mounted on a rover body. Sensor physics must determine the planning abstraction~\cite{LIT_NEXTSMELL2018}.

\subsection{Radiation observation and uncertainty-aware motion}

CARMA II distinguishes surface-contamination avoidance from gamma-dose-aware routing, demonstrating that these hazards require different sensing arrangements and behaviours. Its relation between detector look-ahead, update timing and travel speed is particularly relevant to preliminary layout reasoning. Yet its experiments use simulated sources. For a proposed rover, this literature motivates a conservative timing and stopping-distance calculation followed by appropriate validation, rather than a claim of established radiation detection performance or nuclear-facility readiness~\cite{LIT_CARMA2023}.

Groves and colleagues show why a radiation measurement should be associated with the robot pose at the relevant sensing time. Their navigation system accounts for detector delay when forming a layered costmap. The evaluated gamma response is emulated using radiofrequency source devices, and RF multipath can influence the resulting map. Thus the paper supports timing-aware navigation architecture and surrogate testing, while neither its plotted readings nor its test conditions establish calibrated ionizing-radiation performance for Zephyron~\cite{LIT_RADNAV2021}.

Ardiny and colleagues explicitly investigate how acquisition stops, field of view and angular resolution influence exploration. Simulated coverage is compared with physical marXbot operation, but the physical radiation sources are infrared range-and-bearing surrogates. Reproducing that result graph can explain why a sensor's observation model changes mission efficiency, with that distinction stated in the caption. It should not be described as a radiation-dose experiment or used to assign a numerical detection probability to another vehicle~\cite{LIT_RADHOT2019}.

\begin{figure}[htbp]
\centering
\includegraphics[width=6.650in,height=3.200in,keepaspectratio]{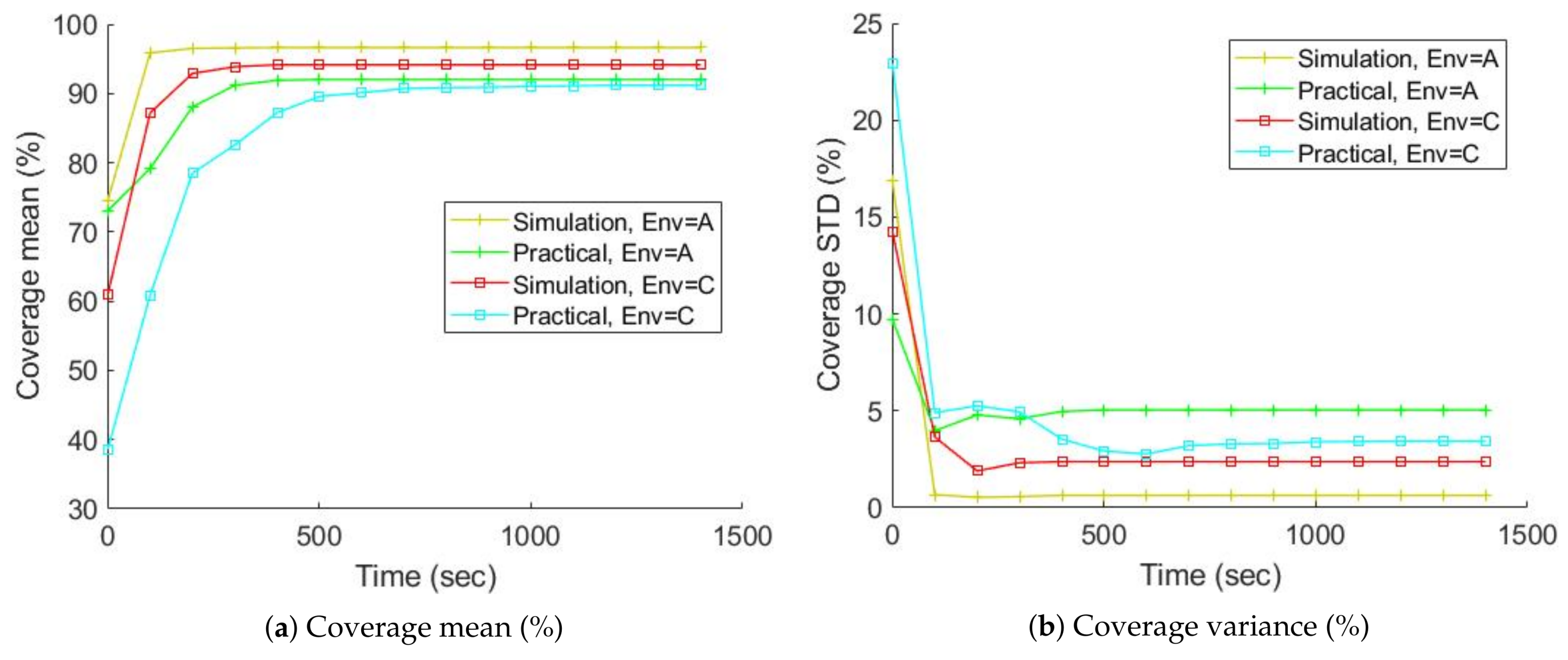}
\caption{Published simulated and physical-robot radiation-map coverage comparisons~\cite{LIT_RADHOT2019}. The left panel reports mean coverage over exploration time, averaged over ten trials; the right reports dispersion as labelled in the source. Physical trials used infrared range-and-bearing signals to emulate radiation. Reproduced from source Fig. 18(a,b), \copyright{} 2019 the authors, under CC BY 4.0 (\url{https://creativecommons.org/licenses/by/4.0/}). Panels arranged side by side; plotted data unchanged. These results concern another platform.}
\label{fig:Radiation_comparison}
\end{figure}

Probabilistic source search provides a further basis for separating observation from inference. Huo and colleagues combine Bayesian source estimation with information-based decisions, evaluating simulation and a barrier-free physical experiment. Obstacles and multiple-source extensions remain outside that demonstrated setting. A proposed Zephyron estimator should state its source, geometry and count-statistical assumptions, and its evaluation should independently test environments that violate them rather than extrapolating open-space localization behaviour to rubble or shielded structures~\cite{LIT_RADSEARCH2020}.

Adaptive sampling on a manipulator-equipped inspection platform also exposes practical constraints: target poses can be unreachable, contact triggers can be spurious, and repeated failed motions can terminate a mission. The limited test series reported by Adams and colleagues motivates recording feasibility rejection, recovery and termination events alongside successful samples. It does not provide a general reliability estimate. For Zephyron, these events belong in a proposed evaluation protocol before autonomous intervention is claimed~\cite{LIT_RADADAPT2025}.

\subsection{Water observation and sample traceability}

The water-monitoring USV developed by Chang and colleagues combines situated sensing, bottle collection and a sequence that flushes residual liquid before sampling. Those workflow ideas are relevant to a ground rover that reaches an accessible water edge or container. Its buoyancy, propulsion and navigation results do not transfer to wheeled mobility. Likewise, a pH observation or simple threshold classification does not establish water potability; the proposed workflow specifies the parameter and the need for subsequent reference analysis~\cite{LIT_WATER2021}.

Ryu's open-source water platform connects multiparameter sensing with local storage and cellular reporting. A particularly useful limitation is that positions had to be joined to cloud sensor records using external GPS timestamps during post-processing. This supports acquiring time, location, sensor status and sample identity together in the proposed Zephyron record. Attractive spatial interpolation can assist visualization, but remains dependent on sampling coverage, alignment and calibration rather than constituting independent field validation~\cite{LIT_IDRONE2022}.

\subsection{Reproducible hardware and achievable autonomy}

ROMR demonstrates the value of publishing build information, hardware interfaces and software alongside an accessible mobile platform. Its payload characterization is explicitly associated with flat-ground conditions. The transferable contribution is reproducibility and declared test context, not a payload number for a visually different rover. Zephyron should similarly distinguish selected components, measured as-built properties and analytically estimated requirements when a physical prototype becomes available for characterization~\cite{LIT_ROMR2023}.

SMARTmBOT provides a small ROS2 platform with exposed camera and sensor interfaces and repeatability-oriented demonstrations. Its evaluated control tasks use external motion-capture positioning, and the accessible document is a preprint. It therefore supports modular interface design and controlled laboratory evaluation, while offering no basis to equate an embedded camera and wireless link with standalone outdoor localization. External infrastructure should be stated whenever it is necessary for a demonstration's success~\cite{LIT_SMART2022}.

Cordie and colleagues examine continued mobility after steering failures through configuration changes and module removal, combining simulation with physical trials. Physical module removal is manual, which must not be relabelled autonomous self-repair. For a rover whose main structure is to remain fixed, the useful implications concern service access, replaceable assemblies and well-defined degraded operation. Those choices should be evaluated against added mass, interfaces and energy demands rather than assumed to improve every mission~\cite{LIT_MODULAR2024}.

Navigation algorithms also impose specific hardware requirements. LIO-SAM evaluates LiDAR--inertial estimation with different datasets and loop-closure or GNSS configurations; it presupposes suitable sensors, calibration and computation. ORB-SLAM3 evaluates visual and visual--inertial configurations using established datasets and carefully defined trajectory-error alignment. Together they support a requirement-led localization choice. Neither paper permits a benchmark accuracy value to be assigned to an untested Zephyron sensor package or to a rendered camera housing~\cite{LIT_LIOSAM2020},~\cite{LIT_ORBSLAM2021}.

The Marathon 2 navigation study makes operational endurance and recovery observable through repeated waypoint routes, trajectory logs and event recording. Its lessons support reporting recoveries separately from human interventions: an autonomous recovery may be successful system behaviour, whereas a silently assisted run obscures capability. This comparison concerns the implementation tested in that historical study; it does not characterize all subsequent versions of the navigation framework~\cite{LIT_MARATHON2020}.

ResQbot shows how intervention evaluation can use instrumented surrogates and controlled viewing conditions. Its casualty-loading study uses a dummy, so it is not clinical evidence for handling people. For Zephyron's small arm, the appropriate transfer is methodological: define a representative object, approach geometry, contact constraints and repeatable task, then measure outcomes. The presence of a gripper does not itself demonstrate rescue extraction, chemical sampling or force-controlled manipulation~\cite{LIT_RESQBOT2019}.

\subsection{Evidence maturity across representative platforms}

The following comparison groups systems by the role relevant to this design. It avoids ranking robots by headline accuracy or payload values obtained under incompatible conditions.

\begingroup
\small
\setlength{\tabcolsep}{4pt}
\renewcommand{\arraystretch}{1.14}
\begin{longtable}{@{}>{\raggedright\arraybackslash}p{0.353\linewidth}>{\raggedright\arraybackslash}p{0.316\linewidth}>{\raggedright\arraybackslash}p{0.294\linewidth}@{}}
\caption{Representative platforms and the limits of transfer}\label{tab:1}\\
\toprule
\rowcolor{tablehead} \textbf{Platform or study} & \textbf{Relevant contribution} & \textbf{Transfer boundary for Zephyron} \\
\midrule
\endfirsthead
\toprule
\rowcolor{tablehead} \textbf{Platform or study} & \textbf{Relevant contribution} & \textbf{Transfer boundary for Zephyron} \\
\midrule
\endhead
\bottomrule
\endfoot
Quince & Mission-driven mechanical and operational redesign~\cite{LIT_QUINCE2011} & Different tracked platform and deployment conditions \\[2pt]
Coal-mine search robot & Environment sensing and operational field issues~\cite{LIT_COAL2017} & Other hardware and qualification context \\[2pt]
SmokeBot sensing study & MOX dynamics and baseline adaptation~\cite{LIT_GAS2019} & Different sensing system and controlled chemicals \\[2pt]
Information-driven gas mapping & Sampling-route and map-quality relationship~\cite{LIT_INFOMAP2023} & Simulation ground truth differs from field truth \\[2pt]
Radiation-aware navigation & Pose and measurement-timing association~\cite{LIT_RADNAV2021} & Evaluation used radiofrequency source surrogates \\[2pt]
ROMR & Reproducible hardware and declared test conditions~\cite{LIT_ROMR2023} & Flat-ground characterization does not transfer as a rating \\[2pt]
Zephyron in this study & Selected geometry, component sizing, and mission models & Design and analysis; no new physical performance tests \\[2pt]
\end{longtable}
\endgroup

\FloatBarrier
\section{System integration and mechanical arrangement}

\subsection{Structure and service access}

The revised body retains the central faceted cowl and carbon-sheet visual treatment. An aluminum ladder structure carries wheel modules, payload mounts, the battery compartment, and the two raked solar supports. The body panels provide enclosure and access surfaces; no composite laminate strength is inferred from their carbon appearance. If structural carbon-fiber laminates are selected during manufacture, ply sequence, resin system, fiber orientation, insert design, and coupon properties must be specified independently of the render material.

The wheel modules remain at the four corners. Front and rear mounting blocks connect through the side rails and outer longitudinal members, preserving the photographed stance. Corner joints use visible fasteners and support plates in the visualization to communicate load paths and maintainable assembly. A closed underside tray protects the proposed low-mounted battery and power distribution region, while side access panels allow inspection of wiring and electronics. Cable loops are retained near moving joints and restrained along fixed members so that articulation does not require a taut harness.

The solar panel remains at the elevated rear position, with its original support concept. Replacing the earlier unspecified panel with a selected 395 by 305 mm module affects rear overhang, wind loading, and center of mass. A support hinge height near 720 mm is retained as a packaging selection. This location makes stationary collection practical, but it also raises the system center of mass and exposes the panel to branch or doorway contact. The panel mounting structure must consequently be included in stability and clearance tests; solar benefit cannot be considered separately from mechanical cost.

\begin{figure}[htbp]
\centering
\includegraphics[width=0.467\linewidth]{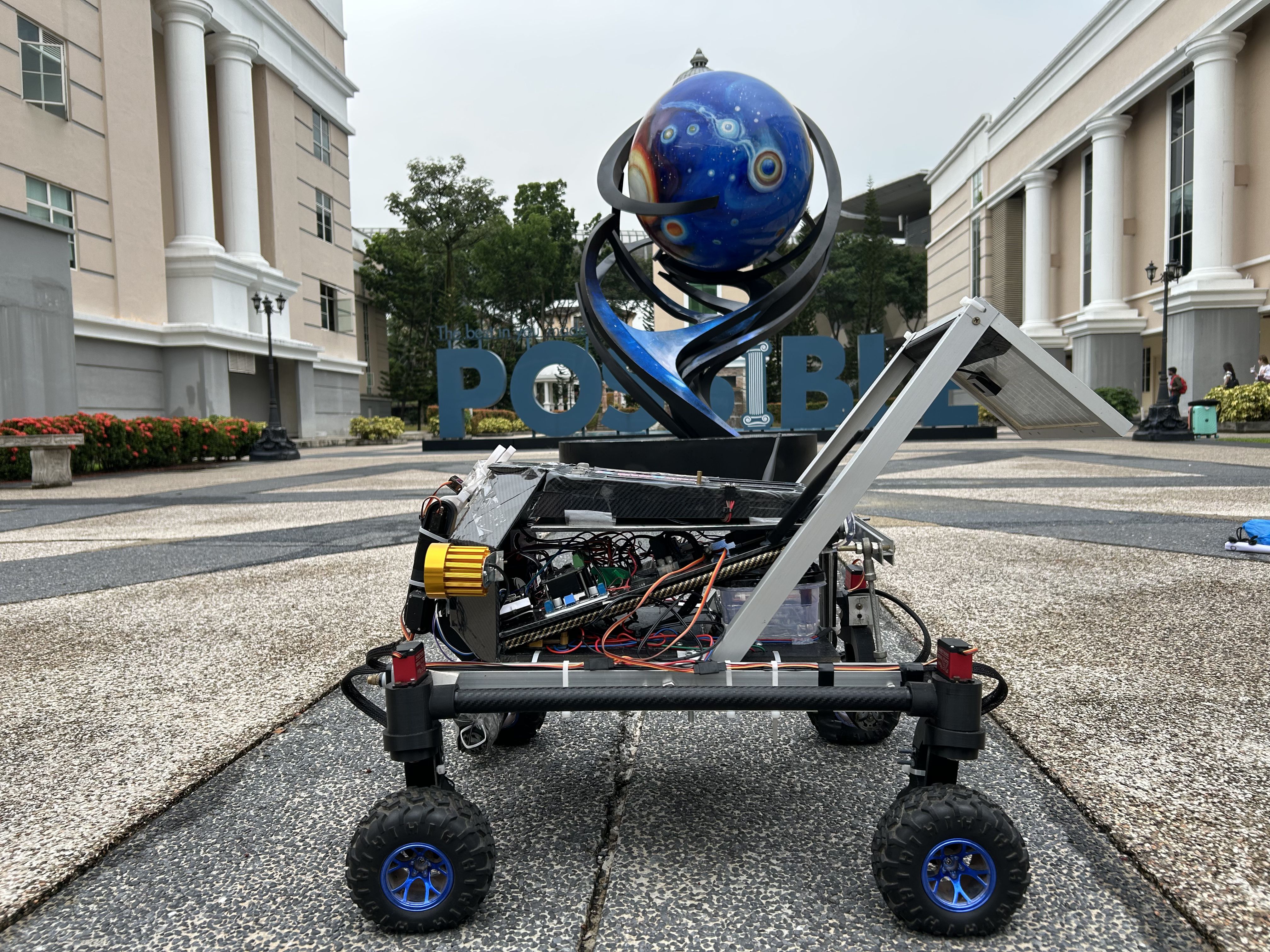}
\hfill
\includegraphics[width=0.467\linewidth]{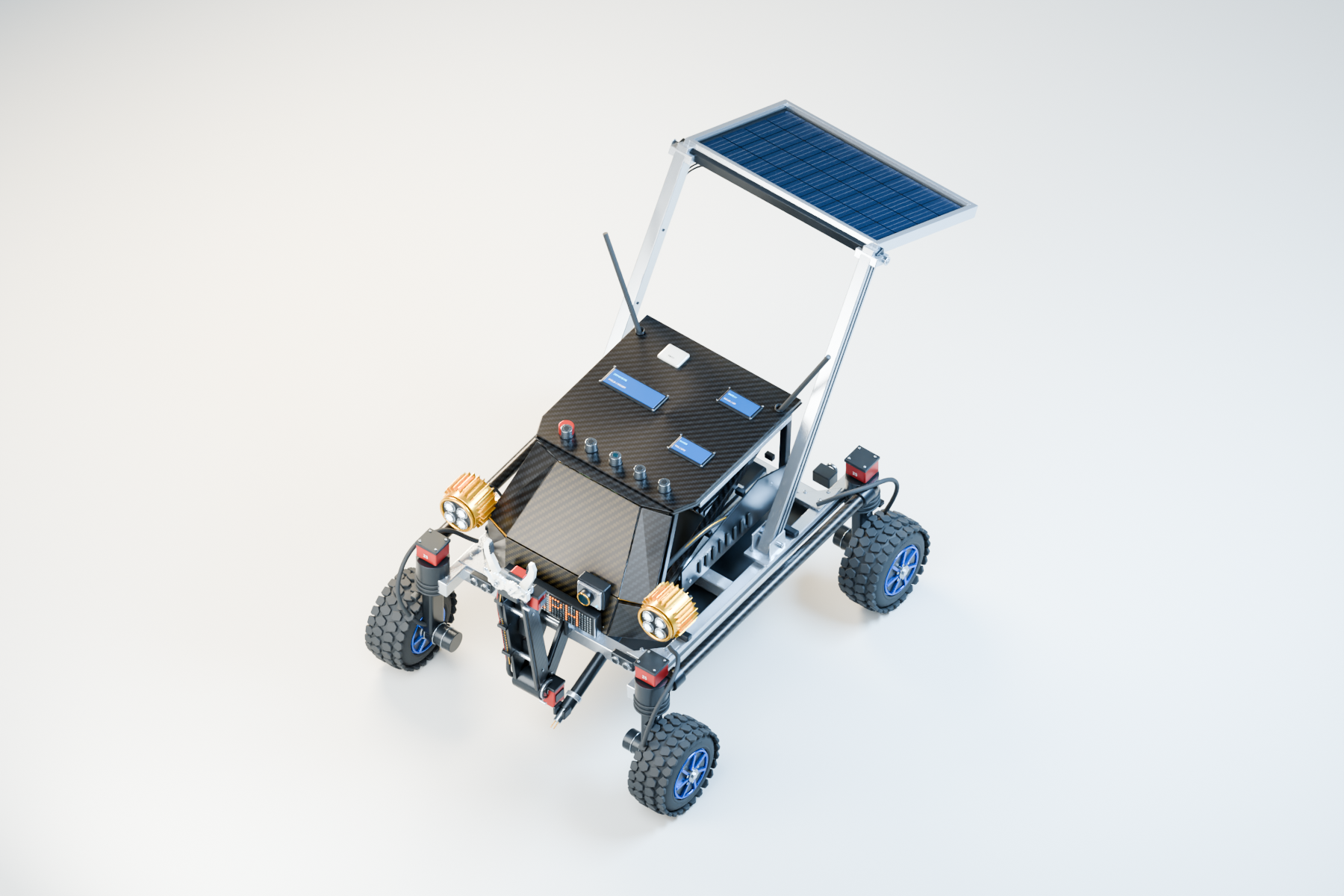}
\caption{Physical configuration and proposed engineering design. Left: author-supplied outdoor side photograph showing the chassis and solar support. Right: Blender visualization preserving the four-wheel stance, carbon-patterned body, front manipulator, lamps and elevated rear panel. Battery protection, enclosure details and component packaging in the rendering are design proposals. The rendering is not a photograph or an experimentally validated digital twin.}
\label{fig:MODEL}
\end{figure}

The front manipulator remains centered on the nose. Its folded pose protects the forward reach envelope during travel, while a deployed pose supports controlled retrieval demonstrations. Front work lamps remain on either side of the cowl, and the camera retains a forward-facing nose position. These fixed placements impose task constraints. For example, a fixed underslung water probe can contact a raised sample well without moving the entire rover into standing water, but it does not automatically provide arbitrary-depth sampling. A future probe deployment mechanism would require a separate design and validation step.

\subsection{Coordinating traction and steering}

Four-wheel propulsion and wheel-angle steering are treated as distinct subsystems. The visible steering servos do not justify assuming that the original control software coordinated all knuckles. For the proposed chassis, define a body frame with x forward and y left, wheel-center coordinates (\(x_i\),\(y_i\)), desired body velocity (\(v_x\),\(v_y\)), and yaw rate \(\omega\). The no-slip instantaneous velocity at wheel i is

\begin{equation}
\begin{gathered}\mathbf{v}_i=\begin{bmatrix}v_x-\omega y_i\\v_y+\omega x_i\end{bmatrix},\\
\delta_i=\operatorname{atan2}(v_y+\omega x_i,v_x-\omega y_i),\\
\Omega_i=\frac{\|\mathbf{v}_i\|}{r}\end{gathered}
\label{eq:1}
\end{equation}

This kinematic relation provides a command target, not a guarantee of rolling without slip. Steering limits, actuator slew rate, wheel speed saturation, and measured geometry constrain its use. In a symmetric turn with zero commanded lateral velocity, front and rear wheels on one side have equal speed magnitudes and opposite steering angles; two grouped traction channels can therefore be compatible with that restricted maneuver. General lateral motion requires revisiting the available independent drive channels. A controller must explicitly document its supported modes rather than mixing skid-steer equations with angled wheels without explanation.

For initial experiments, the operator should select a defined steering mode, and the controller should limit wheel motion while steering transitions occur. Encoder signs, left/right motor polarity, and wheel-angle zero offsets are calibrated on a fixture before ground trials. A stale command or a disagreement between commanded and measured motion triggers a controlled stop. Emergency stopping must remain independent of the remote inference computer and dashboard. This separation prevents loss of a high-level application from removing the ability to stop the rover.

\begin{figure}[htbp]
\centering
\includegraphics[width=6.800in,height=3.850in,keepaspectratio]{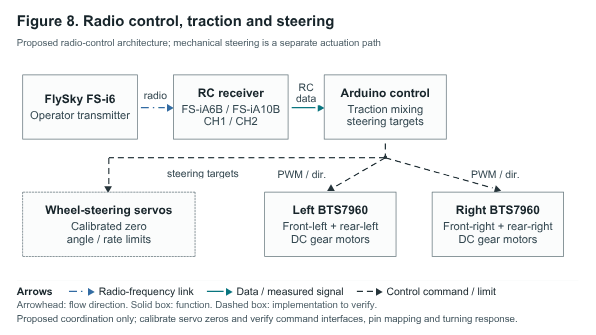}
\caption{Radio control traction and steering. Proposed architecture with arrow conventions defined in the legend. Native editable Canva design, rendered by Canva.}
\label{fig:CANVA_F02}
\end{figure}

\subsection{Electrical and data interfaces}

The proposed power system separates the solar charging path, protected battery pack, motor supply, regulated logic supply, and switched mission loads. The selected LiFePO4 pack requires a chemistry-matched charging profile. Its internal protection and balancing functions are not replaced by a fuse or a generic charger. A main disconnect and appropriately rated branch protection support maintenance and fault isolation. The rendered power tray illustrates packaging only; final conductor sizing, connector rating, converter efficiency, inrush behavior, and thermal margins require a completed circuit design.

Motor and servo return currents can disturb low-level analog measurements if grounds and supply paths are poorly arranged. The acquisition design should therefore document reference voltage, analog filtering, sampling timing, and the relationship between power ground and signal ground. Water and gas channels require separate calibration metadata even when they share one analog-to-digital converter. Radiation pulses require a defined counting interval and handling of counter overflow. Position and vision records require a consistent timebase so that data can be compared across devices.

Each logged record should include a monotonic timestamp, wall-clock timestamp when available, sequence number, sensor identifier, raw signal, converted estimate, unit, calibration identifier, quality flag, and rover operating state. Position records should retain fix validity and the antenna-to-sensor offset. Recording raw values as well as converted outputs makes later recalibration possible without pretending that a new fit was used during the original run. The same principle applies to camera records: model identity and frame timestamps are retained separately from detection outputs.

\begin{figure}[htbp]
\centering
\includegraphics[width=6.800in,height=3.850in,keepaspectratio]{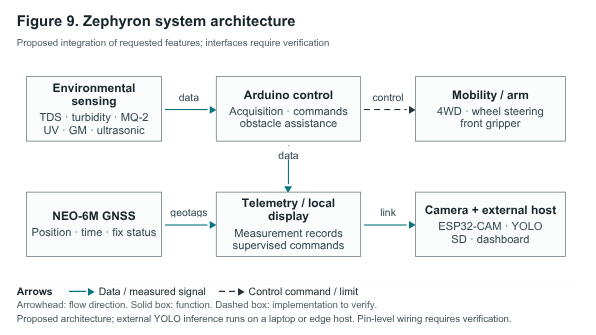}
\caption{System architecture. Proposed architecture with arrow conventions defined in the legend. Native editable Canva design, rendered by Canva.}
\label{fig:CANVA_F01}
\end{figure}

\FloatBarrier
\section{Engineering models and calculated results}

\subsection{Design scope and a reproducible baseline}

The revised Zephyron concept combines supervised ground mobility, a front manipulator, environmental sensing, camera acquisition, remote inference, and solar-assisted battery operation. The physical arrangement preserves the four-wheel stance, carbon-patterned central enclosure, forward gripper, and elevated rear panel evident in the reference photographs. The model is a design visualization rather than an as-built metrology record. Photographs support relative arrangement, but perspective, lens distortion, hidden mounting surfaces, and the absence of a scale reference prevent exact dimensional recovery. Accordingly, the geometry below is selected for a coherent new build and must be checked against purchased components before manufacturing.

The wheel diameter is selected as 165 mm with a 75 mm tread envelope. Wheel-center track is 538 mm and wheelbase is 570 mm, giving a nominal wheel envelope of 735 mm longitudinally and 613 mm transversely. The selected chassis envelope is 600 by 390 mm, cowl top approximately 450 mm above ground, and rear solar support hinge approximately 720 mm above ground. A 130 mm central-chassis clearance is a minimum packaging requirement, not a uniform minimum across all underbody parts. The evaluated Blender geometry places the lowest central chassis component approximately 191.1 mm above the studio ground; the folded arm and fixed probe have approximately 96.1 mm and 141.3 mm clearance, respectively. These are model geometry values, not measurements of the photographed rover. This clearance does not equal obstacle-climbing height: approach geometry, tire compliance, wheel torque, center of mass, and terrain determine traversability.

The mass budget is 10 kg dry plus 2 kg of chassis-carried mission payload. These are allocation targets, not measured mass or demonstrated carrying capacity. The manipulator's grasped-object capability is a separate constraint. A grasped object and its adapters count toward the same 12 kg gross budget and reduce the remaining chassis-carried allocation unless the analyses are repeated at a higher gross mass. The battery should occupy a low, central compartment; the elevated panel and arm should be included in the center-of-mass calculation for every operating pose. Components mounted above the chassis increase overturning moment even if overall mass remains within budget. A later bill of materials must replace allowance masses with weighed parts and include fasteners, wiring, enclosures, guards, sample containers, and manufacturing tolerances.

The motor selection is four Pololu 4755 gearmotors. Manufacturer specifications identify a 102.08:1 gearbox, encoder, 100 rpm free-running speed at 12 V, approximately 0.2 A free-running current, 210 g mass, and 37 by 72.5 mm body. Extrapolated stall values are 34 kgf\ensuremath{\cdot}cm (3.334 N\ensuremath{\cdot}m) and 5.5 A. More relevant design restrictions are the manufacturer's recommended continuous gearbox load of 10 kgf\ensuremath{\cdot}cm (0.981 N\ensuremath{\cdot}m) and general brushed-motor guidance to operate at no more than 25\% of stall current~\cite{ENG_MOTOR}. The manufacturer values characterize components under specified conditions; they do not establish the performance of a complete rover.

\begingroup
\small
\setlength{\tabcolsep}{4pt}
\renewcommand{\arraystretch}{1.14}
\begin{longtable}{@{}>{\raggedright\arraybackslash}p{0.353\linewidth}>{\raggedright\arraybackslash}p{0.316\linewidth}>{\raggedright\arraybackslash}p{0.294\linewidth}@{}}
\caption{Selected geometry and component baseline}\label{tab:2}\\
\toprule
\rowcolor{tablehead} \textbf{Quantity} & \textbf{Selected value} & \textbf{Evidence status} \\
\midrule
\endfirsthead
\toprule
\rowcolor{tablehead} \textbf{Quantity} & \textbf{Selected value} & \textbf{Evidence status} \\
\midrule
\endhead
\bottomrule
\endfoot
Wheel diameter \ensuremath{\times} tread envelope & 165 \ensuremath{\times} 75 mm & New geometric selection \\[2pt]
Track / wheelbase & 538 / 570 mm & New geometric selection \\[2pt]
Dry / chassis payload / gross mass & 10 / 2 / 12 kg & Budget targets \\[2pt]
Commanded translational speed cap & 0.35 m\ensuremath{\cdot}s\textsuperscript{-}\textsuperscript{1} & Proposed operating limit \\[2pt]
Gearmotor & Four 12 V, 100 rpm encoder units & Component selection \\[2pt]
Battery & Bioenno BLF-1206A, 12 V label, 6 Ah & Component selection \\[2pt]
Solar module & Newpowa NPA20S-12J, 20 W & Component selection \\[2pt]
Manipulator links & 0.146 and 0.264 m & Selected model geometry \\[2pt]
\end{longtable}
\endgroup

\subsection{Mass allocation and geometric stability}

The 10 kg dry-mass target is an allocation constraint. Documented component masses consume part of that target before structure, wheel assemblies, wiring, enclosure, sensors, and the arm are finalized. The remaining allocation is not a measured mass and should not be distributed among unselected parts merely to make a finished-looking bill of materials.

\begingroup
\small
\setlength{\tabcolsep}{4pt}
\renewcommand{\arraystretch}{1.14}
\begin{longtable}{@{}>{\raggedright\arraybackslash}p{0.353\linewidth}>{\raggedright\arraybackslash}p{0.316\linewidth}>{\raggedright\arraybackslash}p{0.294\linewidth}@{}}
\caption{Dry mass allocation and mission payload allowance}\label{tab:3}\\
\toprule
\rowcolor{tablehead} \textbf{Mass allocation} & \textbf{Value} & \textbf{Basis} \\
\midrule
\endfirsthead
\toprule
\rowcolor{tablehead} \textbf{Mass allocation} & \textbf{Value} & \textbf{Basis} \\
\midrule
\endhead
\bottomrule
\endfoot
Four selected drive motors & 0.840 kg & Four times 0.210 kg manufacturer value~\cite{ENG_MOTOR} \\[2pt]
Selected battery pack & 0.700 kg & Manufacturer value~\cite{ENG_BATTERY} \\[2pt]
Selected solar module & Approximately 1.433 kg & Conversion of documented 3.16 lb~\cite{ENG_PANEL} \\[2pt]
Subtotal of these components & Approximately 2.973 kg & Calculated allocation \\[2pt]
Remaining dry-mass allowance & Approximately 7.027 kg & 10 kg target minus subtotal \\[2pt]
Chassis-carried payload target & 2.000 kg & Gross-budget allowance shared with grasped objects; arm capability assessed separately \\[2pt]
\end{longtable}
\endgroup

For component masses \(m_j\) with center-of-mass coordinates \(z_j\), the assembled center-of-mass height is

\begin{equation}
h_{CG}=\frac{\sum_j m_jz_j}{\sum_jm_j}
\label{eq:2}
\end{equation}

Arm motion changes the horizontal center of mass as well as its height. The raised solar panel contributes a moment that must be included even when the rover is stationary. For a rigid vehicle on a planar lateral slope, with centered mass and support width b, the ideal geometric tipping angle is

\begin{equation}
\theta_{tip,ideal}=\tan^{-1}\left(\frac{b}{2h_{CG}}\right)
\label{eq:3}
\end{equation}

This expression describes loss of static support in a simplified geometry. Tire deformation, suspension movement, edge contact, dynamic acceleration, soil failure, and uncertainty in mass position can reduce the available margin. It must not be used as an operational slope rating. With the selected track of 0.538 m and illustrative center-of-mass heights of 0.20, 0.26, and 0.32 m, the ideal angles are approximately 53.4, 46.0, and 40.1 degrees. These heights are sensitivity assumptions, not values inferred from photographs. The independently proposed 10 degree grade test remains a controlled validation condition.

\begin{figure}[htbp]
\centering
\includegraphics[width=6.800in,height=3.850in,keepaspectratio]{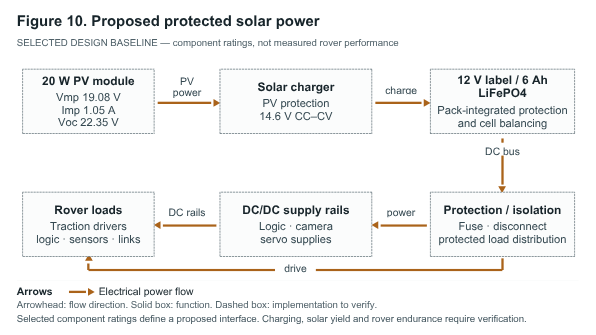}
\caption{Solar charging and protected power. Proposed architecture with arrow conventions defined in the legend. Native editable Canva design, rendered by Canva. The drive branch requires the proposed regulated 12 V motor rail; this functional diagram omits its detailed converter stage.}
\label{fig:CANVA_F05}
\end{figure}

\subsection{Mobility, traction, and geometric uncertainty}

For wheel diameter D in metres and rotational speed n in revolutions per minute, the no-slip peripheral speed is

\begin{equation}
v_0=\pi Dn/60
\label{eq:4}
\end{equation}

Using D = 0.165 m and n = 100 rpm gives v\ensuremath{_{0}} = 0.864 m\ensuremath{\cdot}s\textsuperscript{-}\textsuperscript{1}. This is a no-load kinematic ceiling, not an achieved ground speed. The selected 0.35 m\ensuremath{\cdot}s\textsuperscript{-}\textsuperscript{1} command cap corresponds to approximately 40.5 wheel rpm. Wheel encoders permit speed feedback but do not directly measure ground speed when tires slip. A controller should therefore limit acceleration and reject stale commands independently of any camera inference result.

For the straight-knuckle, differential-drive operating approximation, let r be wheel radius in metres, \ensuremath{\omega}R and \ensuremath{\omega}L right and left equivalent wheel angular speeds in rad\ensuremath{\cdot}s\textsuperscript{-}\textsuperscript{1}, and b the effective track in metres. Then

\begin{equation}
v=\frac{r}{2}(\omega_R+\omega_L),\qquad \dot\psi=\frac{r}{b}(\omega_R-\omega_L)
\label{eq:5}
\end{equation}

These equations are a reduced model. Four-wheel skid steering introduces lateral slip during turning, making the effective track surface dependent. If individual steering knuckles are actively angled, a steering-geometry model must replace this approximation. Skid-steer literature motivates identifying instantaneous centers of rotation or equivalent terrain-specific parameters rather than assuming ideal rolling during every turn~\cite{ENG_SKID}. Unequal rolling radii and effective wheelbase error also create systematic odometry bias~\cite{ENG_ODOM}.

For an illustrative equal-angular-speed command, set the right rolling radius to (1 + \ensuremath{\epsilon}) times the left radius. The resulting curvature is approximately \ensuremath{\kappa} = \ensuremath{\epsilon}/[b(1 + \ensuremath{\epsilon}/2)], with \ensuremath{\kappa} in m\textsuperscript{-}\textsuperscript{1} and dimensionless mismatch \ensuremath{\epsilon}. Over centerline path length s, heading error is \ensuremath{\kappa}s, forward displacement is sin(\ensuremath{\kappa}s)/\ensuremath{\kappa}, and lateral deviation is [1 \ensuremath{-} cos(\ensuremath{\kappa}s)]/\ensuremath{\kappa}. At b = 0.538 m, a 1\% mismatch produces approximately 10.6\ensuremath{{}^{\circ}} heading change and 0.92 m lateral deviation after a 10 m path in this idealized calculation. Tire wear, inflation, payload compression, and surface slip can change effective radius. The calculation demonstrates sensitivity, not a predicted field trajectory.

A first traction estimate for steady ascent is

\begin{equation}
F_{req}=mg(\sin\theta+C_{rr}\cos\theta),\quad \tau_w=F_{req}r/4
\label{eq:6}
\end{equation}

Here m is gross mass in kg, g = 9.81 m\ensuremath{\cdot}s\textsuperscript{-}\textsuperscript{2}, \ensuremath{\theta} grade angle, Crr a dimensionless rolling-resistance coefficient, and \ensuremath{\tau}w torque per driven wheel in N\ensuremath{\cdot}m under equal sharing. With the selected 12 kg mass, 10\ensuremath{{}^{\circ}} grade scenario, Crr = 0.04 assumption, and r = 0.0825 m, the required force is approximately 25.08 N and wheel torque 0.517 N\ensuremath{\cdot}m. The corresponding minimum friction coefficient in this simplified steady model is tan \ensuremath{\theta} + Crr \ensuremath{\approx} 0.216. Neither the friction coefficient nor rolling resistance has been measured for the intended terrain.

\begin{figure}[htbp]
\centering
\includegraphics[width=6.300in,height=4.800in,keepaspectratio]{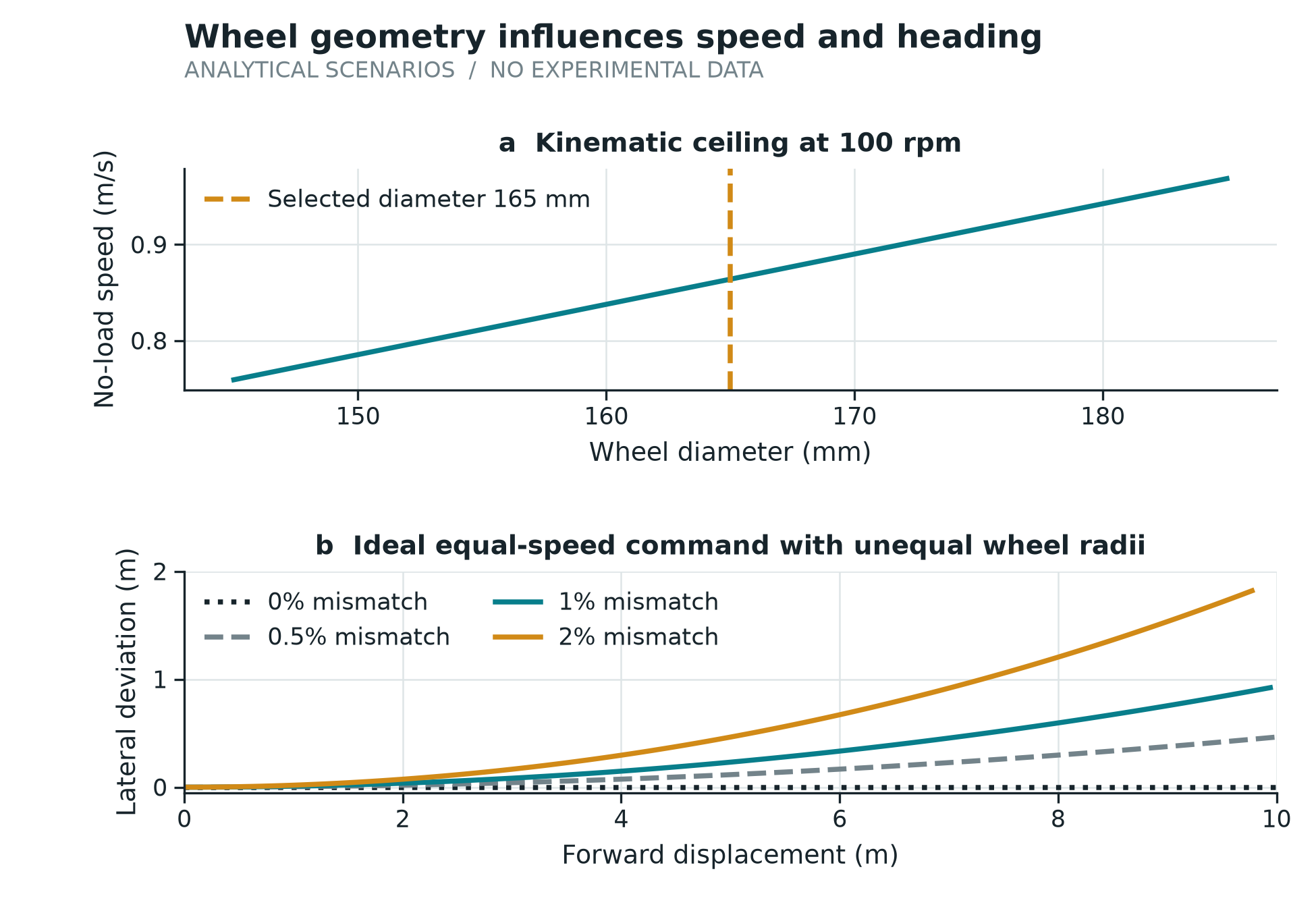}
\caption{Wheel diameter and systematic steering sensitivity. Analytical scenario; no experimental data. The top panel uses peripheral speed at 100 rpm, not loaded ground speed. The bottom panel assumes equal angular speeds, 538 mm effective track and 0--2\% right-to-left rolling-radius mismatch over a 10 m centerline path. Skid turning and active steering require additional models. Wheel-diameter variation is a sensitivity study, not photographic measurement~\cite{ENG_MOTOR},~\cite{ENG_SKID},~\cite{ENG_ODOM}.}
\label{fig:ENG_F04}
\end{figure}

Equal wheel sharing is most plausible on a level, sufficiently stiff support surface with a centered load. On uneven terrain, the effective traction contribution can become concentrated at fewer contacts. For normalized traction shares \(\alpha_i\) that sum to one, the corresponding wheel requirement is

\begin{equation}
\tau_i=\alpha_i F_{req}r,\qquad \sum_i\alpha_i=1
\label{eq:7}
\end{equation}

As a sensitivity case, two effective driven contacts carrying the same total ascent demand each require approximately 1.035 N\ensuremath{\cdot}m, twice the equal-four-contact result. That value exceeds the selected motor's 0.981 N\ensuremath{\cdot}m continuous gearbox recommendation. This is not a simulated suspension result: it is an allocation bound showing that contact distribution can remove the apparent sizing margin. A future traction test should measure wheel contact and load transfer while varying payload placement and obstacle geometry. A nominal four-wheel drive architecture cannot be assumed to maintain four equally useful contacts throughout a rubble maneuver.

The torque-sizing model and the 35 W drive-energy scenario also answer different questions. The former estimates a required mechanical load at a declared grade, while the latter is a selected battery-side increment for mission planning. Multiplying an interpolated winding current by 12 V would not establish battery power at the capped wheel speed without a PWM, motor-loss, and converter model. These analyses are therefore not combined into a claim of runtime on a 10 degree slope. The physical validation should record voltage, current, wheel speed, and grade together so that a coupled energy model can replace the separate preliminary scenarios.

\begin{figure}[htbp]
\centering
\includegraphics[width=6.300in,height=4.800in,keepaspectratio]{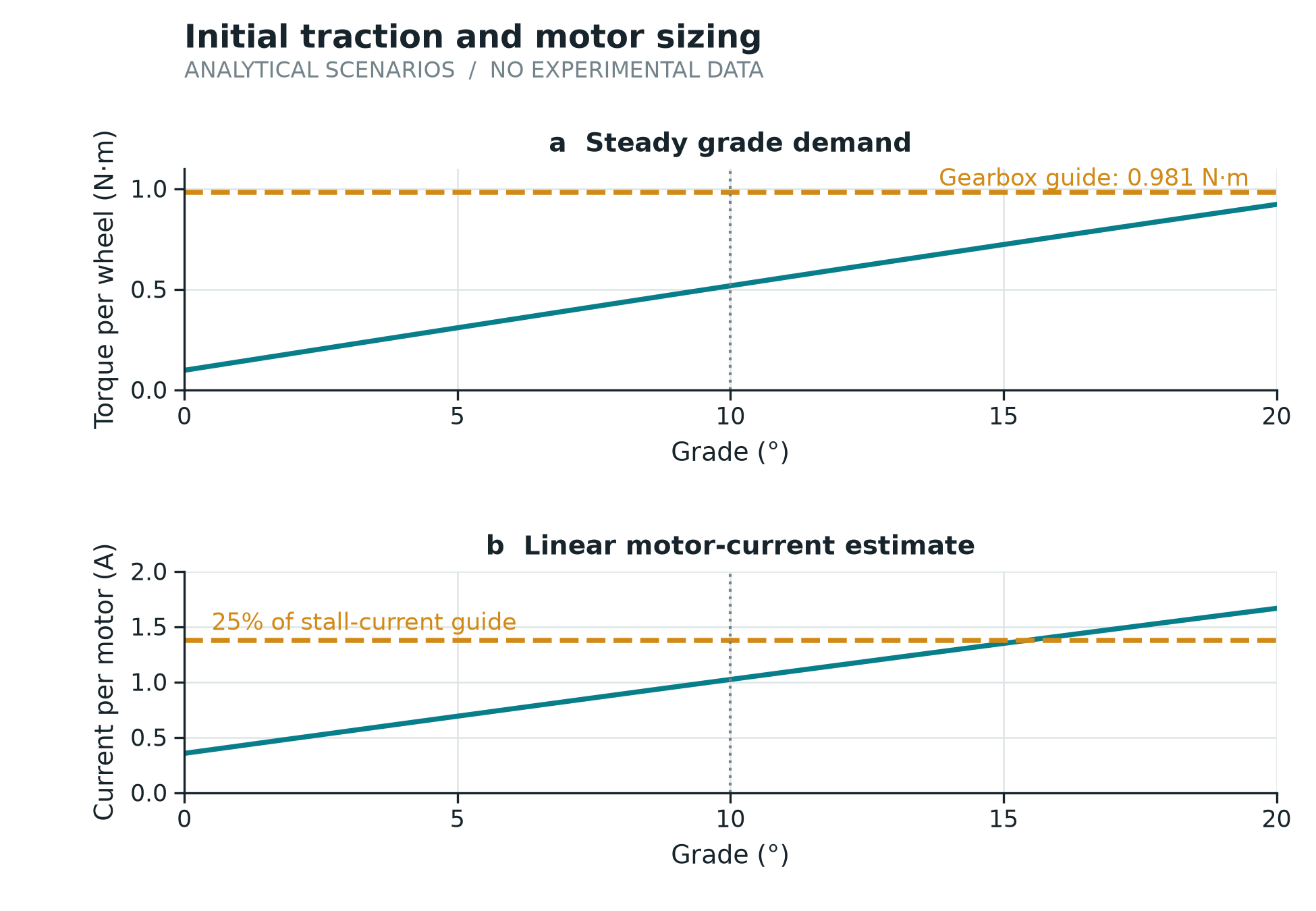}
\caption{Steady slope and motor sizing. Analytical scenario; no experimental data. Equal wheel loading, 12 kg gross mass, 165 mm wheels and rolling coefficient 0.04 are assumed. Wheel torque follows the steady grade model. Current is a linear estimate from the motor specifications at 12 V. Dashed lines are manufacturer recommendations. The 10\ensuremath{{}^{\circ}} marker is a proposed test condition. Starts, skid turning, sinkage, heat and PWM effects are excluded~\cite{ENG_MOTOR},~\cite{ENG_SKID}.}
\label{fig:ENG_F03}
\end{figure}

A linear interpolation between motor free-running and extrapolated stall current gives I \ensuremath{\approx} I\ensuremath{_{0}} + (Is \ensuremath{-} I\ensuremath{_{0}})\ensuremath{\tau}w/\ensuremath{\tau}s, approximately 1.02 A per motor for this load. This interpolation is an initial sizing estimate at the documented motor conditions, not a PWM power model or thermal guarantee. The estimated current of 1.02 A is below the 1.375 A guideline obtained from 25\% of 5.5 A. The calculated wheel torque of 0.517 N\ensuremath{\cdot}m is below the 0.981 N\ensuremath{\cdot}m continuous gearbox recommendation. Starts, sinkage, turning scrub, load transfer, and obstacles can exceed this estimate. Sustained current limiting, stall detection, temperature measurement, and a load-specific motor test are therefore required. A 10\ensuremath{{}^{\circ}} slope is a proposed validation condition, not a gradeability rating.

\subsection{Manipulator load budgeting}

The grasp demonstrator uses selected link lengths L\ensuremath{_{1}} = 0.146 m and L\ensuremath{_{2}} = 0.264 m. It must not inherit the chassis payload target. For a horizontal arm, assume upper-link mass m\ensuremath{_{1}} = 0.12 kg, forearm mass m\ensuremath{_{2}} = 0.16 kg, and wrist/gripper mass mh = 0.13 kg. These are explicit mass allocations to be replaced after weighing. For grasped mass mp, the static shoulder and elbow torques are

\begin{equation}
\tau_S=g[m_1L_1/2+m_2(L_1+L_2/2)+(m_h+m_p)(L_1+L_2)],
\label{eq:8}
\end{equation}

\begin{equation}
\tau_E=g[m_2L_2/2+(m_h+m_p)L_2]
\label{eq:9}
\end{equation}

For mp from 0.15 to 0.25 kg, calculated shoulder gravity torque is approximately 1.65--2.05 N\ensuremath{\cdot}m. Applying an assumed factor of two for design allowance produces a 3.3--4.1 N\ensuremath{\cdot}m shoulder requirement. This factor does not replace a dynamic model, shock analysis, servo thermal characterization, or structural safety assessment. An actuator's advertised stall torque cannot be interpreted as continuous usable torque. Consequently, the model is labelled a grasp demonstrator and no 0.25 kg payload rating is asserted.

The next design step is to calculate actuator torques over a joint-angle grid using actual link centers of mass and acceleration profiles, then assess current, duty, gearbox wear, joint bearing load, and allowable deflection. Grip retention also depends on contact friction and finger geometry; supporting an object against gravity is different from extracting it from debris. The proposal should restrict initial tests to inert objects in a controlled fixture and report object geometry, jaw opening, measured grasp force, success definition, and release behavior.

\begin{figure}[htbp]
\centering
\includegraphics[width=6.800in,height=3.850in,keepaspectratio]{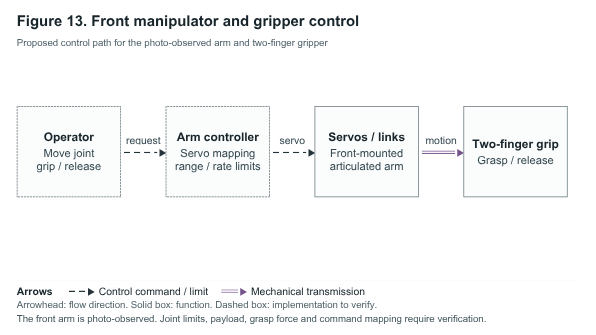}
\caption{Front manipulator. Proposed architecture with arrow conventions defined in the legend. Native editable Canva design, rendered by Canva.}
\label{fig:CANVA_F13}
\end{figure}

The payload contribution also gives a useful sensitivity check. For a horizontal extended reach of \(L_1+L_2=0.410\ \mathrm{m}\), the shoulder sensitivity is \(\partial\tau/\partial m_p=g(L_1+L_2)=4.022\ \mathrm{N\,m\,kg^{-1}}\). Adding only 50 g therefore increases the static moment by approximately 0.201 N m, or the two-times sizing requirement by 0.402 N m. Gripper adapters and cable attachments must be included in the same load budget rather than treated as negligible accessories. The calculation also explains why the chassis-carried payload allocation cannot serve as a manipulator rating. Bearing reactions, mounting compliance, cable drag and transient acceleration remain separate loads to characterize during bench evaluation.

\begin{figure}[htbp]
\centering
\includegraphics[width=6.300in,height=4.800in,keepaspectratio]{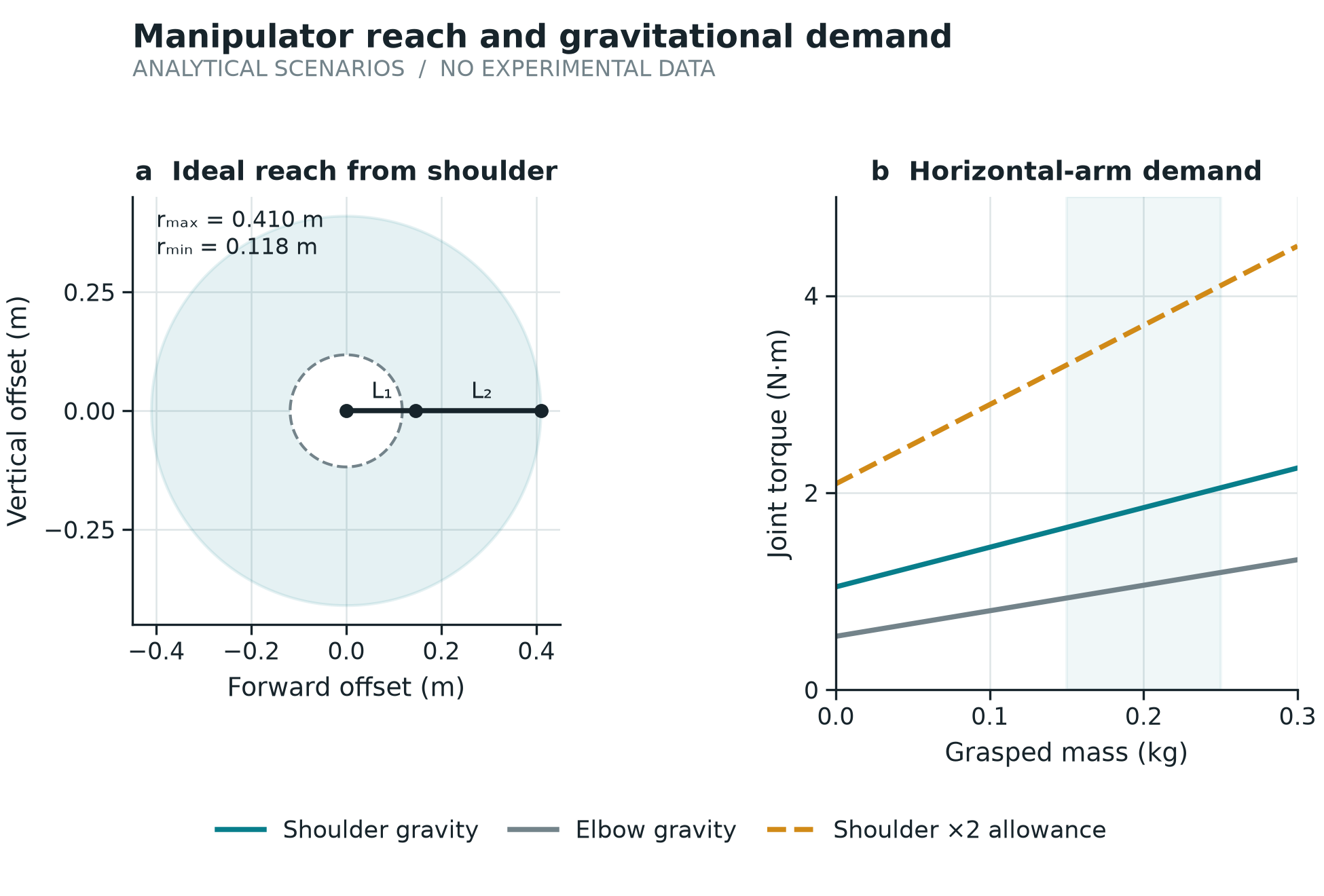}
\caption{Arm workspace envelope and torque budget. Analytical scenario; no experimental data or payload rating. The ideal annulus uses selected 146 mm and 264 mm links and ignores joint limits, collisions, cables and gripper length. Horizontal torque assumes 0.12/0.16 kg link masses and a 0.13 kg hand. The shaded 0.15--0.25 kg band is only a grasp-load scenario. The \ensuremath{\times}2 allowance does not validate dynamics or actuator thermal capability. Both annular boundaries and torque curves are generated from archived CSV data.}
\label{fig:ENG_F05}
\end{figure}

\subsection{Energy architecture and mission duty cycle}

The selected Bioenno BLF-1206A is documented as a 12 V, 6 Ah LiFePO\ensuremath{_{4}} pack with integrated protection and balancing, 12 A maximum continuous discharge, and a 24 A two-second pulse rating. Its documented compatible charger output is 14.6 V~\cite{ENG_BATTERY}. The conservative label-based nominal energy is E\ensuremath{_{n}} = 12 \ensuremath{\times} 6 = 72 Wh. This calculation deliberately uses the manufacturer's stated 12 V designation; a different nominal-cell convention must not silently change the baseline to 76.8 Wh. Actual delivered watt-hours depend on the discharge profile, cutoff, temperature, age, and cell state.

The selected 20 W Newpowa module has a 395 \ensuremath{\times} 305 \ensuremath{\times} 23 mm envelope, 19.08 V maximum-power voltage, 1.05 A maximum-power current, and 22.35 V open-circuit voltage~\cite{ENG_PANEL}. These distinct electrical quantities should appear in the power schematic. Multiplying nominal battery-system voltage by short-circuit current is not a measurement of available PV power. A solar charger matched to the pack's chemistry and protection system is required. BQ24650 documentation provides one suitable architectural reference for programmable multi-cell CC/CV charging with input-voltage regulation~\cite{ENG_BQ24650}. Pack protection, cell monitoring, and balancing are separate functions~\cite{ENG_BQ76920}. The proposed circuit remains a design requirement until its complete schematic and component ratings are established.

The motor supply requires regulation and current management across the battery's charge range. Four extrapolated motor stall currents sum to 22 A, exceeding the pack's 12 A continuous limit. This does not justify routine reliance on its brief pulse rating. A proposed continuous per-motor limit of 1.3 A, coordinated acceleration, fusing, and under-voltage handling should be verified against the finished supply. The battery-side auxiliary budget of 5 W includes electronics and an operating gas-sensor heater; lighting, sample pumps, arm motion, and wheel-steering actuation or loaded holding require separate logged increments. Any steering consumption included in a future drive increment must be identified explicitly. The motor-rail converter must cover the full pack-voltage range; maintaining 12 V below a 12 V input would require boost capability or a declared reduced-voltage operating mode. The remote inference laptop is outside the onboard energy boundary and its power must be reported separately for a system-wide comparison.

For the mission calculation, selected usable-energy fraction fu = 0.80 gives Eu = 57.6 Wh. Define battery-side drive increment Pd = 35 W as an explicit scenario, auxiliary power Pa = 5 W, drive duty d between zero and one, panel rating Pstc = 20 W, irradiance G in W\ensuremath{\cdot}m\textsuperscript{-}\textsuperscript{2}, and combined PV derating \ensuremath{\eta}pv = 0.75. Then

\begin{equation}
\bar P_{load}=P_a+dP_d,\quad P_{pv}=\eta_{pv}P_{stc}G/1000,
\label{eq:10}
\end{equation}

\begin{equation}
t=E_u/(\bar P_{load}-P_{pv}),\quad \bar P_{load}>P_{pv}
\label{eq:11}
\end{equation}

Time t is in hours because energy is in Wh and power in W. The 35 W increment is a planning parameter awaiting power logging; it is not derived from the disavowed paper. At continuous movement and no sunlight, this scenario gives 1.44 h. At 25\% drive duty, darkness gives 4.19 h. Constant 500 W\ensuremath{\cdot}m\textsuperscript{-}\textsuperscript{2} irradiance gives 7.5 W modeled PV contribution and 9.22 h at the same duty. That latter value is conditional on constant weather and operating assumptions; it is not an endurance result or a forecast for a particular day.

When modeled PV exceeds mean load, the script reports a conditionally non-depleting state rather than an infinite runtime. Nightfall, shading, storage saturation, battery charge-current limits, and time-varying demand remain outside this steady approximation. For target mission duration \(t_m\), define \(B=E_u/t_m+P_{pv}-P_{aux}\). If \(B<0\), no driving duty satisfies the duration: the auxiliary load alone exceeds the energy budget. Otherwise \(d_{\max}=\min(1,B/P_d)\). A 20 h dark mission is therefore infeasible even at zero driving duty under the stated 57.6 Wh and 5 W assumptions. A six-hour mission permits about 13.1\% driving in darkness, 34.6\% at constant 500 W\ensuremath{\cdot}m\textsuperscript{-}\textsuperscript{2}, or 56.0\% at constant 1000 W\ensuremath{\cdot}m\textsuperscript{-}\textsuperscript{2} under these assumptions.

\begin{figure}[htbp]
\centering
\includegraphics[width=6.300in,height=4.800in,keepaspectratio]{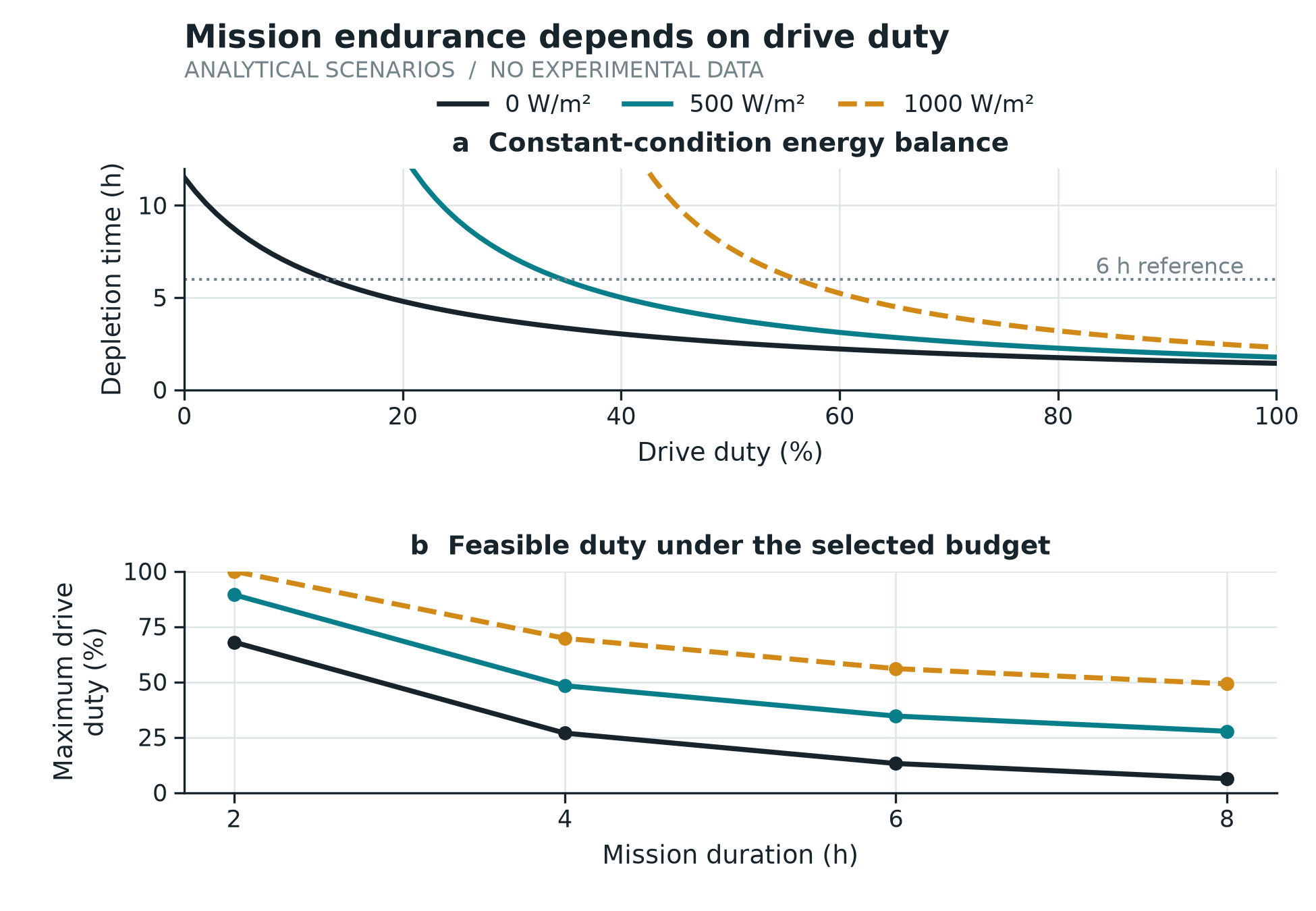}
\caption{Energy-limited mission duration. Analytical scenario; no experimental data. A 72 Wh label-based capacity, 80\% usable fraction, 5 W auxiliary load, 35 W driving increment and 0.75 PV derating define the balance. Irradiance is constant. The top panel is displayed only to 12 h; non-depleting branches are conditional, not unlimited endurance. The lower panel shows the feasible 2--8 h scenarios; the archive also records an explicitly infeasible 20 h dark case. Additional arm, steering, light and pump demand is excluded~\cite{ENG_BATTERY},~\cite{ENG_PANEL},~\cite{ENG_PVWATTS}.}
\label{fig:ENG_F01}
\end{figure}

An empty-to-full 72 Wh recharge using 15 W modeled effective full-sun input has an energy-only lower bound of 4.8 h with rover load disabled. Charge taper and changing irradiance lengthen that time. PV temperature sensitivity can be evaluated separately using Pdc = Pstc(G/1000)[1 + \ensuremath{\gamma}(Tc \ensuremath{-} 25)], where \ensuremath{\gamma} = \ensuremath{-}0.0038 K\textsuperscript{-}\textsuperscript{1} is the selected module's documented coefficient and Tc is cell temperature in \ensuremath{{}^{\circ}}C~\cite{ENG_PANEL}. Such dependence and system losses are consistent with the structure of established PV models~\cite{ENG_PVWATTS}. The separate temperature calculation must not be multiplied by the combined derating again without explaining which losses are already included.

\begin{figure}[htbp]
\centering
\includegraphics[width=6.300in,height=4.800in,keepaspectratio]{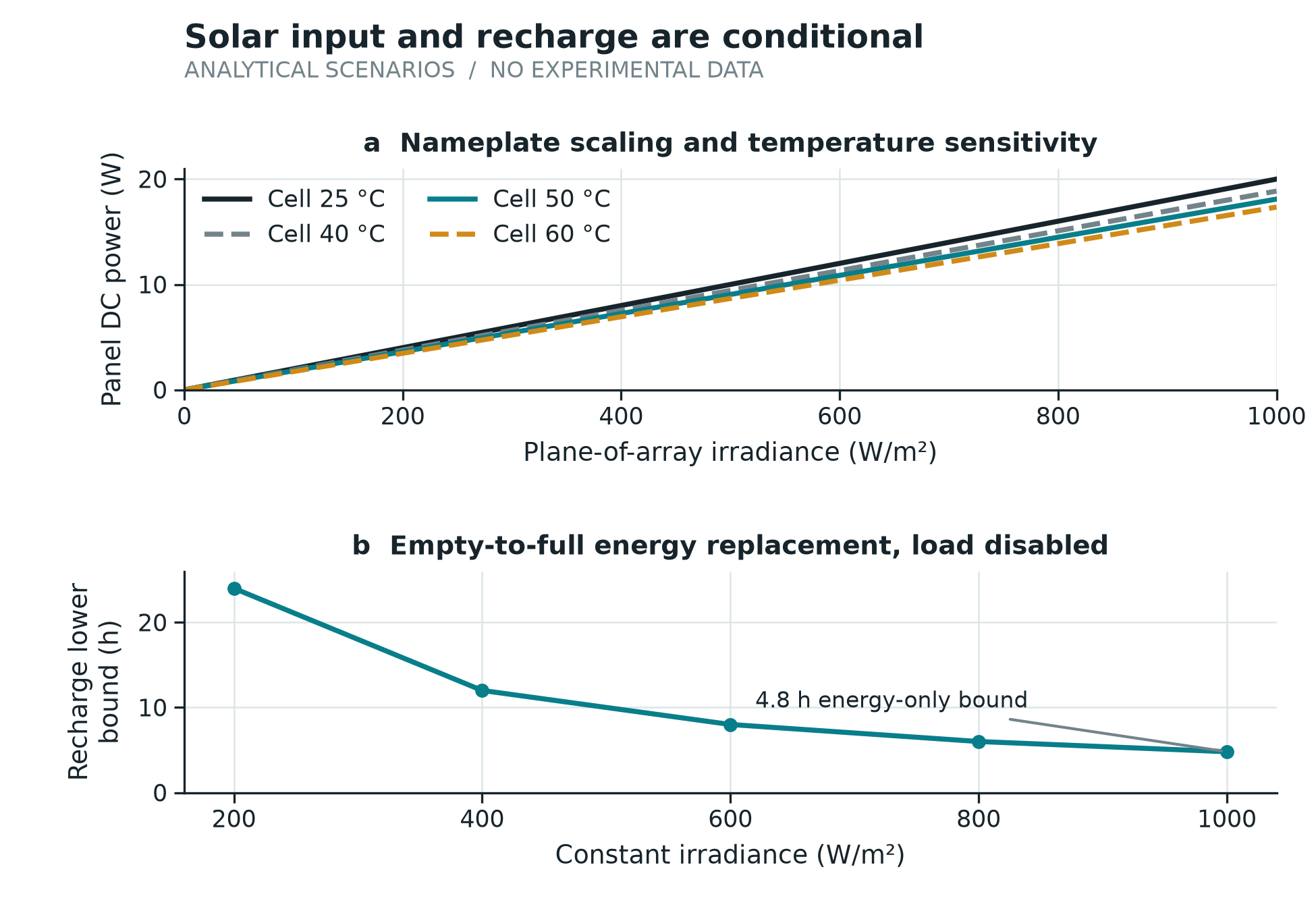}
\caption{Solar power sensitivity and recharge lower bound. Analytical scenario; no experimental data. The upper panel uses the selected 20 W module temperature coefficient without the lumped mission derating. The lower panel separately assumes 0.75 combined derating, 72 Wh replacement and no rover load. These losses must not be counted twice. Recharge values are energy-only lower bounds; weather, charge taper, shading and controller headroom are omitted. The module power-temperature coefficient is \ensuremath{-}0.38\% per kelvin~\cite{ENG_BATTERY},~\cite{ENG_PANEL},~\cite{ENG_BQ24650}.}
\label{fig:ENG_F02}
\end{figure}

\subsection{Calculated energy scenarios}

The scenarios below use the same 57.6 Wh usable-energy assumption, 5 W auxiliary load, 35 W motion increment, and 0.75 combined photovoltaic derating. Arm, steering actuation/holding, work-lamp, and any pump loads require additional terms.

\begingroup
\small
\setlength{\tabcolsep}{4pt}
\renewcommand{\arraystretch}{1.14}
\begin{longtable}{@{}>{\raggedright\arraybackslash}p{0.353\linewidth}>{\raggedright\arraybackslash}p{0.316\linewidth}>{\raggedright\arraybackslash}p{0.294\linewidth}@{}}
\caption{Calculated energy scenarios}\label{tab:4}\\
\toprule
\rowcolor{tablehead} \textbf{Scenario} & \textbf{Mean net battery demand} & \textbf{Calculated duration} \\
\midrule
\endfirsthead
\toprule
\rowcolor{tablehead} \textbf{Scenario} & \textbf{Mean net battery demand} & \textbf{Calculated duration} \\
\midrule
\endhead
\bottomrule
\endfoot
Continuous motion without sunlight & 40.00 W & 1.44 h \\[2pt]
25 percent motion duty without sunlight & 13.75 W & 4.19 h \\[2pt]
25 percent duty with constant 500 W m\textsuperscript{-}\textsuperscript{2} irradiance & 6.25 W & 9.22 h \\[2pt]
Stationary, no sunlight, auxiliary load only & 5.00 W & 11.52 h \\[2pt]
\end{longtable}
\endgroup

These values are outputs of a steady energy balance, not endurance observations. A scenario that predicts non-depletion during a sunny interval still requires finite battery capacity and a time-varying calculation spanning shade and darkness.

\subsection{Environmental measurements and uncertainty}

For gas screening, the MQ-2 is a nonselective metal-oxide sensor. Winsen specifies a flammable-gas range of 300--10000 ppm, heater consumption no greater than 950 mW, and standard preheating of at least 48 hours~\cite{ENG_MQ2}. Initial conditioning should not be confused with the stabilization time following every power interruption, which must be characterized experimentally. The heater budget and recovery time argue against treating sensor duty cycling as an energy saving with no measurement consequence. Temperature, humidity, oxygen, and mixed gases are part of the measurement problem.

For a conventional divider circuit, sensor resistance is Rs = RL(Vc/Vout \ensuremath{-} 1), where RL is load resistance in ohms and Vc and Vout are voltages. A gas-specific calibration can fit log C = a log(Rs/R\ensuremath{_{0}}) + b, where C is concentration in ppm, R\ensuremath{_{0}} is a reference resistance defined by a stated exposure, and a and b are fitted coefficients. These coefficients are not supplied by the present work. An MQ-2 voltage must not be labelled a selective CO concentration or a certified atmosphere-clearance result. The defensible initial output is a calibrated response or screening flag with interferent and out-of-range indicators. Proposed chamber work requires traceable gas standards, a reference analyzer, documented environmental conditions, and separately defined alarm metrics.

\begin{figure}[htbp]
\centering
\includegraphics[width=6.800in,height=3.850in,keepaspectratio]{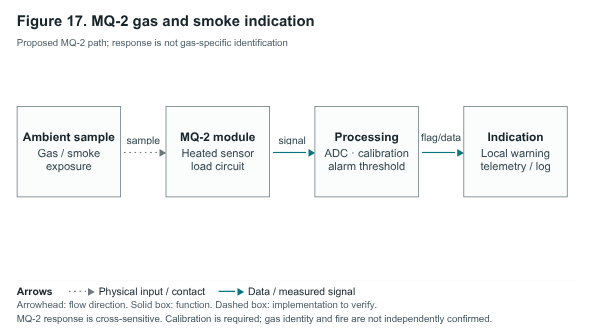}
\caption{Gas and smoke indication. Proposed architecture with arrow conventions defined in the legend. Native editable Canva design, rendered by Canva.}
\label{fig:CANVA_F04}
\end{figure}

Water sensing requires a similar distinction between electrical response and chemical interpretation. DFRobot's SEN0244 specification is \ensuremath{\pm}10\% of a 1000 ppm full scale at 25 \ensuremath{{}^{\circ}}C, equivalent to \ensuremath{\pm}100 ppm, rather than \ensuremath{\pm}10\% of every indicated value~\cite{ENG_TDS}. The module has no temperature probe. Its example compensates voltage using V\ensuremath{_{2}}\ensuremath{_{5}} = V/[1 + 0.02(T \ensuremath{-} 25)] and converts with C = 0.5(133.42V\ensuremath{_{2}}\ensuremath{_{5}}\textsuperscript{3} \ensuremath{-} 255.86V\ensuremath{_{2}}\ensuremath{_{5}}\textsuperscript{2} + 857.39V\ensuremath{_{2}}\ensuremath{_{5}})~\cite{ENG_TDS_CODE}. The coefficients are a vendor example and should not be treated as a universal relation for floodwater mixtures. Conductance depends on dissolved ionic composition, and field calibration and temperature reporting remain essential~\cite{ENG_USGS_EC}.

A partial uncertainty example uses a selected 5 V, 10-bit ADC. Quantization step q = 5/1024 V and a uniform-quantization assumption give uV = q/\ensuremath{\surd}12. With temperature standard uncertainty uT = 0.5 \ensuremath{{}^{\circ}}C assumed, the first-order output uncertainty is

\begin{equation}
u_C^2=(\partial C/\partial V)^2u_V^2+(\partial C/\partial T)^2u_T^2
\label{eq:12}
\end{equation}

This example omits calibration residuals, reference voltage error, sensor drift, fouling, ionic composition, and covariance. It must therefore be called a partial budget, not sensor accuracy. Full-scale manufacturer limits are also not automatically standard uncertainties. A defensible expanded uncertainty needs an explicitly justified distribution and coverage factor, following measurement-uncertainty practice~\cite{ENG_GUM}. The accompanying calculation illustrates why a smooth ADC trace does not establish a chemically accurate result.

\begin{figure}[htbp]
\centering
\includegraphics[width=6.800in,height=3.850in,keepaspectratio]{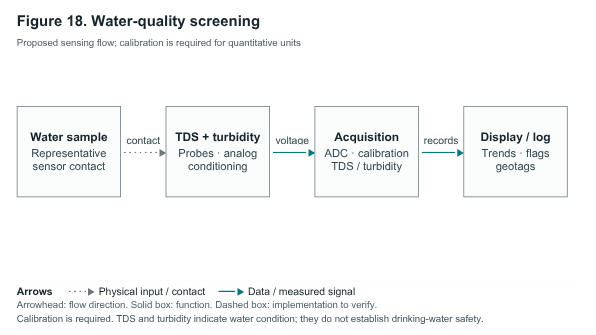}
\caption{Water-quality screening. Proposed architecture with arrow conventions defined in the legend. Native editable Canva design, rendered by Canva.}
\label{fig:CANVA_F03}
\end{figure}

The SEN0189 is suitable for exploring optical changes, but a DFRobot-hosted FAQ post describes it as qualitative and discourages conversion to standard NTU~\cite{ENG_TURB_FAQ}. Its stated electronic response below 500 ms does not establish sample equilibration, settling behavior, or low-turbidity accuracy~\cite{ENG_TURB}. A claim of quantitative low-NTU performance would require validation against a suitable nephelometer and standards across the intended range, including blank behavior and contamination controls~\cite{ENG_EPA1801}. Percentage error is undefined at a zero reference; blank bias and detection capability need absolute units. Neither turbidity nor conductivity alone establishes microbiological safety. The EPA's 500 mg\ensuremath{\cdot}L\textsuperscript{-}\textsuperscript{1} TDS value is a secondary aesthetic guideline~\cite{ENG_EPA_TDS}, while WHO turbidity guidance is process dependent~\cite{ENG_WHO_TURB}.

A proposed water-measurement run begins with a recorded sensor identity, calibration version and independent reference value. The operator acquires a blank and standards spanning the intended operating range, then measures samples under a declared rinse, immersion and stabilization procedure. Temperature is measured with a separately identified instrument and associated with the same sample interval. Standards used to fit a conversion are kept distinct from check solutions used to assess it. Replicate readings from one cup characterize short-term repeatability; they do not create independent evidence across water matrices or sites. Carryover is examined by alternating low and high check solutions, and a failed post-run check invalidates an automatic assumption that the intervening readings remained calibrated. The log retains raw voltage, reference voltage, temperature, elapsed immersion time, conversion version and quality flags. Acceptance limits for drift, blank response, repeatability and reference disagreement are specified before testing. Values outside the validated range remain flagged or unreported rather than extrapolated into apparently precise concentration or turbidity estimates.

\begin{figure}[htbp]
\centering
\includegraphics[width=6.300in,height=4.800in,keepaspectratio]{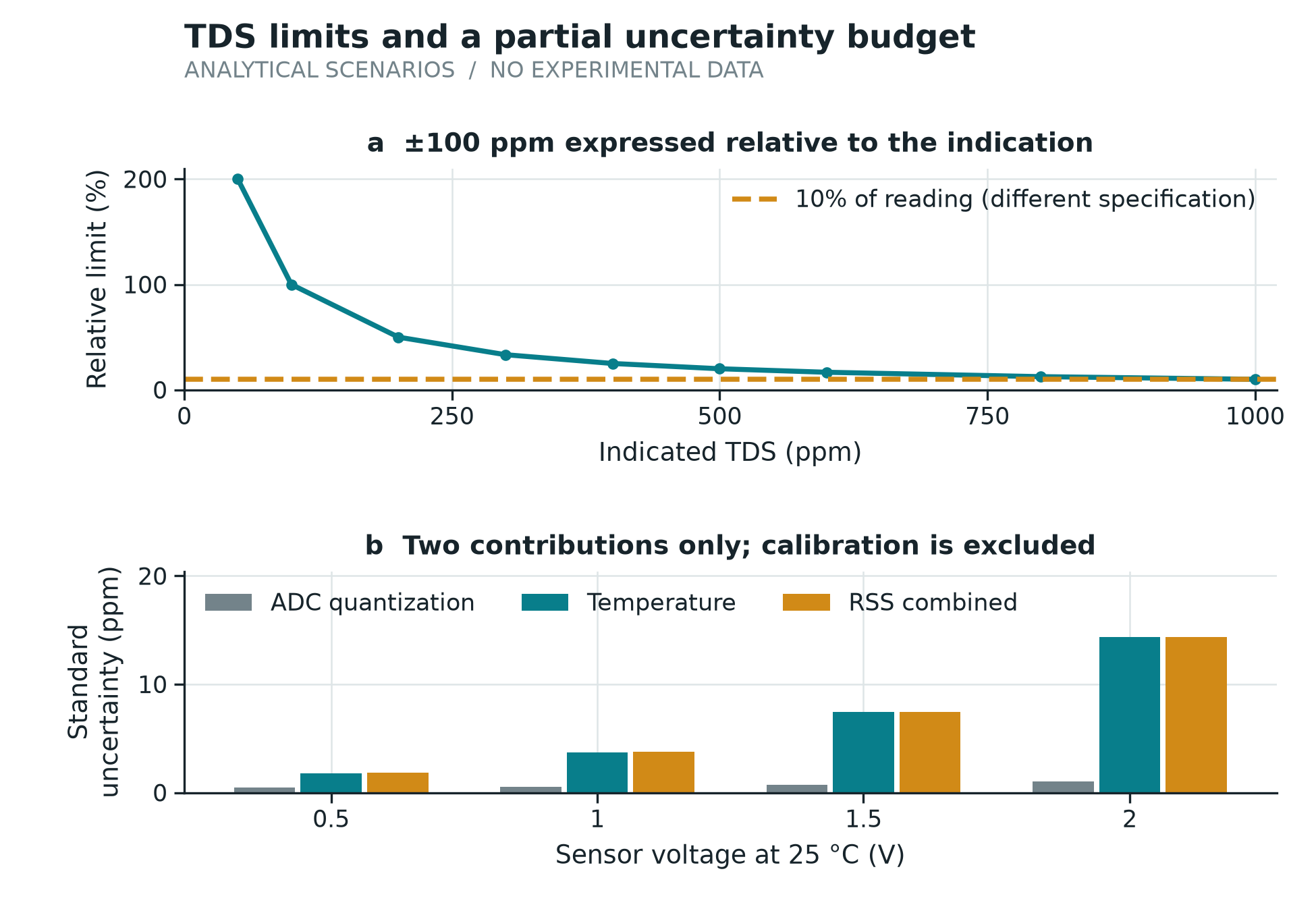}
\caption{Full-scale TDS specification and partial uncertainty. Calculated specification interpretation and analytical scenario; no experimental data. The SEN0244 specification is \ensuremath{\pm}10\% of its 1000 ppm full scale, equivalent to \ensuremath{\pm}100 ppm. The lower panel propagates selected 10-bit, 5 V ADC quantization and assumed 0.5 \ensuremath{{}^{\circ}}C temperature standard uncertainty. Calibration, matrix effects, drift, reference voltage, fouling and covariance are excluded, so the result is only a partial uncertainty budget. The manufacturer full-scale limit is not a probabilistic standard uncertainty~\cite{ENG_TDS},~\cite{ENG_TDS_CODE},~\cite{ENG_GUM}.}
\label{fig:ENG_F07}
\end{figure}

\subsection{Radiation counts and UV interpretation}

The design may include a GM counting channel, but its tube, high-voltage supply, pulse-conditioning circuit, energy response, and calibration must be specified before interpreting dose rate. GM pulse counting does not identify photon energy, and detector dead time can distort elevated rates~\cite{ENG_IAEA_GM}. The appropriate first output is counts over a stated live interval, instrument status, and a background estimate. A universal conversion from counts per minute to dose rate must not be invented.

For ideal Poisson counts with expected count \(\Lambda\) over live time \(t\), the rate estimator \(N/t\) has standard deviation \(\sqrt{\Lambda}/t\). For sufficiently informative positive counts, \(\sqrt N/t\) is a plug-in estimate, giving relative estimate \(1/\sqrt N\)~\cite{ENG_IAEA_STATS}. Low and zero counts require a declared Poisson interval procedure; zero observations do not establish zero rate. After zero counts, for example, the one-sided 95\% upper rate bound is \(-\ln(0.05)/t\). The planning curves use expected counts \(\Lambda=\lambda t\), not observations. A 1 count\ensuremath{\cdot}s\textsuperscript{-}\textsuperscript{1} scenario yields approximately 12.9\% relative counting uncertainty at 60 s and 5.8\% at 300 s. These are expected statistical limits before calibration or environmental contributions, not observed rover readings. For independent sample and background windows, the observed-count plug-in variance estimate is

\begin{equation}
r_{net}=N_s/t_s-N_b/t_b,\quad u^2(r_{net})=N_s/t_s^2+N_b/t_b^2
\label{eq:13}
\end{equation}

Low net rates can consequently require stationary integration. The natural research question is how a mission policy trades spatial coverage against measurement precision and energy. That question is more defensible than claiming immediate high-accuracy radiation mapping from unspecified commodity hardware.

\begin{figure}[htbp]
\centering
\includegraphics[width=6.800in,height=3.850in,keepaspectratio]{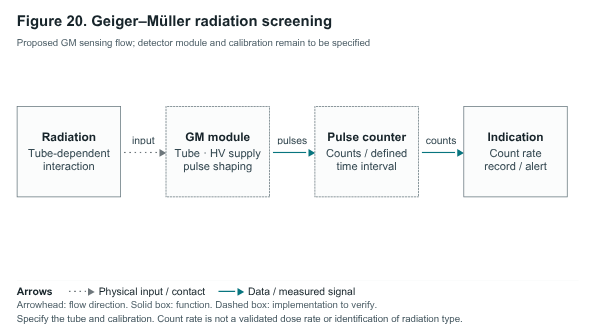}
\caption{Radiation counting. Proposed architecture with arrow conventions defined in the legend. Native editable Canva design, rendered by Canva.}
\label{fig:CANVA_F08}
\end{figure}

\begin{figure}[htbp]
\centering
\includegraphics[width=6.300in,height=4.800in,keepaspectratio]{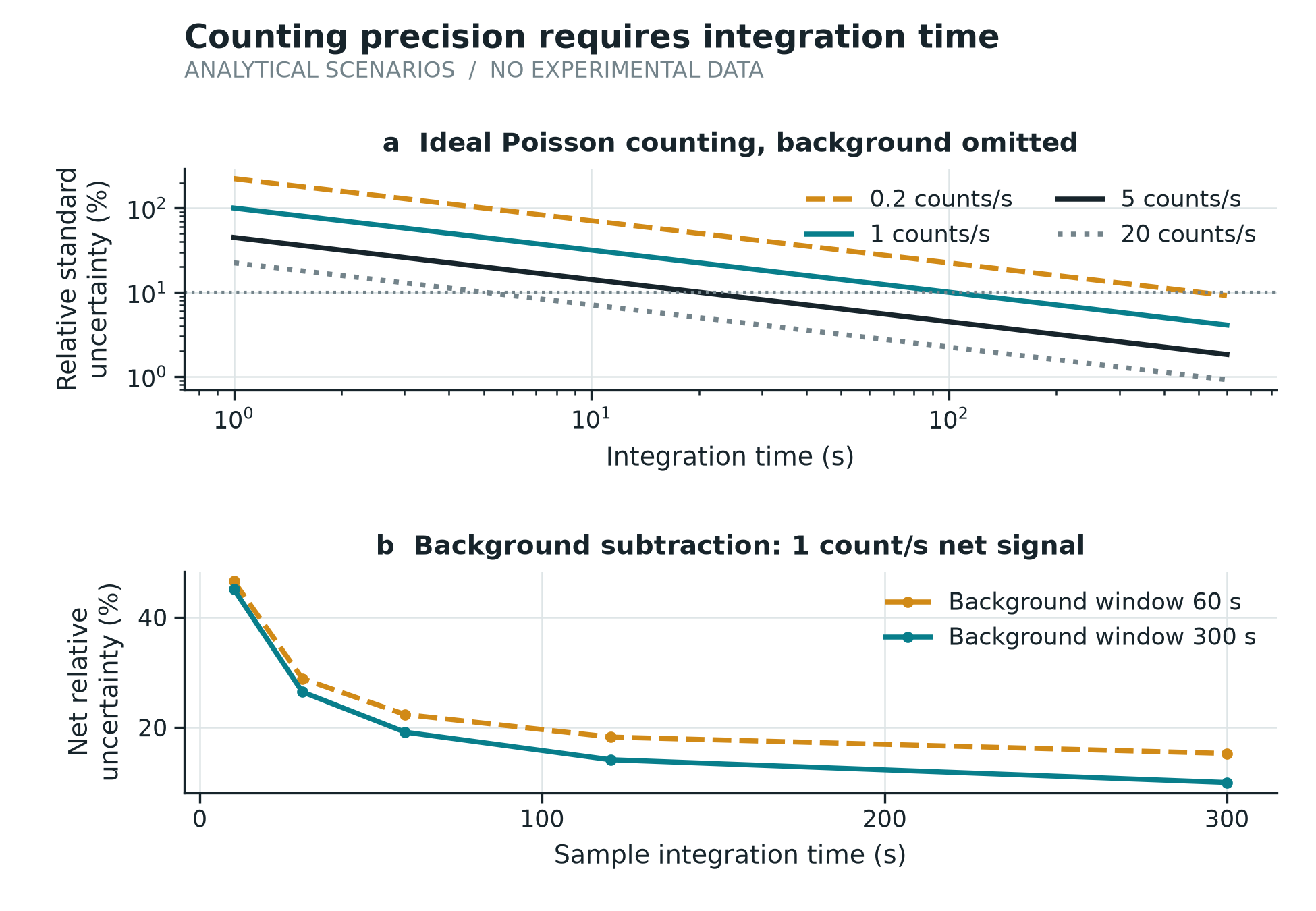}
\caption{Count integration and background uncertainty. Analytical scenario; no experimental radiation data. The upper panel uses ideal Poisson expected counts without background or dead time. The lower panel assumes a 1 count/s net signal and 1 count/s background, with independent sample and background windows. Calibration, energy response and background drift are omitted. These curves support integration-time planning, not dose-rate or safety claims~\cite{ENG_IAEA_GM},~\cite{ENG_IAEA_STATS}.}
\label{fig:ENG_F06}
\end{figure}

Solar UV sensing must specify spectral response, optical window, angular response, orientation, and reference instrument. UVI is a dimensionless index proportional to erythemally weighted solar UV irradiance. Exposure dose requires time integration and orientation/context; a single uncalibrated photodiode channel cannot substitute for that spectral weighting~\cite{ENG_UVI}. A threshold of six marks the high-UV category, but protective recommendations begin at three. The dashboard should retain the actual estimated index, uncertainty or quality flag, timestamp, and sensor orientation. No UV accuracy, personal exposure estimate, or forecast capability is claimed for the present design.

\begin{figure}[htbp]
\centering
\includegraphics[width=6.800in,height=3.850in,keepaspectratio]{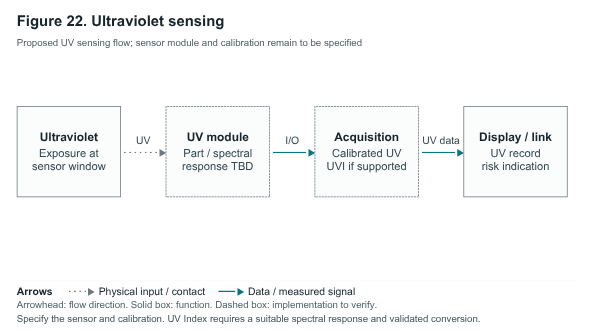}
\caption{UV monitoring. Proposed architecture with arrow conventions defined in the legend. Native editable Canva design, rendered by Canva.}
\label{fig:CANVA_F07}
\end{figure}

\subsection{Positioning and vision as distributed measurements}

The NEO-6M provides an economical outdoor geotagging option. Its data sheet lists a 27 s cold-start figure and 2.5 m horizontal CEP under specified static conditions; CEP is a 50\% containment measure, not a 95\% radial-error bound~\cite{ENG_NEO6}. A GNSS evaluation must state reference coordinates, antenna, satellite geometry, date and time, environment, sampling interval, outages, and statistical definition. Consecutive fixes are temporally correlated and should not be counted as independent repetitions merely because many records were logged. The application should distinguish antenna position from the location of a deployed sampling tip and flag data acquired while the rover was moving.

Geotagging also requires a geometric reference. Let \(\mathbf p_a^w\) be antenna position in the chosen world frame, \(R_b^w\) the body-to-world rotation, and \(\mathbf r_{a\rightarrow s}^b\) the fixed antenna-to-sensor displacement expressed in the body frame. The corresponding sensor position is

\begin{equation}
\mathbf p_s^w=\mathbf p_a^w+R_b^w\mathbf r_{a\rightarrow s}^b
\label{eq:14}
\end{equation}

For an articulated sampling point, the displacement becomes a function of measured joint configuration. An antenna fix alone cannot establish that point's location when attitude or joint state is unknown. The prototype should therefore store antenna fixes, pose estimates and mounting offsets separately, rather than overwriting them with a single apparently precise coordinate. Verification should compare stationary fixtures and repeated approaches to surveyed locations. These checks distinguish antenna positioning error, coordinate-transformation error and measurement-timestamp mismatch before spatial maps are interpreted.

\begin{figure}[htbp]
\centering
\includegraphics[width=6.800in,height=3.850in,keepaspectratio]{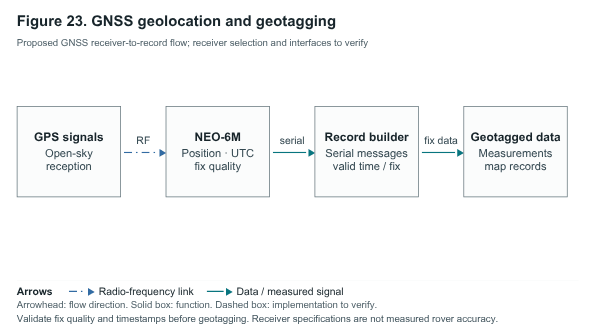}
\caption{GNSS geotagging. Proposed architecture with arrow conventions defined in the legend. Native editable Canva design, rendered by Canva.}
\label{fig:CANVA_F06}
\end{figure}

The camera architecture explicitly separates capture from inference. The ESP32 camera transmits JPEG frames to a laptop or other edge computer that runs object detection. ESP32 memory and camera-driver constraints justify recording frame size, compression quality, buffers, PSRAM configuration, firmware version, and competing Wi-Fi/SD activity~\cite{ENG_ESP32},~\cite{ENG_CAMERA}. The MJPEG2SD project reports recording throughput for a particular camera board, SD card, clock, and quality setting~\cite{ENG_MJPEG}. Those values are software-author benchmarks, not rover measurements, and its motion-detection timing must not be relabelled YOLO inference time.

Scaled-YOLOv4 reports YOLOv4-tiny at 22.0\% COCO AP and 42.0\% AP50 with 443 FPS on an RTX 2080Ti, under the authors' configuration~\cite{ENG_SCALED}. AP50 and AP averaged across IoU thresholds differ~\cite{ENG_COCO}. Neither score establishes person-finding reliability in dust, debris, occlusion, low light, or unfamiliar viewpoints. The external GPU throughput is not used as a Zephyron latency estimate. The proposed evaluation will report the selected edge hardware, model file checksum, inference runtime, precision, input size, confidence threshold, nonmaximum suppression, and task-specific held-out evaluation.

An ideal capture-opportunity and processing-delay model is

\begin{equation}
T=1/(2f_c)+8S/R+T_{decode}+T_{infer}+T_{post}+T_{display}+T_{queue}
\label{eq:15}
\end{equation}

Here fc is capture rate in frames\ensuremath{\cdot}s\textsuperscript{-}\textsuperscript{1}, S compressed frame size in bytes, and R network throughput in bit\ensuremath{\cdot}s\textsuperscript{-}\textsuperscript{1}; all times are seconds. The ideal mean acquisition-opportunity wait 1/(2fc) assumes events arrive uniformly between captured frames. When capture opportunities are discarded or throttled, the additional wait for an accepted frame is not represented by this term and must be measured from the accepted-stream schedule. For selected scenarios fc = 5 FPS, S = 40000 bytes, R = 2 Mbit\ensuremath{\cdot}s\textsuperscript{-}\textsuperscript{1}, inference 50 ms, and other processing 50 ms, with queueing omitted, T = 360 ms. At 0.35 m\ensuremath{\cdot}s\textsuperscript{-}\textsuperscript{1} the rover travels 0.126 m during this interval. This is not a safe stopping-distance estimate: braking, jitter, retransmission, detection failure, and operator response are additional terms. Bounded queues or latest-frame processing should be evaluated to prevent stale visual decisions. These are conditional, queue-free sums of an ideal capture-opportunity wait and accepted-frame processing. They are not sustainable event-to-display means for an overloaded stream. If acquisition exceeds service capacity, source throttling or replacement/discard is necessary before a growing queue forms. Capture frequency is not delivered-frame frequency. For example, 20 FPS at 40,000 bytes per frame offers 6.4 Mbit/s; a 2 Mbit/s link carries at most 6.25 such frames/s before overhead, requiring at least 68.75\% throttling/discard in this ideal case. The archived partial bottleneck upper bound excludes stage-concurrency and additional processing constraints and is not a throughput result.

INT8 quantization maps a floating value approximately as x = scale(q \ensuremath{-} zero\_point). Eight-bit representations can reduce weight storage relative to FP32, but realized speed depends on the execution hardware and supported operators~\cite{ENG_QUANT}. Static post-training quantization requires representative calibration inputs and may change detection quality~\cite{ENG_ONNX}. Calibration images and final test images must be disjoint at the scene or sequence level. No percentage speedup, model-size saving, or unchanged accuracy is asserted before implementing and measuring both models. Report end-to-end latency distributions and energy as well as inference-only timing; throughput and latency must not be treated as interchangeable.

\begin{figure}[htbp]
\centering
\includegraphics[width=6.300in,height=4.800in,keepaspectratio]{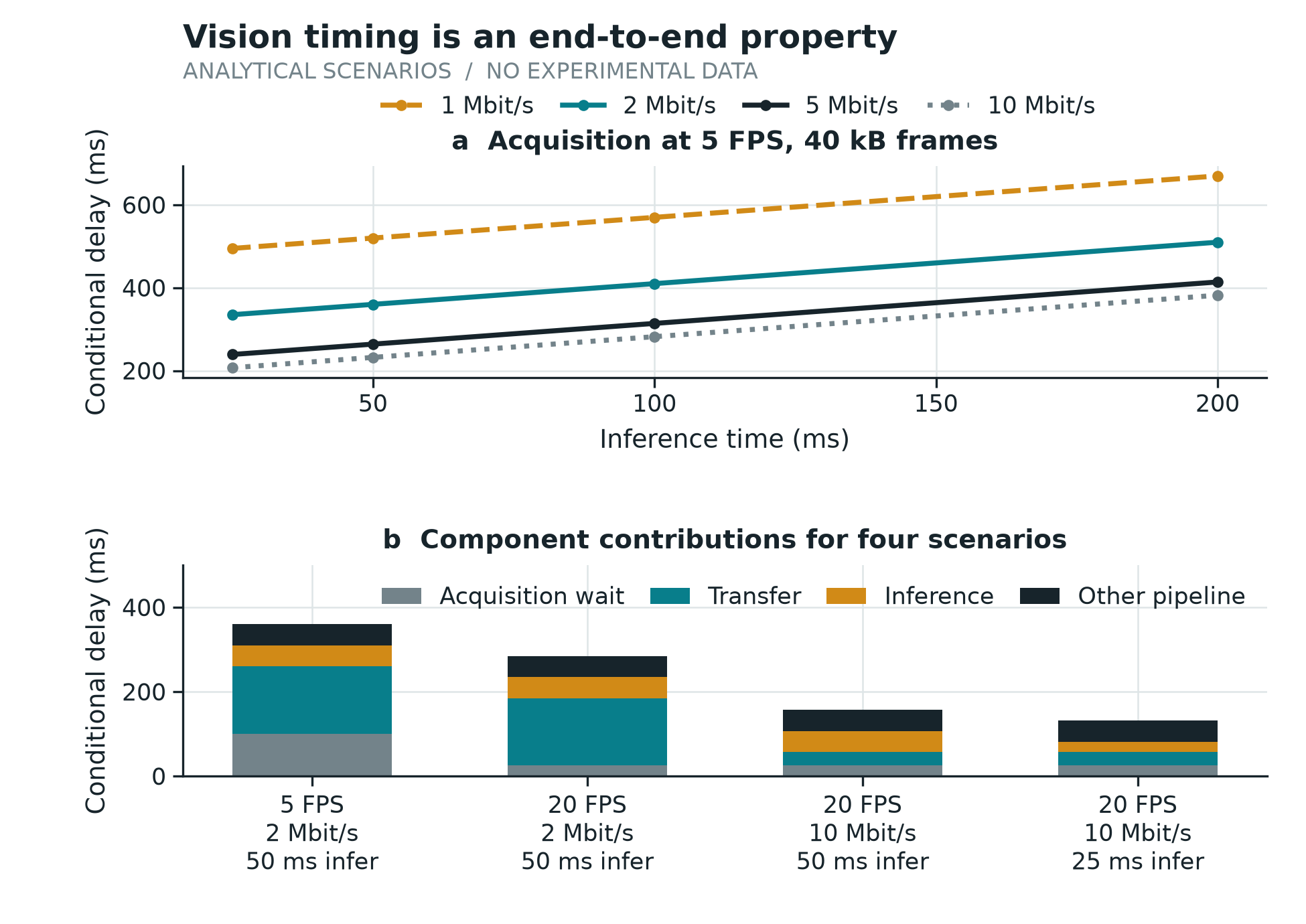}
\caption{Vision pipeline latency scenarios. Analytical scenario; no hardware timing data. Each sum combines ideal capture-opportunity wait 1/(2fc) with processing and transfer of an accepted frame: 8S/R for 40,000 bytes plus inference and 50 ms for other stages. Prior queueing, retransmission, jitter and missed detections are omitted. Capture rates exceeding network capacity require throttling or discard; skipped capture opportunities add waiting not modeled here. Thus overloaded cases are conditional timing sums, not sustainable event-to-display means. The 360 ms sum gives 0.126 m travel at 0.35 m/s, excluding braking and human response. Inference is assigned to a remote edge host; other pipeline time includes decoding, postprocessing and display~\cite{ENG_CAMERA},~\cite{ENG_MJPEG},~\cite{ENG_SCALED},~\cite{ENG_ONNX}.}
\label{fig:ENG_F09}
\end{figure}

\FloatBarrier
\section{Mission design and quality-aware sampling}

\subsection{A common mission state model}

The mission controller is organized around initialization, supervised travel, stationary acquisition, review, and return or shutdown. Initialization verifies available sensors, battery state, storage capacity, and communication status. Travel moves the platform toward an operator-selected inspection point under a defined steering mode. Acquisition holds the rover while a sensor stabilizes or integrates. Review attaches quality flags and presents results to the operator. Return is triggered by an explicit command or a budget threshold. These states describe proposed behavior; no autonomous deployment is asserted.

The design motivation is that different channels require different evidence windows. A radiation count estimate may improve with longer integration; a metal-oxide gas sensor may require conditioning and recovery; a water probe requires representative contact with a sample; and a camera frame may become stale while waiting in a network queue. A uniform fixed sampling rate does not solve these distinct constraints. The rover should therefore associate each action with an expected energy cost, duration, movement requirement, and measurement-quality condition.

Propagate estimated usable energy on intervals that resolve load and charging changes. Let \(P_{\mathrm{load},a,j}\) and \(P_{\mathrm{pv},j}\) denote battery-side load and available charging power for action \(a\) over interval \(\Delta t_j\) in seconds. The charging contribution respects charger and pack limits. With selected usable capacity \(E_{\max}\), apply capacity saturation at each interval:

\begin{equation}
E_{j+1}=\min\!\left(E_{\max},\max\!\left[0,E_j+\frac{\Delta t_j}{3600}\left(P_{\mathrm{pv},j}-P_{\mathrm{load},a,j}\right)\right]\right)
\label{eq:16}
\end{equation}

The factor 1/3600 converts watt-seconds to watt-hours. A return/recovery reserve is a route-dependent planning allocation, not a universal percentage. If \(\underline E_j\) denotes a conservative energy estimate and \(E_{\mathrm{reserve},j}\) the required reserve at that state, the path condition is

\begin{equation}
\min_j\left(\underline E_j-E_{\mathrm{reserve},j}\right)\geq0
\label{eq:17}
\end{equation}

An action is feasible only when this condition, actuator constraints and sensor-quality prerequisites hold throughout the predicted trajectory. A depleted or reserve-violating intermediate state rejects the candidate before any clipping can conceal the violation; subsequent sunshine cannot retroactively make it feasible. Capacity saturation must also occur before a later discharge interval. For example, charging from 50 Wh at a net 15 W for one hour reaches the selected 57.6 Wh usable-capacity limit; a following hour at net 15 W discharge ends at 42.6 Wh, rather than 50 Wh. The calculation is a constructed energy example. Battery-state uncertainty, converter loss, weather uncertainty and temporal resolution belong in the feasibility margin. Holding position consumes energy and requires its own feasible fallback budget; otherwise the supervisor requests a defined recovery or shutdown action.

\subsection{Precision and coverage tradeoffs}

For an expected stationary count rate lambda and desired relative counting standard uncertainty epsilon, the ideal Poisson planning time is

\begin{equation}
t_{min}=\frac{1}{\lambda\epsilon^2}
\label{eq:18}
\end{equation}

This expression applies to an ideal single-rate count process. It does not include uncertainty in background subtraction, calibration, dead time, or a changing source field. Nevertheless, it explains the design tradeoff: reducing relative uncertainty by a factor of two requires four times as many expected counts. At an assumed one count per second, a 10 percent target requires 100 s, while a 5 percent target requires 400 s. Those durations consume battery energy and reduce the number of locations that can be visited in a fixed mission period.

For gas mapping, traversal speed and sensor dynamics determine the spatial interpretation of a reading. A first-order response approximation has time constant \(\tau_s\) and a characteristic travel distance

\begin{equation}
\ell_s=v\tau_s
\label{eq:19}
\end{equation}

This is a response-length scale, not a calibrated plume-location error. Transport, turbulence, platform wake, and gas-specific adsorption can produce more complex behavior. The relation still shows why a sensor voltage recorded at the current GNSS fix may reflect exposure encountered earlier along the path. Stop-and-sample operation, response identification, and explicit lag handling should be compared before displaying a continuous concentration map~\cite{LIT_GAS2019},~\cite{LIT_GADEN2017}.

Candidate actions can be ranked by a proposed utility

\begin{equation}
J(a)=w_I\,\Delta I(a)-w_E\,\frac{\Delta E(a)}{E_{usable}}-w_T\,\frac{\Delta t(a)}{T_{mission}}-w_Q\,q_{invalid}(a),
\label{eq:20}
\end{equation}

subject to the feasibility conditions above. \(\Delta I\) is an explicitly chosen information or coverage benefit; \(q_{invalid}\) is a penalty for invalid or low-quality measurements. The weights are design parameters and require a declared selection procedure. No optimality, reinforcement-learning performance, or superiority over a fixed policy is claimed. The formulation is useful because it identifies what must be measured to justify a future scheduling decision and prevents an energy-only objective from rewarding rapid collection of uninterpretable data.

\subsection{Water, gas, and radiation tasks}

The water task approaches a controlled sample station, positions the fixed probe in contact with the sample, records raw and compensated signals, and stores a geotag and sampling identifier. A separate container is required for independent reference analysis. The illustrated well in the Blender scene is a sampling fixture selected to suit the existing probe position. It is not evidence of waterproofing or an ability to drive through floodwater.

The gas task approaches a representative source region under operator supervision, stabilizes the sensor, and records response with environmental context. The illustrated streamlines are explanatory graphics; they are not a computational fluid dynamics result or measured concentration field. A combustible-gas screening channel cannot establish that an atmosphere is safe for entry, and the present rover has no demonstrated explosion-protection classification. The proposed research tests should use a suitable controlled facility and a reference instrument rather than uncontrolled hazardous releases.

\begin{figure}[htbp]
\centering
\includegraphics[width=6.650in,height=2.600in,keepaspectratio]{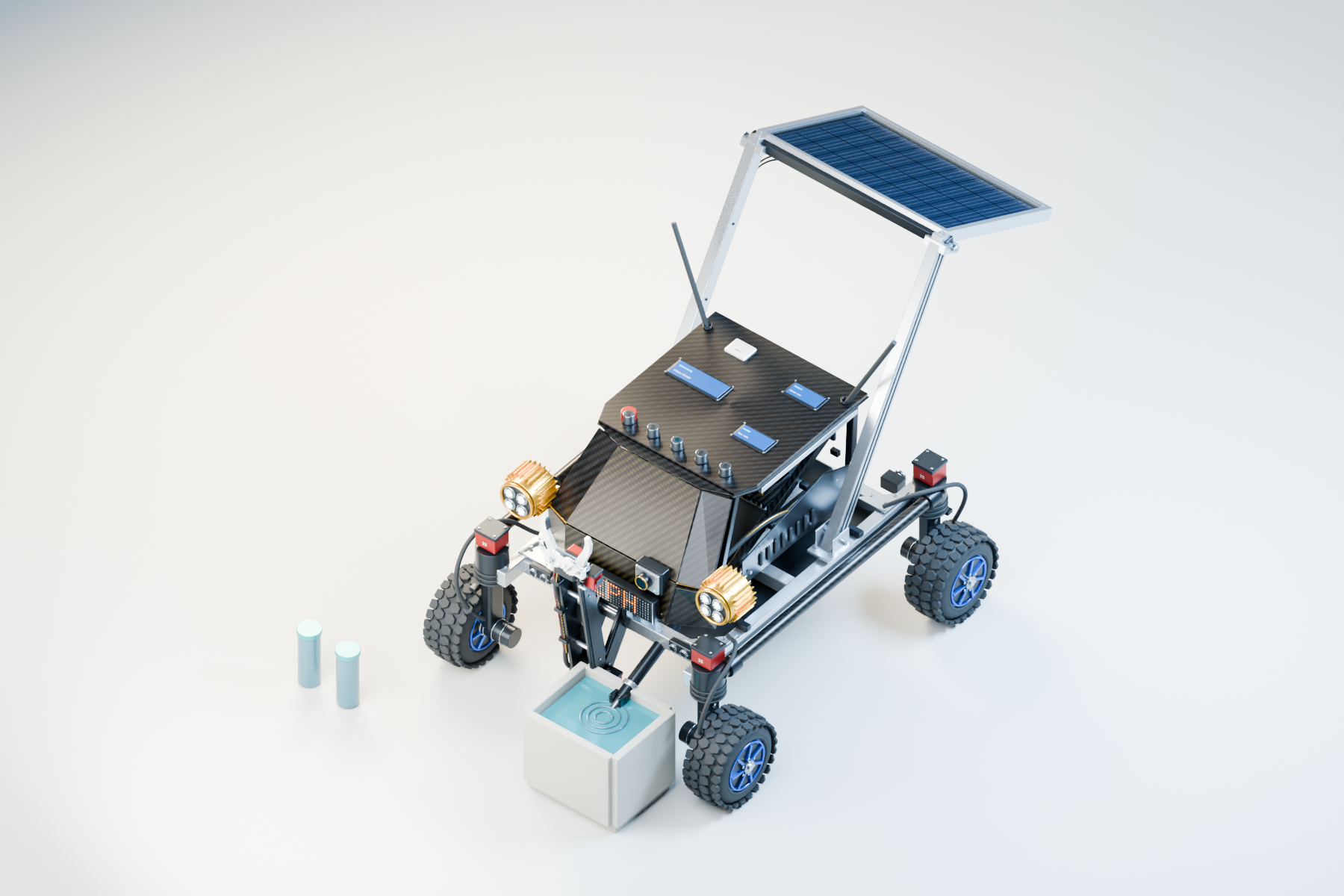}
\caption{Water screening at a raised sample fixture suited to the fixed probe location. The separate vials represent reference samples; immersion and sensor conversion require physical validation. Blender mission illustration of proposed behavior.}
\label{fig:MISSION_WATER}
\end{figure}

\begin{figure}[htbp]
\centering
\includegraphics[width=6.650in,height=2.600in,keepaspectratio]{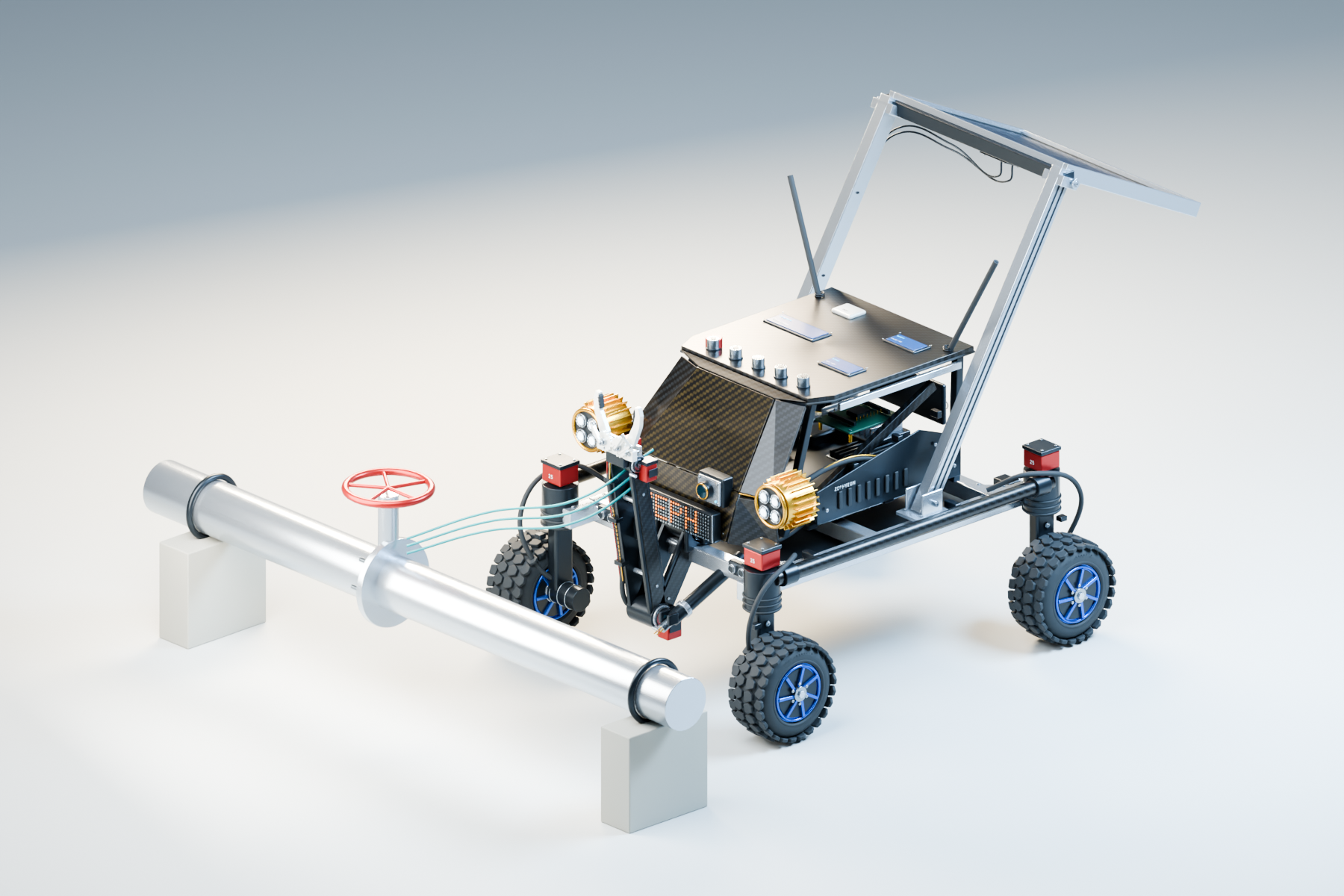}
\caption{Gas-source inspection around a process-pipe fixture. Teal streamlines are explanatory graphics, not measured gas concentrations or a fluid-dynamics simulation. Blender mission illustration of proposed behavior.}
\label{fig:MISSION_GAS}
\end{figure}

The radiation task consists of stationary count windows at defined waypoints. A map should retain integration time and statistical uncertainty, not only a colored scalar value. Background counts and source counts are acquired with documented geometry and live intervals. The Blender containers and sampling grid are illustrative training objects. Published radiation-robot studies provide useful experimental designs, but results obtained using radiation surrogates must remain identified as surrogate experiments~\cite{LIT_RADNAV2021},~\cite{LIT_RADHOT2019}.

\begin{figure}[htbp]
\centering
\includegraphics[width=6.650in,height=3.050in,keepaspectratio]{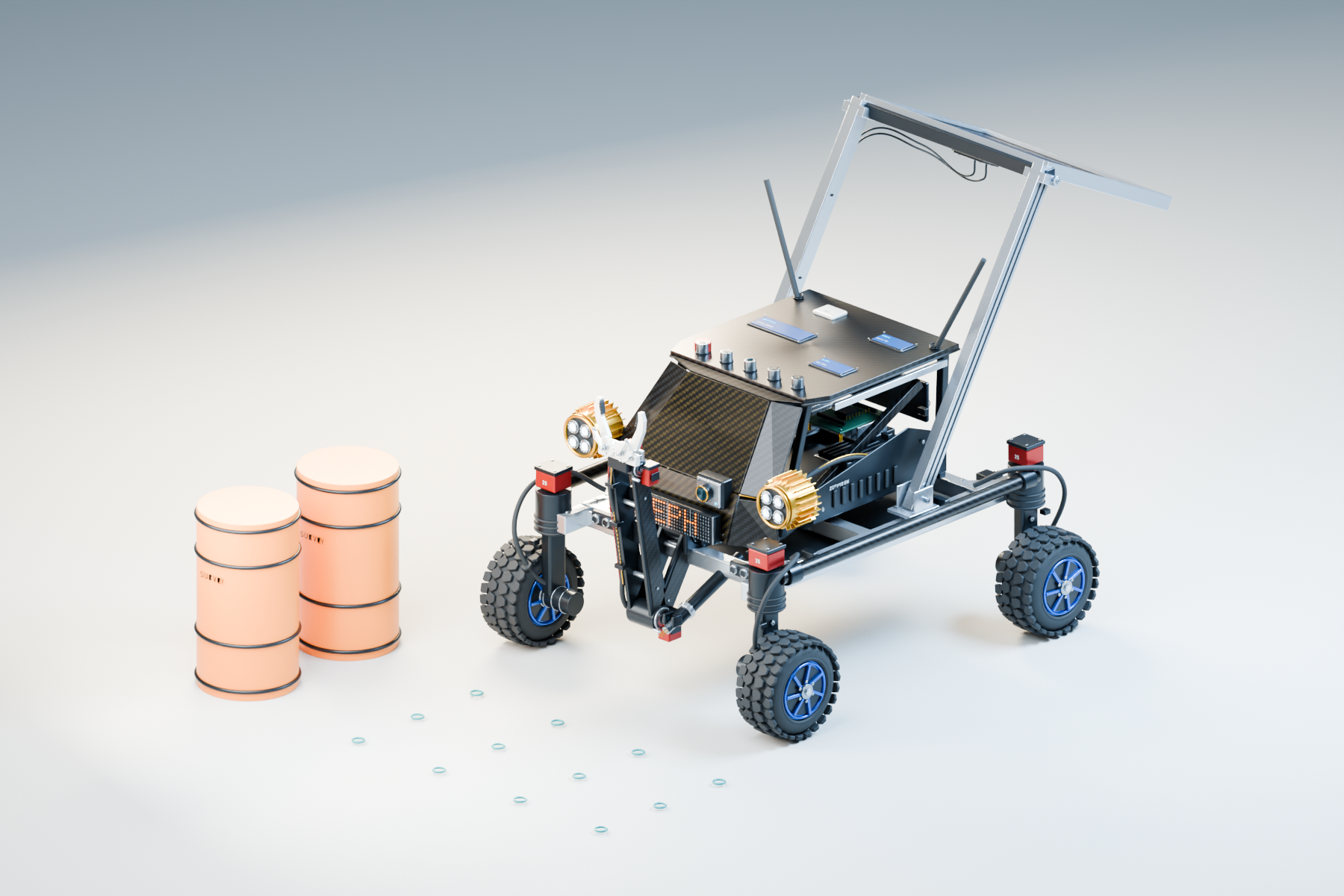}
\caption{Stationary radiation survey at a waypoint grid. Containers are illustrative training objects; no source activity, measured count rate, or dose map is represented. Blender mission illustration of proposed behavior.}
\label{fig:MISSION_RAD}
\end{figure}

\subsection{Mobility, retrieval, and stationary monitoring}

The controlled obstacle task evaluates approach geometry and stopping behavior before attempting a traversal. The rendered rubble layout depicts an inspection fixture, not an accomplished climb. Ground clearance alone cannot determine whether a wheel can lift onto an obstacle: torque, tire contact, friction, mass distribution, and the remaining wheel contacts determine the maneuver. Test reports should state obstacle shape, height, approach angle, surface, and success criterion, with current and temperature traces during each attempt.

\begin{figure}[htbp]
\centering
\includegraphics[width=6.650in,height=3.050in,keepaspectratio]{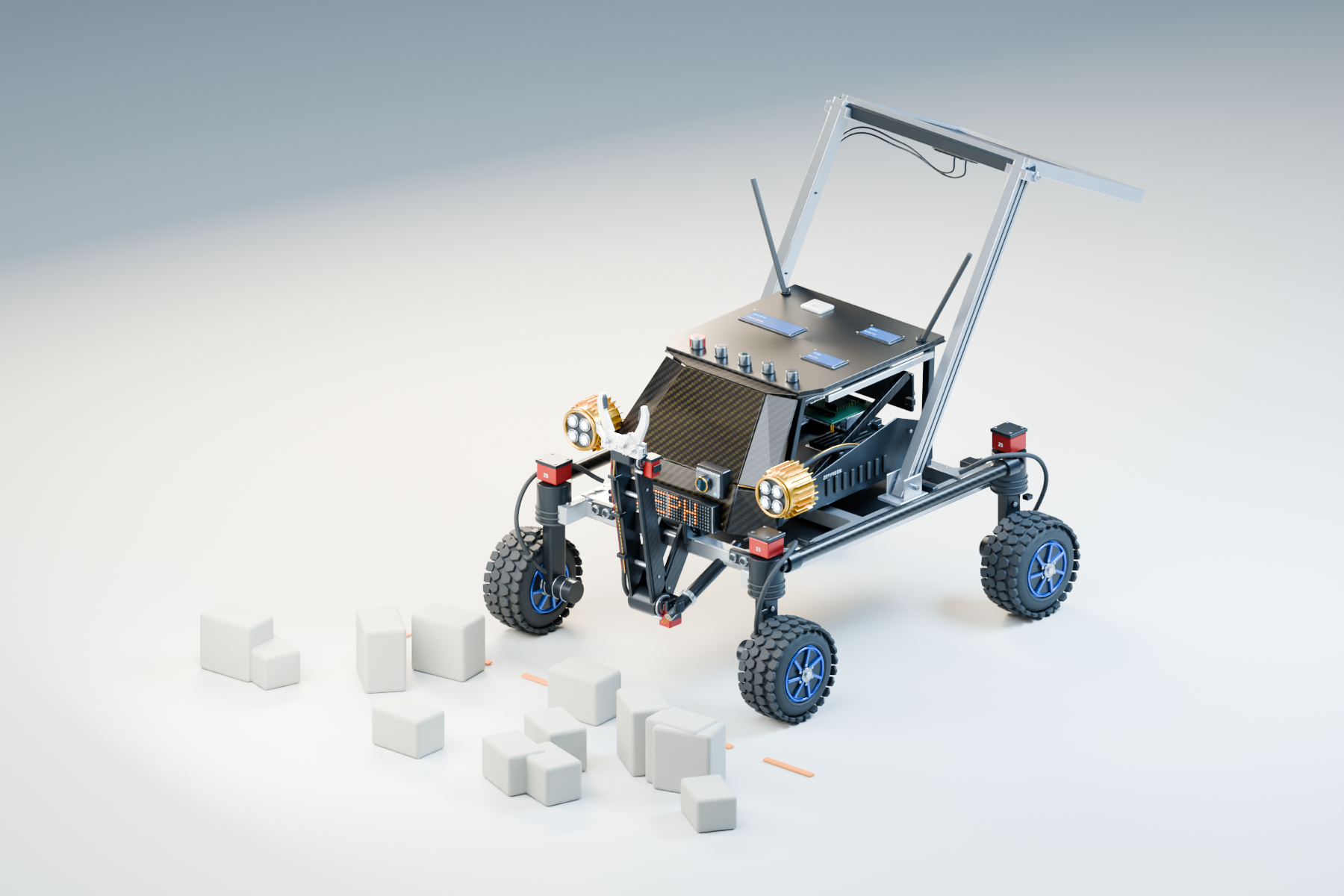}
\caption{Supervised approach to a controlled obstacle fixture. Blocks and stop line illustrate a proposed test arrangement; no traversability or braking result is shown. Blender mission illustration of proposed behavior.}
\label{fig:MISSION_MOBILITY}
\end{figure}

The retrieval task separates approach, arm deployment, grasp closure, lifting, retention, transport, and release. A successful closure command is not equivalent to retaining an object. The design must evaluate each stage against jaw geometry, friction, load, and actuator limits. The deployed model illustrates a lightweight canister retrieval arrangement. Its pose is an editable visualization and does not establish a rated grasp capacity or verified collision-free motion.

\begin{figure}[htbp]
\centering
\includegraphics[width=6.650in,height=3.050in,keepaspectratio]{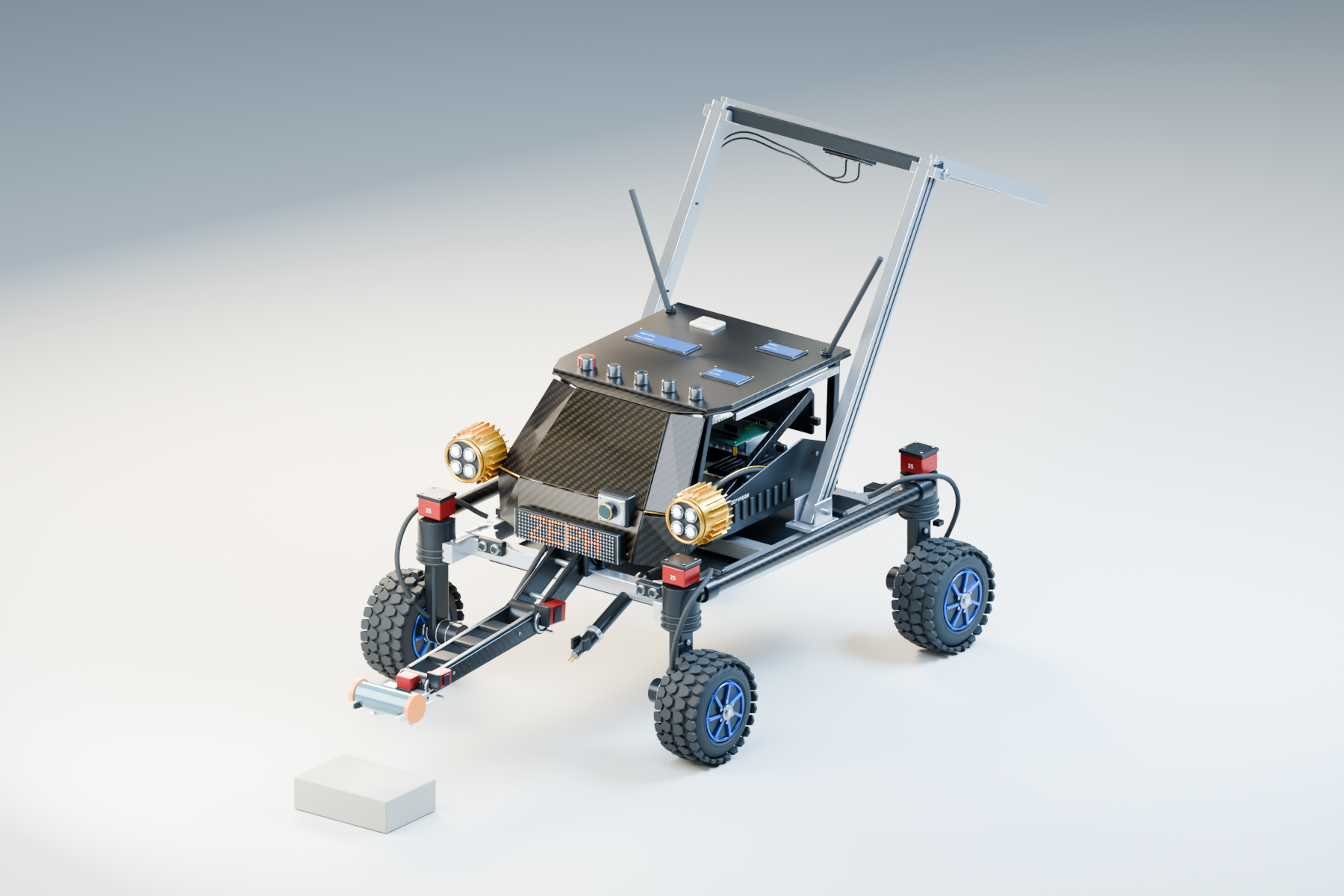}
\caption{Deployed manipulator approaching a lightweight sample canister. The pose communicates reach and task geometry; it does not establish continuous actuator capability, grasp force, or rated payload. Blender mission illustration of proposed behavior.}
\label{fig:MISSION_ARM}
\end{figure}

Stationary solar operation reduces the movement duty while preserving selected sensing and logging functions. The elevated rear panel remains fixed in the illustrated collection pose. The solar model must use incident irradiance on the panel plane and include temperature and conversion losses, rather than assuming rated output throughout a day. UV monitoring requires a specified optical response and orientation; placing a sensor near a solar module does not calibrate it. Antenna and panel placement should be inspected for shadowing and electromagnetic interference during integration.

\begin{figure}[htbp]
\centering
\includegraphics[width=6.650in,height=3.050in,keepaspectratio]{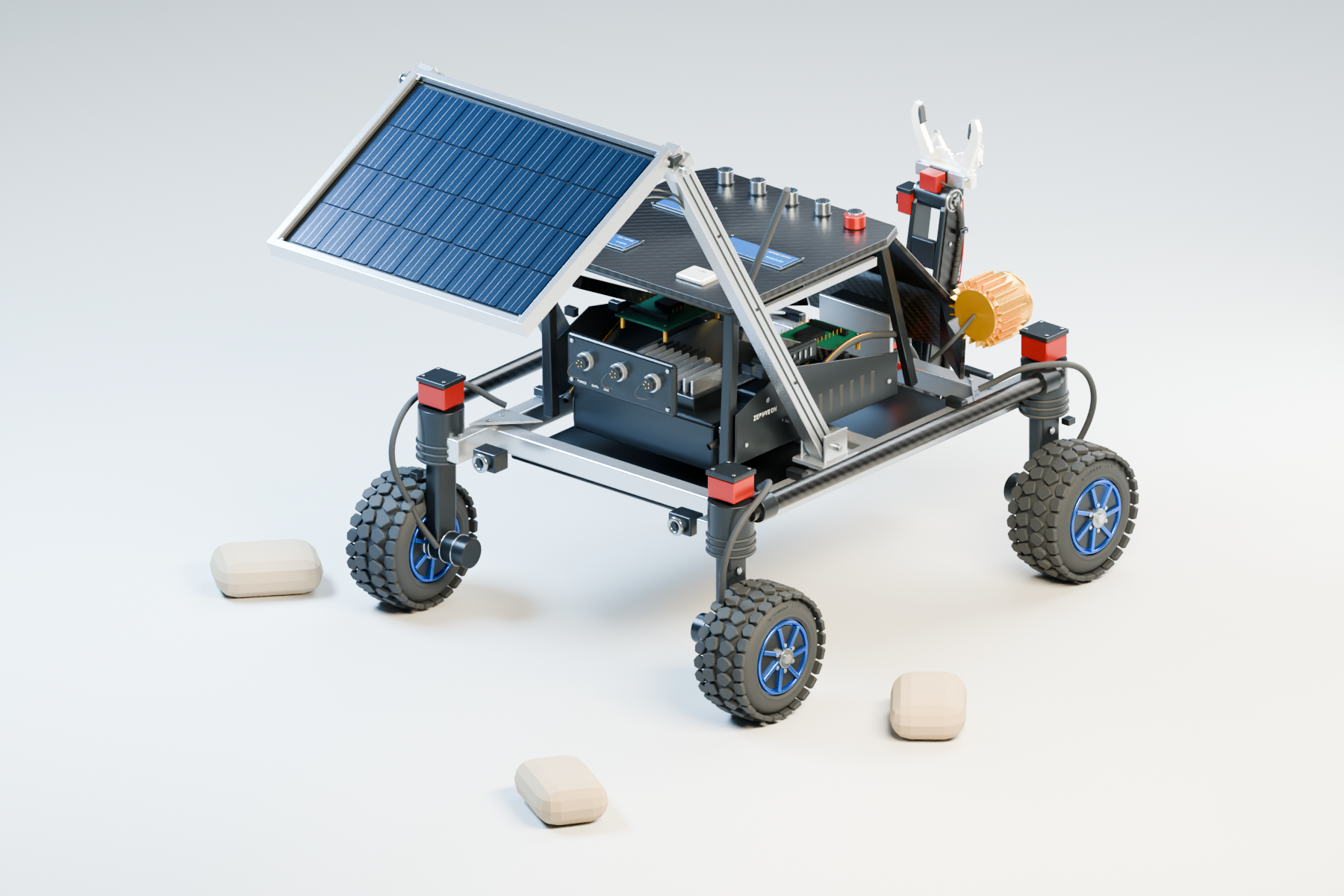}
\caption{Stationary solar-assisted monitoring with the selected rear panel arrangement. Solar input and UV response require their own calibration and operating conditions. Blender mission illustration of proposed behavior.}
\label{fig:MISSION_SOLAR}
\end{figure}

\FloatBarrier
\section{Vision inference, data integrity and validation methodology}

\subsection{Distributed camera and inference architecture}

The proposed vision subsystem separates image acquisition on the rover from object detection on an external computer. An ESP32-CAM acquires compressed camera frames, transmits them over Wi-Fi, and associates each frame with a sequence identifier and timing metadata. The external host decodes the frame, applies the detector's documented preprocessing, performs inference, and forwards the image and associated detections to the operator interface. This allocation reflects the camera driver's memory and buffering constraints; it does not imply that a full YOLO detector has been demonstrated on the ESP32. Larger image formats, additional frame buffers and simultaneous wireless communication require explicit memory-budget verification~\cite{ENG_ESP32},~\cite{ENG_CAMERA}. The accompanying camera and recording architectures identify the processing boundary and archival branch.

The acquisition record will specify the camera module, available memory, firmware commit, pixel format, frame dimensions, JPEG settings, buffer policy, and wireless configuration. The host record will specify processor, accelerator, operating system, inference-runtime version, model hash, input dimensions, numerical precision and postprocessing parameters. These records are experimental inputs. Published detector or recording-firmware benchmarks establish the feasibility of candidate software, but their throughput and accuracy will not be transferred to Zephyron~\cite{ENG_MJPEG},~\cite{ENG_SCALED}.

The offered video load depends on both frame frequency and compressed-frame size. For acquisition frequency \(f_c\) and mean JPEG size \(\bar B\) bytes, the payload-only rate is

\begin{equation}
R_{\mathrm{payload}}=8 f_c\bar B
\label{eq:21}
\end{equation}

This estimate excludes protocol overhead, retransmissions and telemetry. Experiments will therefore log actual transmitted bytes, received sequence identifiers, frame-size distributions and missing frames. A bounded buffer retaining the newest available frame is proposed for live inference. Discarding superseded frames limits queue growth, while a separate recording branch preserves the requested archival behavior. The recording policy, storage failures and any resulting resource contention must be logged rather than concealed as detector failures.

\begin{figure}[htbp]
\centering
\includegraphics[width=6.800in,height=3.850in,keepaspectratio]{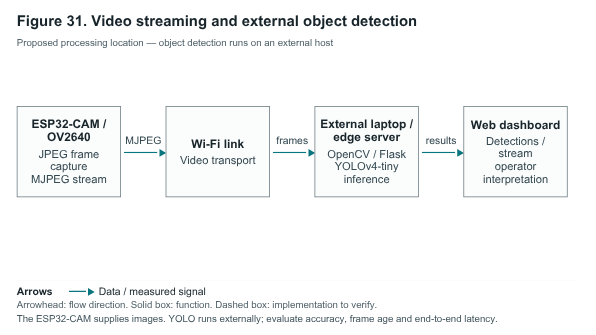}
\caption{Camera capture and external inference. Proposed architecture with arrow conventions defined in the legend. Native editable Canva design, rendered by Canva.}
\label{fig:CANVA_F10}
\end{figure}

\begin{figure}[htbp]
\centering
\includegraphics[width=6.800in,height=3.850in,keepaspectratio]{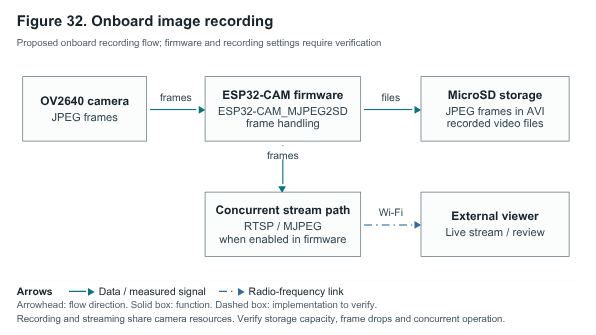}
\caption{Local recording. Proposed architecture with arrow conventions defined in the legend. Native editable Canva design, rendered by Canva.}
\label{fig:CANVA_F11}
\end{figure}

\subsection{Candidate detectors and training objective}

The proposed model comparison makes the external-host inference task explicit. MobileNetV3-Small with an SSDLite head is the compact candidate; MobileNetV3-Large with the same head tests the cost of additional feature capacity. Both combinations have primary-paper precedents~\cite{ENG_MOBILENETV3}. YOLOv4-tiny remains a third comparator~\cite{ENG_SCALED}. None is claimed to be installed, trained or measured on Zephyron. Their role is to locate a useful accuracy, latency and energy tradeoff on the selected host, rather than infer suitability from a model name.

The selected comparison input is one RGB frame resampled to 320 by 320 pixels. This resolution is a proposed common starting point, not a recovered prototype setting. Preprocessing will record color order, intensity normalization, aspect-ratio handling and the inverse box transform. Outputs are class identifiers, scores and bounding boxes in the original image coordinates, with the capture identifier attached. An initial label guide will distinguish visible persons, designated retrieval objects and obstacle categories that annotators can define consistently. A two-dimensional box alone will not supply object distance or traversability.

\begingroup
\small
\setlength{\tabcolsep}{4pt}
\renewcommand{\arraystretch}{1.14}
\begin{longtable}{@{}>{\raggedright\arraybackslash}p{0.236\linewidth}>{\raggedright\arraybackslash}p{0.236\linewidth}>{\raggedright\arraybackslash}p{0.236\linewidth}>{\raggedright\arraybackslash}p{0.236\linewidth}@{}}
\caption{Proposed model comparisons and required evidence}\label{tab:5}\\
\toprule
\rowcolor{tablehead} \textbf{Candidate} & \textbf{Input and output} & \textbf{Comparison purpose} & \textbf{Required evidence} \\
\midrule
\endfirsthead
\toprule
\rowcolor{tablehead} \textbf{Candidate} & \textbf{Input and output} & \textbf{Comparison purpose} & \textbf{Required evidence} \\
\midrule
\endhead
\bottomrule
\endfoot
MobileNetV3-Small + SSDLite & RGB; boxes and class scores & Compact host detector & Held-out AP, event recall and latency \\[2pt]
MobileNetV3-Large + SSDLite & Same image task & Additional feature capacity & Matched preprocessing and runtime \\[2pt]
YOLOv4-tiny & Same image task & Different detector family & Same test encounters and host \\[2pt]
Calibration curve / random forest & Voltage and measured temperature; conductivity & Simple versus nonlinear calibration & Independent reference samples and batch-held-out error \\[2pt]
\end{longtable}
\endgroup

For the SSD-family candidates, the declared reference objective combines class cross-entropy and smooth-L1 localization error over encoded anchor offsets~\cite{ENG_SSD}:

\begin{equation}
L_{\mathrm{det}}=\frac{L_{\mathrm{cls}}+\alpha L_{\mathrm{loc}}}{N_+},\qquad N_+>0
\label{eq:22}
\end{equation}

Here the losses, positive-anchor count \(N_+\) and weight \(\alpha\) are dimensionless. Anchor matching, negative mining and loss normalization will be retained from the pinned implementation. The original SSD convention assigns zero loss when no anchor is matched; the selected implementation's handling of negative-only images will therefore be checked explicitly. Background-only scenes remain essential in evaluation. A proposed initial training search uses pretrained weights with a replaced class head, SGD with momentum 0.9, learning rates 0.001 and 0.0003, and three recorded random seeds. These are search settings, not optimized values. Epoch limits, augmentation and early stopping will be fixed before test access; model selection will use validation encounters.

\subsection{Leakage control and calibrated interpretation}

The dataset protocol below remains authoritative. Its dependency groups will be assigned to training, validation, calibration and test partitions before fitting. The supplied hash-based splitter uses selected proportions 55/15/15/15 and a saved seed; it preserves assignments when additional groups are appended. It does not discover correlated observations or guarantee balanced classes. Site, run and repeated-object dependencies must be resolved first, and empty or poorly represented partitions require more acquisition or a narrower claim. A final site-held-out evaluation is preferable when independent sites permit it. Test groups will never enter augmentation, threshold selection, probability calibration or quantization calibration.

If class logits are available, a scalar temperature may be fitted on a separate calibration subset, following classification-calibration practice~\cite{ENG_TEMPERATURECAL}:

\begin{equation}
\begin{gathered}p_{ik}(T)=\frac{\exp(z_{ik}/T)}{\sum_j\exp(z_{ij}/T)},\\
T^*=\operatorname{argmin}_{T>0}\left[-\sum_i\log p_{i,y_i}(T)\right]\end{gathered}
\label{eq:23}
\end{equation}

Logits \(z\), temperature \(T\) and probabilities are dimensionless. For an anchor detector, the calibration record must define the anchor labels, background sampling and matching procedure; these cannot change after observing test reliability. This operation concerns class probabilities and does not calibrate localization correctness. Emitted-detection reliability, duplicate penalties and missed-object recall will consequently remain separate outcomes under the subsequent evaluation protocol. Calibration will be omitted if sufficient independent calibration groups or compatible logits are unavailable.

Sensor learning has a narrower justification. A candidate random-forest regressor may map the water channel's raw voltage in volts and independently measured sample temperature in degrees Celsius to reference electrical conductivity at 25 degrees Celsius, in microsiemens per centimetre. This requires a calibrated reference meter, a separate temperature measurement and controlled sample handling; these instruments are proposed experimental requirements. Labels must not be generated from the same vendor voltage equation being assessed~\cite{ENG_TDS_CODE},~\cite{ENG_USGS_EC}. The forest prediction is

\begin{equation}
\widehat{\kappa}_{25}(\mathbf{x})=\frac{1}{B}\sum_{b=1}^{B}h_b(\mathbf{x})
\label{eq:24}
\end{equation}

Each tree predicts conductivity in the same units; \(B\) is the tree count~\cite{ENG_RANDOMFOREST}. A selected 200-tree candidate will use squared-error splits, bootstrap sampling and validation-selected depth and leaf size. It will be compared with a fitted low-order calibration curve using the same inputs. Grouping by physical sample batch and acquisition day prevents repeated aliquots from leaking across partitions. Report bias, MAE and RMSE in conductivity units and residuals against temperature and reference level. Tree-to-tree spread will not be presented as a calibrated uncertainty interval. Without reference labels, the forest will not be fitted; a single MQ-2 signal will not be assigned gas identities or safety categories by analogy, and water conductivity will not become a potability label.

\subsection{Executable checks and deployment acceptance}

The archive contains standard-library Python implementations for deterministic splitting, finite-box IoU, class-aware one-to-one matching and conservative frame admission. These execute without ML frameworks. They test data handling with explicitly constructed fixtures; their outputs are not detector performance. The matching helper covers ordinary boxes at one threshold, while reported COCO AP still requires the official evaluator and its ignore/crowd rules~\cite{ENG_COCO}.

Let \(\widehat A\) be estimated image age in seconds and \(e\) a nonnegative clock-error bound in seconds. The proposed admission check is

\begin{equation}
\mathrm{admit}=I_{\mathrm{identity}}I_{\mathrm{quality}}
\mathbf{1}[\widehat A-e\geq0]\mathbf{1}[\widehat A+e\leq A_{\max}]
\label{eq:25}
\end{equation}

The identity flag includes session, sequence and displayed-image agreement. An unknown offset fails admission. A standard uncertainty must be converted to a justified bound before use; passing this gate does not establish collision avoidance. Listing 1 demonstrates the boundary with chosen timestamps.

\begin{minipage}{\linewidth}
\begin{lstlisting}[caption={Conservative frame-age admission using explicit clock bounds},label={lst:1}]
from validation_primitives import admit_frame

# Constructed fixture: all time quantities are seconds.
accepted = admit_frame(
    capture_s=10.0, now_s=10.5, offset_s=0.25,
    error_s=0.0625, limit_s=0.3125,
    identity_ok=True, quality_ok=True)
assert accepted  # Age interval: [0.1875, 0.3125] s.
assert not admit_frame(
    10.0, 10.5, 0.25, 0.0625, 0.30, True, True)
assert not admit_frame(
    10.0, 10.5, None, 0.0625, 0.3125, True, True)
assert not admit_frame(
    10.0, 10.5, 0.25, 0.0625, 0.3125, False, True)
print({"fixture_accepted": accepted})
\end{lstlisting}
\end{minipage}

Listing 2 makes the duplicate penalty executable. Both listings use the included \texttt{validation\_primitives.py}; complete runnable files are supplied alongside it.

\begin{minipage}{\linewidth}
\begin{lstlisting}[caption={One reference object cannot produce two true positives},label={lst:2}]
from validation_primitives import (
    BoxRecord, match_detections, rates)

# Constructed boxes, not outputs from a trained detector.
truth = [BoxRecord("target", (0, 0, 2, 2))]
predictions = [
    BoxRecord("target", (0, 0, 2, 2), 0.9),
    BoxRecord("target", (0, 0, 2, 2), 0.8)]
counts = match_detections(truth, predictions, 0.5, 0.5)
assert counts == {"tp": 1, "fp": 1, "fn": 0}
assert rates(counts)["precision"] == 0.5
empty = match_detections([], [])
assert empty == {"tp": 0, "fp": 0, "fn": 0}
assert rates(empty)["precision"] is None
print({"fixture_counts": counts})
\end{lstlisting}
\end{minipage}

Model deployment requires an additional verification stage. Archive weights and hashes, framework/export versions, host specifications and preprocessing. Compare floating-point and INT8 exports on identical saved frames; inspect operator fallback and output agreement before timing. Quantization inputs must come from representative development groups~\cite{ENG_QUANT},~\cite{ENG_ONNX}. Measure warm-up separately, then inference distributions, peak memory, full-pipeline age, dropped frames and host energy under matched recording/network conditions. A model will be selected from validation evidence subject to declared resource limits, then assessed once on the held-out test protocol. No trained weights, accuracy improvement or runtime advantage is asserted here.

\subsection{Timing, frame age and supervised operation}

Timing will be instrumented at acquisition, transmission, host reception, decode completion, inference completion and presentation. Each result must retain the identifier of the image on which it was computed; displaying a detection over a later image would produce an invalid association even if both images appeared plausible. A monotonic clock and an explicit boot/session identifier will distinguish sequence resets. Host-side intervals can be measured directly, whereas cross-device intervals require a documented clock-offset estimate and its uncertainty. A browser-render acknowledgment will be identified as such; physical display delay requires an optical check.

Let \(t_{c,i}^{r}\) denote a defined acquisition event on the rover clock, \(t_{d,i}^{h}\) the presentation event on the host clock, and \(\hat\theta\) the estimated host-minus-rover clock offset. The estimated end-to-end latency of frame \(i\) is

\begin{equation}
\hat L_i=t_{d,i}^{h}-t_{c,i}^{r}-\hat\theta
\label{eq:26}
\end{equation}

The acquisition event must be identified precisely. A timestamp taken when a frame buffer becomes available is not automatically an exposure timestamp. Where exposure timing is unavailable, the reported interval will be named for those measured endpoints and characterize the acquisition delay using a timed optical event visible to the camera. If clock synchronization cannot be bounded, host-reception-to-presentation latency will be reported as a partial interval; it will not be relabelled as complete scene-to-operator latency. Median and upper-tail latency will be reported per run, together with frame loss, frame rejection and delivered-frame frequency.

The display will retain the capture identifier, age indication and communication state. If a result exceeds a preregistered age limit, it will be marked stale and excluded from current-scene assistance. The proposed detector provides operator information, while a separate command supervisor manages motion authority. Loss of inference must not silently transform into an affirmative ``clear path'' indication. The basic avoidance and manual-control functions require their own validation independently of recognition performance.

The selected design cap of \(v_{\max}=0.35\ \mathrm{m\,s^{-1}}\) is an operating constraint, not a measured capability or demonstrated safe speed. A candidate freshness limit must be related to available clearance and measured stopping behavior. Under a constant-deceleration approximation, a conservative design inequality is

\begin{equation}
d_{\mathrm{clear}}\geq v_{\max}(A_{\max}+T_{\mathrm{command}})
+\frac{v_{\max}^{2}}{2a_{\min}}+m
\label{eq:27}
\end{equation}

Here \(A_{\max}\) is the admitted image age, \(T_{\mathrm{command}}\) is the subsequent command-path delay, \(a_{\min}\) is a demonstrated minimum braking deceleration within the tested conditions, and \(m\) accounts for measurement and geometric allowances. All quantities other than the selected cap require characterization. This inequality is a proposed screening model; terrain-dependent slip, slopes and unobserved obstacles prevent it from establishing general collision avoidance.

\subsection{Dataset construction and separation of development from evaluation}

A task-specific image corpus will be collected from the rover's intended viewing height, orientation and camera settings. Acquisition will include relevant object classes, unobstructed negative scenes, partial occlusions, changing illumination, motion blur and visually similar distractors. Class definitions, minimum annotation requirements, treatment of heavily occluded objects and ignore regions will be fixed in an annotation guide. A second annotator will independently review a documented subset; disagreements will be adjudicated before evaluation. Neither a detection label nor environmental-sensor readings will be treated as ground truth merely because the platform produced them.

Images will be grouped by site and acquisition run before splitting. Adjacent video frames, repeated traversals of the same scene and derived crops will remain in the same group. When repeated object identity could reveal a test instance, that identity will also be considered in grouping. A site-held-out test partition is preferred when sufficient independent sites are available. Where this is infeasible, the narrower run-held-out scope and its limitations will be stated explicitly. Randomly distributing neighboring frames across partitions would test recognition of near-duplicate observations rather than transportable mission performance.

The four-way development and evaluation partition described above governs fitting, model selection and calibration. Quantization inputs will be drawn only from representative development groups. Site, run, frame and instance counts will be reported after acquisition; no dataset counts or detection results are currently claimed.

\subsection{Detection metrics and score interpretation}

Localization matching will use intersection over union between a predicted box \(B_p\) and a reference box \(B_g\):

\begin{equation}
\operatorname{IoU}(B_p,B_g)=
\frac{|B_p\cap B_g|}{|B_p\cup B_g|}
\label{eq:28}
\end{equation}

At a declared class, score threshold and IoU threshold, predictions will be matched to reference objects using the selected evaluator's documented procedure. Each reference object can contribute at most one true positive, with unmatched predictions counted as false positives and unmatched eligible references as false negatives. Duplicate detections must therefore remain penalized. For the resulting counts,

\begin{equation}
P=\frac{TP}{TP+FP},\qquad
R=\frac{TP}{TP+FN},\qquad
F_1=\frac{2TP}{2TP+FP+FN}
\label{eq:29}
\end{equation}

Each metric is undefined only when its own denominator is zero, and the underlying counts will be reported. Thus F1 is zero when TP is zero but FP or FN is positive, and undefined when TP, FP and FN are all zero. The direct-count expression equals 2PR/(P + R) when precision and recall are defined and P + R is positive. Per-class results will be retained so that strong performance on a common class cannot obscure missed instances of a less frequent class. Precision and recall at the operating threshold answer a different question from threshold-swept average precision.

The proposed primary detection summary follows the COCO evaluator's IoU thresholds and interpolated precision-recall convention~\cite{ENG_COCO}. With \(C\) eligible classes, the threshold set \(\mathcal T=\{0.50,0.55,\ldots,0.95\}\), and interpolated precision \(p^{*}_{c,\tau}(r)\),

\begin{equation}
\begin{gathered}\operatorname{AP}_{c,\tau}
=\frac{1}{101}\sum_{j=0}^{100}p^{*}_{c,\tau}(j/100),
\\
\operatorname{AP}
=\frac{1}{10C}\sum_{c=1}^{C}\sum_{\tau\in\mathcal T}\operatorname{AP}_{c,\tau}\end{gathered}
\label{eq:30}
\end{equation}

The primary summary will use at most 100 detections per image, the all-area range, and the documented COCO ignore/crowd semantics. The evaluator version and annotation conversion will be fixed and disclosed. AP50 will be named separately from AP averaged across IoU thresholds. Application-specific event recall will supplement these metrics: an event is a predefined object encounter, not every correlated frame containing that object. The definition of a timely event detection must be established before testing.

Detector scores will not be assumed to be calibrated probabilities. A proposed reliability analysis will compare score bins against the fraction of emitted predictions that satisfy the fixed class-and-localization matching criterion. For \(M\) eligible emitted predictions, scores \(s_i\), and correctness indicators \(y_i\in\{0,1\}\), a complementary score diagnostic is

\begin{equation}
\operatorname{BS}_{\mathrm{emitted}}
=\frac{1}{M}\sum_{i=1}^{M}(s_i-y_i)^2
\label{eq:31}
\end{equation}

This conditional Brier score characterizes emitted predictions only; it cannot account for unreported objects and cannot replace recall. Bin counts, matching rules and any fitted score transformation will accompany calibration plots. Results will also be examined by site, illumination, object size and occlusion, provided each subgroup has enough independent evidence to support interpretation.

\subsection{Quantization and matched computational comparisons}

A floating-point detector on the selected external host will establish the reference implementation. An integer-quantized version may then be evaluated using the same architecture and weights lineage. For an affine quantizer with scale \(s\), zero point \(z\) and representable integer limits,

\begin{equation}
\begin{gathered}q=\operatorname{clip}\!\left(\operatorname{round}(x/s)+z,q_{\min},q_{\max}\right),
\\
\hat x=s(q-z)\end{gathered}
\label{eq:32}
\end{equation}

Quantization changes numerical representation; an actual speed improvement depends on operator support, hardware execution and conversion overhead~\cite{ENG_QUANT},~\cite{ENG_ONNX}. The comparison will therefore report model size, memory consumption, warm inference-time distributions, loading/warm-up time and end-to-end performance, alongside detection metrics. Unsupported operators and floating-point fallback must be disclosed.

Floating-point and quantized models will first process identical stored frames with identical preprocessing and matching settings. A second experiment will compare their complete live pipelines under matched camera and network conditions. Additional runs with recording disabled and enabled will isolate the cost of SD activity. Repeated conditions will be interleaved within site/session blocks, with order randomized where practical. Transport losses, thermal state and resource contention will be recorded so that changes in received images are not mistaken for changes in model quality.

\subsection{Uncertainty, comparators and the progression of evidence}

Measurement uncertainty and sampling uncertainty will be treated separately. Timing uncertainty includes timestamp resolution, clock-offset estimation and the definition of acquisition and display events. For a derived quantity \(y=f(\mathbf{x})\), the first-order combined standard uncertainty is

\begin{equation}
\begin{gathered}u_c^2(y)=\sum_i c_i^2u^2(x_i)
+2\sum_{i<j}c_i c_j\operatorname{cov}(x_i,x_j),
\\ c_i=\frac{\partial f}{\partial x_i}\end{gathered}
\label{eq:33}
\end{equation}

The uncertainty budget will identify measured repeatability and externally assigned inputs, and state the coverage convention when an expanded uncertainty is reported~\cite{ENG_GUM}. This propagation does not substitute for a confidence interval on dataset-level detection performance. Because frames within runs are correlated, run- or site-level resampling will be used for proposed interval estimation rather than treating each image as an independent replicate.

Binary encounter outcomes will retain exact counts. For zero misses in \(n\) independent positive encounters, the one-sided upper confidence bound on miss probability is

\begin{equation}
p_{\mathrm{miss},U}=1-\alpha^{1/n}
\label{eq:34}
\end{equation}

This expression uses confidence level \(1-\alpha\) and the binomial independence assumption~\cite{ENG_BINOM}. Repeated frames of one encounter do not increase \(n\). For nonzero misses, an appropriate exact interval will be calculated. Required encounter counts will be set from a prespecified precision or upper-bound objective, not selected retrospectively to support a favorable claim.

The telemetry interface in Figure 33 will retain timestamped measurements, frame identifiers, settings changes and operator interventions. Manipulator trials following Figure 13 will record task conditions and control outcomes without inferring grasp capacity from appearance. The mission sequence in Figure 34 provides the structure for synchronizing these records. Until the proposed experiments are completed, the subsystem remains an engineering design and evaluation plan; no training accuracy, detection improvement, stopping performance or field effectiveness is asserted.

Software verification was performed separately from the proposed model evaluation. Under Python 3.12.14, all 33 deterministic checks of the supplied validation primitives passed with zero failures or errors. The cases cover finite and degenerate boxes, IoU geometry, class-aware matching, score ordering and ties, duplicate predictions, missed objects, threshold boundaries and undefined rates for empty inputs. Timing cases reject unknown clock offsets, nonfinite values, inconsistent identities, invalid quality flags and age intervals extending beyond the admitted boundary. Group-split checks verify a known hash assignment and invariance to input order and appended groups. The two printed listings were also checked against their runnable files and executed. Separately, 12 analytical regression tests passed, including reserve violation, intermediate capacity saturation, infeasible mission duration, network overload and parameter perturbations. The revised generators reproduced 9,026 previously calculated numerical values without a mismatch under the comparison tolerance, while extending the archived curve grids. These checks establish consistency of the declared formulas and software behavior on specified fixtures. They do not establish detector accuracy, calibrated sensor performance, real-time deadlines or field reliability. Source hashes, inputs and test outcomes are retained so the verification can be repeated after a code change.

\begin{figure}[htbp]
\centering
\includegraphics[width=6.800in,height=3.850in,keepaspectratio]{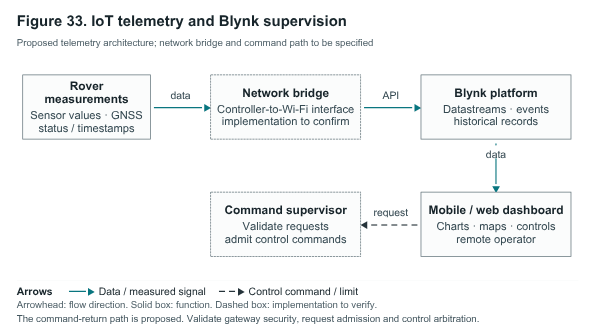}
\caption{IoT telemetry and dashboard. Proposed architecture with arrow conventions defined in the legend. Native editable Canva design, rendered by Canva.}
\label{fig:CANVA_F12}
\end{figure}

\begin{figure}[htbp]
\centering
\includegraphics[width=6.800in,height=3.850in,keepaspectratio]{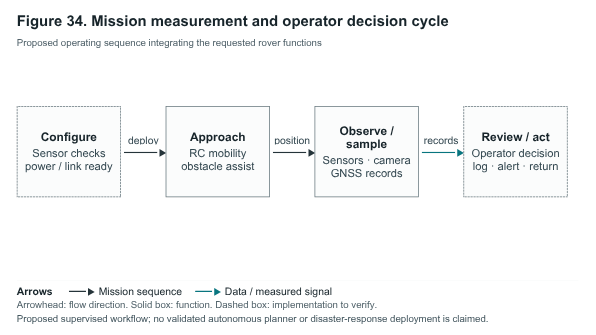}
\caption{Mission measurement and operator decision cycle. Proposed architecture with arrow conventions defined in the legend. Native editable Canva design, rendered by Canva.}
\label{fig:CANVA_F14}
\end{figure}

Evaluation will proceed from documented component feasibility to integrated bench tests, controlled supervised traversals, held-out sites and, only after authorization and successful preceding tests, application-specific field studies. Each stage supports a narrower claim than the next. Manufacturer specifications, mathematical design estimates, simulations, recorded bench observations and field measurements will be identified separately. Comparative motion trials will hold the selected speed cap, route, payload configuration and environmental conditions as consistent as practicable, while comparing teleoperation with the proposed assistance enabled and disabled. Outcomes will include completion, interventions, communication interruptions, stale-frame exposure and contact events, alongside elapsed time.

\begin{figure}[htbp]
\centering
\includegraphics[width=6.300in,height=4.800in,keepaspectratio]{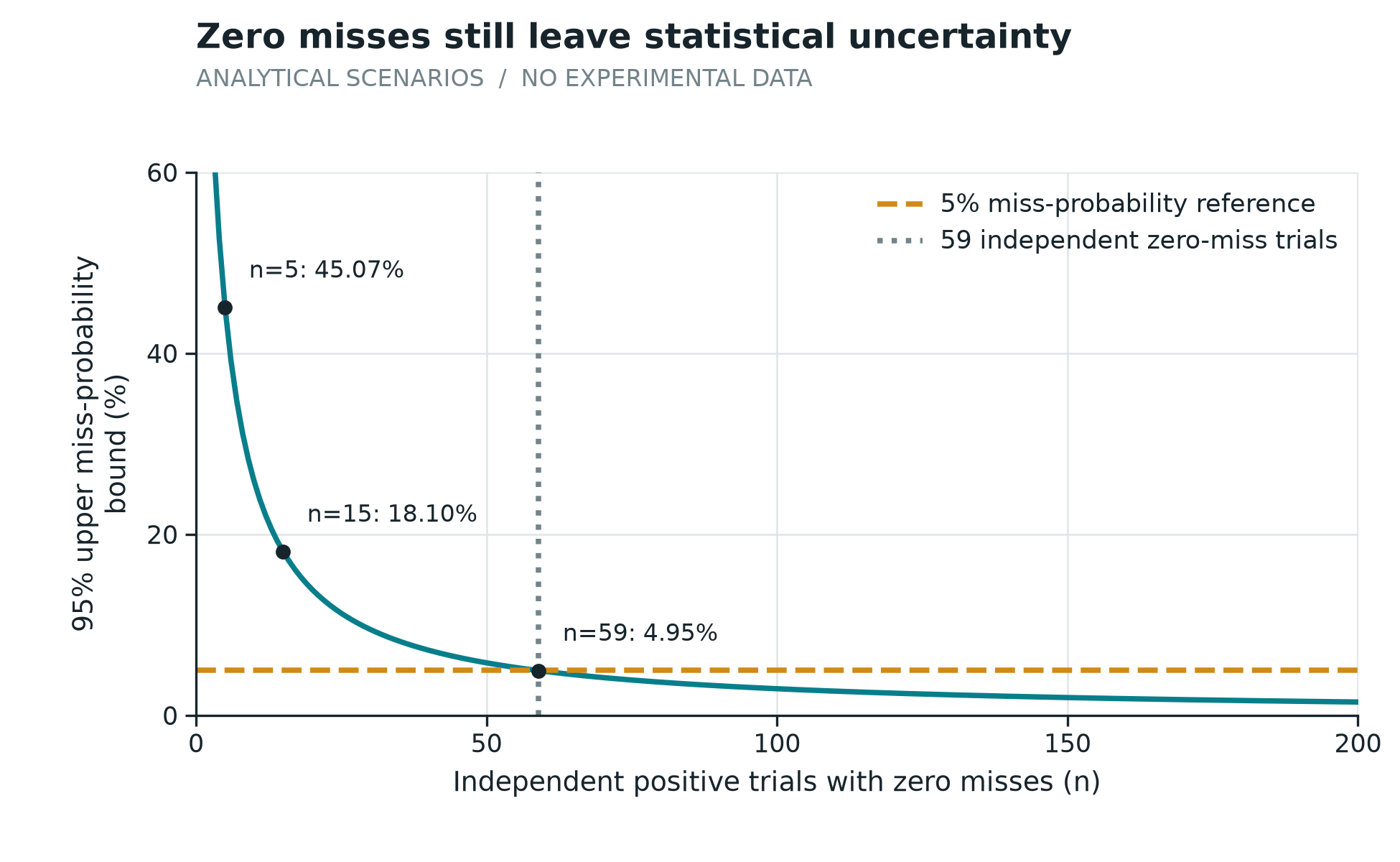}
\caption{Detection trial count and uncertainty. Analytical sample-size calculation; no trials were performed. For zero misses in n independent positive Bernoulli trials, the exact one-sided 95\% upper miss-probability bound is 1 \ensuremath{-} 0.05\textasciicircum{}(1/n). At least 59 zero-miss trials are needed to place this bound below 5\% for the tested condition distribution. Repeated frames or readings from one exposure are not necessarily independent trials~\cite{ENG_BINOM}.}
\label{fig:ENG_F08}
\end{figure}

\begin{figure}[htbp]
\centering
\includegraphics[width=6.650in,height=2.100in,keepaspectratio]{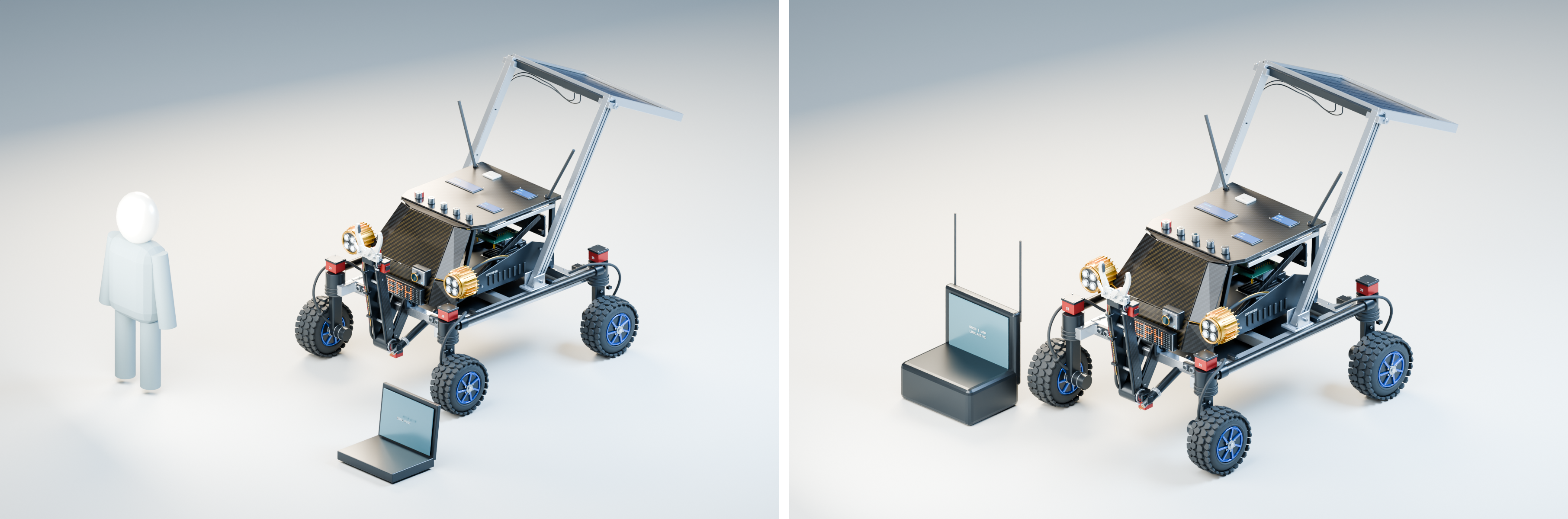}
\caption{Distributed vision and telemetry mission concepts. Left: the forward camera observes a training mannequin while an external terminal represents inference processing. Right: a ground station receives position, logging, and link-status information. Both are Blender illustrations of proposed behavior, not detection results or field trials.}
\label{fig:Distributed_missions}
\end{figure}

\subsection{Validation matrix}

The validation unit should be a run, an exposure, or an independently prepared sample as appropriate. A large number of frames or ADC samples from one event does not by itself increase the number of independent trials.

\begingroup
\small
\setlength{\tabcolsep}{4pt}
\renewcommand{\arraystretch}{1.14}
\begin{longtable}{@{}>{\raggedright\arraybackslash}p{0.353\linewidth}>{\raggedright\arraybackslash}p{0.316\linewidth}>{\raggedright\arraybackslash}p{0.294\linewidth}@{}}
\caption{Proposed subsystem validation matrix}\label{tab:6}\\
\toprule
\rowcolor{tablehead} \textbf{Subsystem} & \textbf{Proposed reference or comparator} & \textbf{Primary recorded outcomes} \\
\midrule
\endfirsthead
\toprule
\rowcolor{tablehead} \textbf{Subsystem} & \textbf{Proposed reference or comparator} & \textbf{Primary recorded outcomes} \\
\midrule
\endhead
\bottomrule
\endfoot
Mobility & Surveyed route and controlled surfaces & Speed, slip, tracking error, current, temperature \\[2pt]
Steering and stopping & Defined geometry and timed command loss & Angle error, response time, stopping distance \\[2pt]
Arm retrieval & Weighed inert objects and force measurement & Retention, joint current, deflection, release failures \\[2pt]
Battery and solar & Calibrated voltage and current logging & Delivered Wh, converter loss, charge state, irradiance \\[2pt]
Water screening & Independent conductivity and turbidity references & Bias, drift, repeatability, temperature, sample identity \\[2pt]
Gas screening & Controlled exposure and reference analyzer & Response and recovery, interferents, misses, false alarms \\[2pt]
Radiation counting & Specified detector reference and background procedure & Counts, live time, background, uncertainty, geometry \\[2pt]
GNSS and telemetry & Surveyed positions and synchronized records & Fix error distribution, outages, loss, frame age \\[2pt]
Computer vision & Site or run held-out annotated sequences & Per-class AP, precision, recall, latency, frame loss \\[2pt]
\end{longtable}
\endgroup

All comparisons use the same declared task definition and record unsuccessful runs. Acceptance thresholds must be fixed before evaluating held-out data. A failed attempt remains part of the outcome distribution even if a later attempt succeeds.

\FloatBarrier
\section{Proximity supervision, illumination, and fault handling}

The following functional diagrams complete the proposed interface description. Their role is to expose dependencies between acquisition, storage, supervision, and indication. Each function requires a tested implementation and a documented failure response before integration acceptance.

The proximity channel provides a short-range input to supervised motion. A reading is accepted only when its acquisition time, validity state and configured range checks are available. A timeout or missing reading is recorded as unavailable; it must not be converted into a declaration that the path is clear. The controller distinguishes this condition from a valid observation beyond a declared intervention distance. Consecutive plausibility checks and hysteresis can reduce rapid command switching, but the filtering delay must be included in the command-age budget.

Coverage is checked using targets at different heights, approach angles and lateral offsets, because the sensing footprint need not coincide with the camera view or the complete wheel-and-arm envelope. A stationary chassis also does not prevent a moving manipulator from contacting an obstacle. The trial protocol therefore tests chassis stopping, steering transitions and arm motion as separate operating states, logging the active state with every intervention. Fault injection should include disconnected sensors, stale messages and a lost operator link. The corresponding stop or hold response is assessed against a declared procedure before the proposed behavior is used in a field trial.

The two front work lamps support illumination and inspection, while the local status display reports operating state. Their acceptance tests should record switched current, supply droop and enclosure temperature alongside camera exposure settings. A brighter image does not by itself improve recognition: glare, backscatter from nearby particles, clipped highlights and automatic-exposure transitions can remove useful scene detail. Matched recordings with lamps enabled and disabled should therefore use the same scene and camera configuration, retaining the exposure metadata. Lamp duty and measured electrical demand enter the mission power ledger explicitly. Electrical switching and dashboard state must agree with the observed lamp state; a status icon alone is insufficient evidence that illumination was available during an acquisition.

\begin{figure}[htbp]
\centering
\includegraphics[width=6.800in,height=3.850in,keepaspectratio]{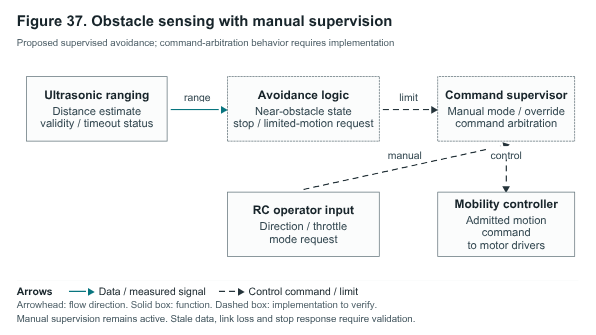}
\caption{Proximity sensing and supervised avoidance. Proposed architecture with arrow conventions defined in the legend. Native editable Canva design, rendered by Canva.}
\label{fig:CANVA_F09}
\end{figure}

\begin{figure}[htbp]
\centering
\includegraphics[width=6.800in,height=3.850in,keepaspectratio]{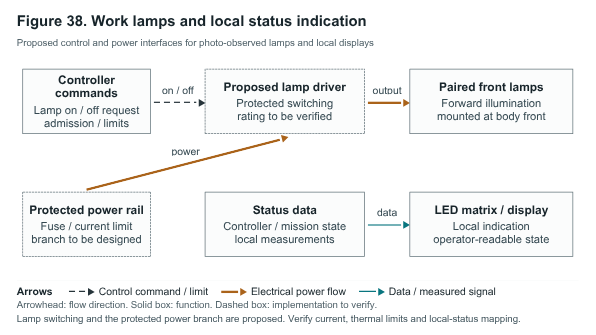}
\caption{Work lamps and local status indication. Proposed architecture with arrow conventions defined in the legend. Native editable Canva design, rendered by Canva.}
\label{fig:CANVA_F15}
\end{figure}

\section{Discussion and engineering implications}

The calculations identify several different limiting mechanisms. Steady grade torque is compatible with an initial motor sizing scenario, but that finding does not include turning scrub, impacts, sinkage, or thermal accumulation. A substantial chassis payload allocation does not imply a similar arm payload because the manipulator amplifies gravity loads through its reach. Photovoltaic input can change the budget for a low-duty mission, yet a small panel does not support unlimited operation under changing weather. These results favor a supervised, intermittently stationary research platform whose limits are measured explicitly.

Measurement quality is also task dependent. Counting uncertainty can often be reduced through longer integration, whereas an uncalibrated turbidity conversion cannot be repaired merely by averaging more voltage readings. Gas-sensor drift and interferents can dominate electronic resolution. GNSS accuracy must retain its statistical definition, and consecutive fixes do not provide independent replications of an environmental trial. Remote computer vision introduces an additional coupling between bandwidth, frame age, and the distance traveled before a result is shown. The proposed logging schema is intended to preserve the information needed to examine these effects.

The photographic design constraint makes this study different from an unconstrained replacement of the rover. The arm, panel supports, wheel stance, lamps, and faceted body remain in their recognizable locations. The selected component baseline introduces engineering specificity within that arrangement, while industrial details such as restrained wiring, inspection covers, serviceable fasteners, and separated power distribution support plausible construction. These details communicate design intent; they do not substitute for ingress testing, structural analysis, or a manufacturing drawing set.

The main research opportunity is a matched evaluation of measurement-aware operation. A future experiment can compare continuous movement with fixed-rate acquisition against stop-and-sample operation using the same hardware, route, exposure conditions, and initial battery state. Outcomes should include valid sample fraction, spatial coverage, uncertainty, total mission energy, missed events, and operator interventions. An energy saving is meaningful only when task quality remains comparable. Conversely, a slower policy may be preferable when its additional dwell time makes an otherwise ambiguous measurement interpretable.

The analytical models are deliberately compact. Equal wheel load sharing, constant rolling resistance, fixed PV derating, an assumed auxiliary load, and simplified link masses reduce the number of unknowns enough to expose tradeoffs. They also limit numerical transfer to field conditions. Future work should replace assumptions with measured rolling resistance and current maps, weighed mass properties, calibrated sensor models, and time-varying irradiance. Uncertainty propagation should then include correlations and model discrepancy, rather than only sensor noise.

\FloatBarrier
\section{Conclusion}

Zephyron has been specified as a literature-informed environmental reconnaissance rover while preserving the photographed mechanical arrangement. The resulting baseline, mission illustrations, subsystem diagrams, and deterministic data make the design inspectable and reproducible. Analytical results identify the separate constraints imposed by traction, manipulator reach, usable battery energy, sensor integration, calibration, and distributed vision latency. The proposed quality-aware mission policy connects those constraints to explicit decisions about travel, sampling, and return reserve. Explicit detector candidates, a conditional conductivity-regression baseline, and executable frame-admission and matching checks connect the system architecture to an assessable software methodology. These elements constitute a design and analysis framework, with code checks establishing implementation consistency rather than learned-model performance. A subsequent experimental study must establish component integration, reference calibration, operating limits, and held-out task performance before the platform can support measured capability claims.

\FloatBarrier

\end{document}